\documentclass{article}

 \usepackage[preprint]{neurips_2025}

\usepackage[utf8]{inputenc} 
\usepackage[T1]{fontenc}    
\usepackage{hyperref}       
\usepackage{url}            
\usepackage{booktabs}       
\usepackage{amsfonts}       
\usepackage{nicefrac}       
\usepackage{microtype}      
\usepackage{xcolor}         

\usepackage{multirow}
\usepackage{graphicx}
\usepackage{color}
\usepackage{natbib}
\usepackage{graphicx}
\usepackage{subcaption}
\usepackage{amsmath}
\usepackage{tabularx}
\usepackage{booktabs}
\usepackage[most]{tcolorbox}
\usepackage{colortbl}
\usepackage{graphicx}
\usepackage{subcaption}
\usepackage{xcolor}
\usepackage{array}
\usepackage{longtable}
\definecolor{headerblue}{RGB}{41, 98, 143}
\definecolor{catblue}{RGB}{210, 228, 245}
\definecolor{rowgray}{RGB}{245, 248, 252}
\usepackage[table]{xcolor}

\usepackage[dvipsnames]{xcolor}
\usepackage{xspace}

\newcommand\llama{\texttt{Llama-3.1-8B-Instruct}}
\newcommand\qwen{\texttt{Qwen3-8B}}
\newcommand\mone{\texttt{m1-7b-23k}}
\newcommand\medreason{\texttt{MedReason-8B}}
\newcommand\medgemma{\texttt{medgemma-1.5-4b-it}}
\newcommand\deepseek{\texttt{Deepseek-R1}}
\newcommand\gpt{\texttt{GPT-5-mini}}

\title{Improving Clinical Diagnosis with Counterfactual Multi-Agent Reasoning}

\author{%
  Zhiwen You$^{1}$\thanks{e-mail: \url{zhiweny2@illinois.edu}} \quad
  Xi Chen$^{2}$ \quad
  Aniket Vashishtha$^{1}$ \quad
  Simo Du$^{3}$ \AND
  Gabriel Erion-Barner$^{4}$ \quad
  Hongyuan Mei$^{5}$\quad
  Hao Peng$^{1}$\quad
  Yue Guo$^{1}$\thanks{Corresponding author. e-mail: \url{yueg@illinois.edu}} \\
  \\
  $^{1}$University of Illinois Urbana-Champaign, Urbana, IL, USA \quad \\
  $^{2}$Independent Researcher \quad \\
  $^{3}$Jacobi Medical Center, Albert Einstein College of Medicine, NY, USA\\
  $^{4}$Department of Emergency Medicine, Beth Israel Deaconess Medical Center, Boston, MA, USA \quad \\
  $^{5}$Leaning machines, Palo Alto, CA, USA \\
}

\begin{document}

\maketitle

\begin{abstract}
Clinical diagnosis is a complex, iterative reasoning process in which clinicians gather evidence, form hypotheses, and continuously test them against alternative explanations. 
In medical training, this reasoning is explicitly developed through counterfactual questioning---e.g., asking how a diagnosis would change if a key symptom were absent or altered---to strengthen differential diagnosis skills.
As large language model (LLM)–based systems are increasingly used for diagnostic support and second opinions, ensuring the interpretability of their recommendations becomes critical. 
However, most existing LLM-based diagnostic agents reason over fixed clinical evidence without explicitly testing how individual findings support or weaken competing diagnoses, limiting the transparency, reliability, and clinical trust of their outputs.
In this work, we propose a counterfactual multi-agent diagnostic framework inspired by clinician training that makes hypothesis testing explicit and evidence-grounded. 
Our framework introduces counterfactual case editing to systematically modify clinical findings and evaluate how these changes affect competing diagnoses. 
We further define the Counterfactual Probability Gap, a method that quantifies how strongly individual findings support a diagnosis by measuring confidence shifts under these edits. 
These counterfactual signals guide multi-round specialist discussions, enabling agents to challenge unsupported hypotheses, refine differential diagnoses, and produce more interpretable reasoning trajectories.
Across three diagnostic benchmarks and seven LLMs, our method consistently improves diagnostic accuracy over prompting and prior multi-agent baselines, with the largest gains observed in complex and ambiguous cases. 
Human evaluation by licensed physicians further indicates that our framework produces more clinically useful, reliable, and coherent reasoning.
These results suggest that incorporating explicit counterfactual evidence verification is an important step toward building reliable AI systems for clinical decision support, aligning machine reasoning more closely with how clinicians are trained to evaluate diagnoses.

\end{abstract}

\section{Introduction}

Clinical diagnosis is a core medical decision-making task that requires physicians to interpret patient findings, compare competing hypotheses, and identify the most likely explanation for a patient's presentation. Despite extensive medical training, diagnostic errors remain a major source of patient harm, leading to delayed treatment, inappropriate interventions, and, in severe cases, preventable morbidity and mortality \cite{ball2015improving, ibrahim2021diagnostic, sox2024medical}. A key contributor is anchoring bias, where clinicians over-rely on initial impressions and insufficiently revise their hypotheses in light of new evidence \cite{sox2024medical, ly2023evidence, vally2023errors, webster2021cognitive}.
In clinical training, physicians are explicitly taught to mitigate such biases through iterative hypothesis testing, including counterfactual questioning (e.g., how a diagnosis would change if a key finding were absent or altered). This process encourages systematic evaluation of evidence and refinement of differential diagnoses.
Large language models (LLMs) have demonstrated strong performance across medical applications, such as diagnosis prediction \cite{chen2025enhancing, wu2025medcasereasoning, liu2025generalist}, medical text summarization \cite{guo2021automated, you2024uiuc_bionlp}, and EHR-based risk prediction \cite{yang2023pyhealth, jiangreasoning}. However, accurate clinical diagnosis requires more than generating plausible answers: it demands the ability to justify decisions, distinguish among competing hypotheses, and revise conclusions when evidence is inconsistent. 
This gap highlights the need for LLM-based systems that support interpretable clinical decision support and second opinions by enabling explicit, evidence-grounded diagnostic reasoning aligned with how clinicians are trained to evaluate diagnoses.

Existing LLM-based approaches to clinical diagnosis largely rely on standard prompting strategies (e.g., zero-shot, few-shot, and chain-of-thought (CoT) prompting \cite{kojima2022large}) or on multi-agent frameworks that simulate multidisciplinary discussion \cite{chen2025enhancing, tang2024medagents, kim2024mdagents, peng2025tree}. While these methods have improved diagnostic prediction, they exhibit several important limitations. 
First, most approaches reason over fixed clinical evidence without explicitly testing whether a diagnosis remains valid under alternative assumptions. In contrast, clinical training emphasizes counterfactual reasoning to rule out competing hypotheses—for example, asking how the diagnosis would change if key findings were absent or different \cite{sox2024medical}. This evidence-testing process is critical for ensuring that a diagnosis is both supported and internally consistent, yet it is rarely made explicit in current LLM-based systems.
Second, although multi-agent frameworks introduce structured discussion\cite{chen2025enhancing, kim2024mdagents}, they typically rely on fixed interaction protocols and do not explicitly verify how individual clinical findings contribute to diagnostic decisions\cite{tang2024medagents, kim2024mdagents, peng2025tree}. As a result, their reasoning remains difficult to audit and may fail to detect unsupported or inconsistent hypotheses.
Third, many existing approaches depend mainly on proprietary LLMs, limiting their applicability in clinical settings where local deployment, data privacy, and interpretability are essential \cite{chen2025enhancing, kim2024mdagents, tang2024medagents}.
Together, these limitations highlight the need for diagnostic systems that make reasoning explicit, verifiable, and grounded in clinical evidence.

To address these challenges, we propose a counterfactual case editing-based multi-agent diagnostic framework that turns the question ``what if not?'' into an explicit evidence-checking step during diagnosis. Rather than reasoning over fixed evidence, the framework enables specialist agents to generate counterfactual variants of the clinical case by modifying selected findings, allowing them to evaluate how competing diagnoses change under alternative evidence conditions.
By comparing model predictions across original and edited cases, the framework identifies which clinical findings are most critical for supporting or distinguishing competing diagnoses. This transforms counterfactual reasoning into an interpretable mechanism for testing the relationship between evidence and diagnostic decisions, rather than relying solely on forward reasoning.
To quantify this process, we introduce the Counterfactual Probability Gap (CPG), which quantifies how diagnostic confidence shifts under targeted evidence operations. Within the multi-agent framework, specialist agents use these signals to refine hypotheses, challenge unsupported conclusions, and guide discussion toward evidence-consistent diagnoses. 
Instead of requiring full consensus, a final judge agent synthesizes the discussion and resolves disagreements based on the most relevant clinical evidence.

We evaluate our framework on the clinical diagnosis task across three datasets (MIMIC-CDM-FI \cite{hager2024evaluation} (MIMIC), MedCaseReasoning \cite{wu2025medcasereasoning}, and ER-Reason \cite{mehandru2025er}) using a diverse set of LLMs, including five open-source models (\llama, \space \qwen, \space \mone, \space \medreason, and \space \medgemma) and two stronger models with undisclosed training data (\deepseek \space and \space \gpt). These models span general-purpose and medically specialized systems, as well as varying scales and reasoning capabilities. 
Our results show that the proposed framework consistently improves diagnostic accuracy across all three datasets, outperforming standard prompting strategies (e.g., zero-shot and few-shot) and strong multi-agent baselines, including MAC \cite{chen2025enhancing}, MedAgents \cite{tang2024medagents}, and MDAgents \cite{kim2024mdagents}.
We further conduct detailed analyses of multi-round discussion behavior and ablation studies to evaluate the contribution of each component of the framework.
Beyond quantitative performance, human evaluation by two licensed physicians (S.D. and G.E.B.) shows that our framework produces reasoning traces with fewer factual errors, stronger logical coherence, and higher clinical trust than standard zero-shot CoT.

Incorporating counterfactual evidence testing enables explicit verification of diagnostic hypotheses, allowing models to move beyond plausible predictions toward more interpretable, evidence-supported decisions. These observations motivate the integration of counterfactual evidence testing into LLM-based diagnostic systems, bringing machine reasoning closer to how clinicians are trained to evaluate and refine diagnoses. Strong performance on open-source models further highlights the potential of our approach for privacy-sensitive and resource-constrained clinical settings, where local deployment and interpretability are essential, pointing toward more trustworthy and deployable clinical decision support systems.

\begin{figure*}[!h]
  \centering
    \includegraphics[width=1\linewidth,height=9.5cm,keepaspectratio]{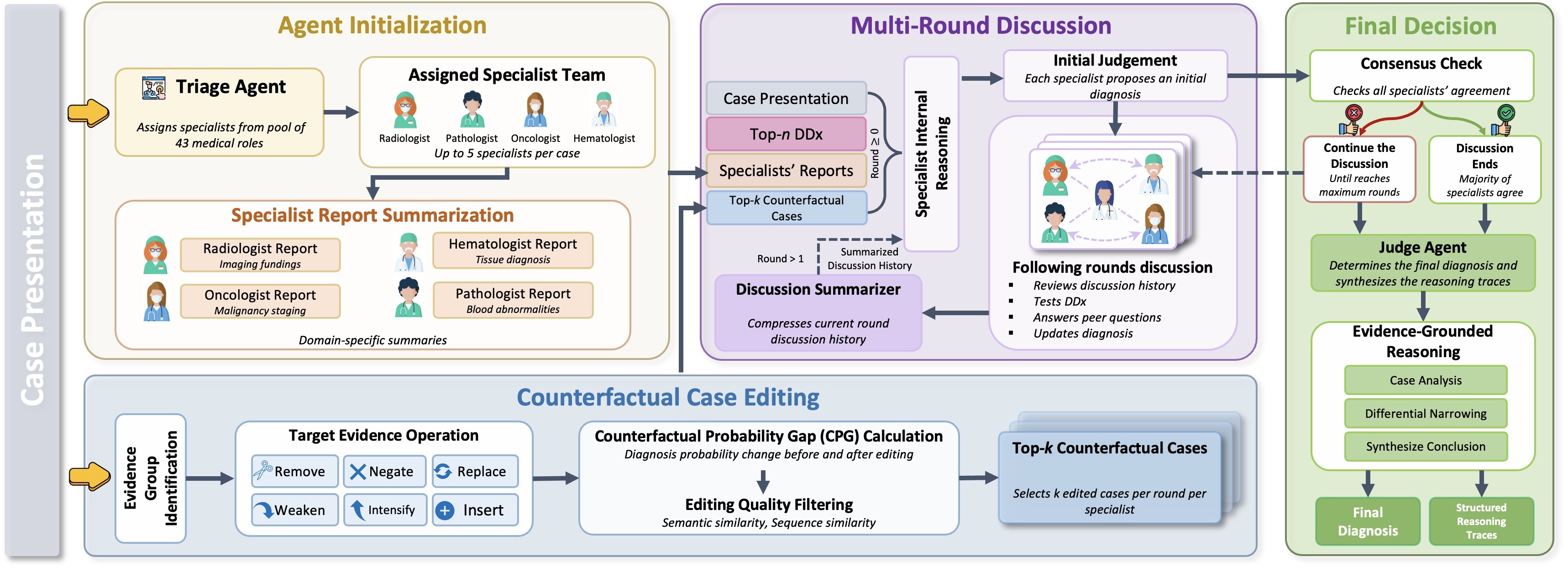}

  \caption{Overview of the proposed counterfactual case editing-based multi-agent diagnostic framework. Given a clinical case, a triage agent selects a set of relevant medical specialists, and an initial differential diagnosis (DDx) is generated to guide subsequent reasoning. During multi-round discussion, each specialist performs counterfactual case editing by modifying targeted clinical findings to test competing diagnostic hypotheses. The impact of these edits is quantified using the Counterfactual Probability Gap (CPG), which measures changes in diagnostic confidence before and after the target evidence operation. These counterfactual signals guide iterative discussion, enabling specialists to explicitly test how individual clinical findings support or weaken competing diagnoses. A final judge agent synthesizes the discussion into an evidence-grounded diagnostic decision and structured reasoning trace.
  }
  \label{fig:system}
\end{figure*}

\section{Results}

\subsection{Overview of the counterfactual multi-agent diagnostic system} \label{sec:overview}
Inspired by clinician training \cite{sox2024medical}, we propose a multi-agent diagnostic framework that introduces counterfactual evidence testing to improve the reliability and interpretability of AI-assisted diagnosis (Figure~\ref{fig:system}). The framework consists of three main stages: specialist assignment, counterfactual evidence testing during multi-round discussion, and evidence-grounded final decision-making.
Our framework organizes multiple specialist agents to collaboratively generate and refine differential diagnoses from a given clinical case. Rather than relying solely on discussion over fixed evidence, the system incorporates a novel counterfactual case editing mechanism that enables agents to explicitly test diagnostic hypotheses by modifying key clinical findings and observing their impact on predicted diagnoses. This process allows the system to identify which findings are most critical for supporting or refuting competing diagnoses.
To quantify the contribution of individual clinical features, we introduce the Counterfactual Probability Gap (CPG), which measures how diagnostic confidence changes under targeted evidence operations. By integrating signals from multiple counterfactual edits, the framework guides multi-round specialist discussions toward more evidence-grounded and interpretable conclusions.
Together, these components transform multi-agent diagnostic reasoning from implicit discussion to explicit hypothesis testing, improving diagnostic accuracy while providing interpretable justification for model predictions. Further implementation details are provided in the Methods (Section~\ref{sec:method}).

\subsection{Performance of diagnosis prediction across LLMs} \label{sec:main-results}

\begin{figure*}[!ht]
  \centering
  \begin{subfigure}[t]{0.75\textwidth}
    \centering
    \includegraphics[width=\linewidth,keepaspectratio]{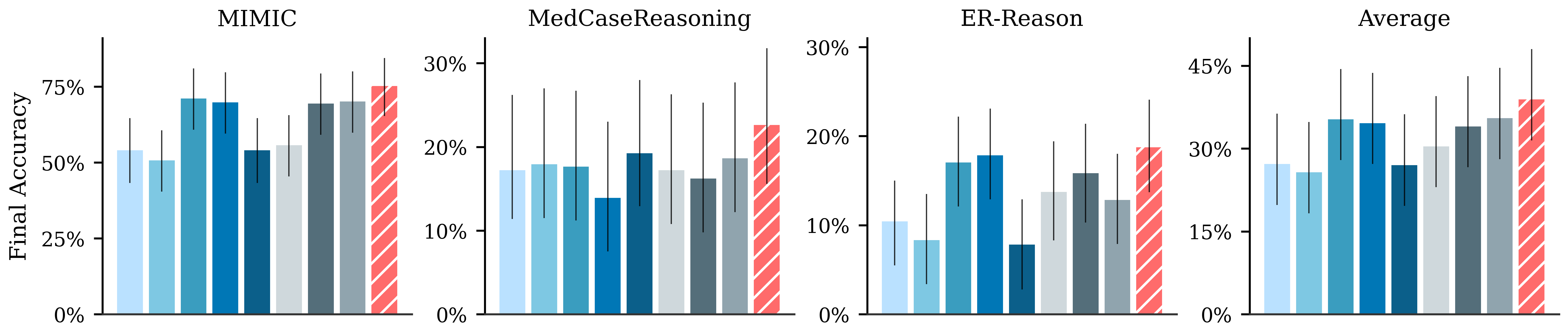}
    \caption{}
    \label{fig:llama-baseline}
  \end{subfigure}

  \begin{subfigure}[t]{0.75\textwidth}
    \centering
    \includegraphics[width=\linewidth,keepaspectratio]{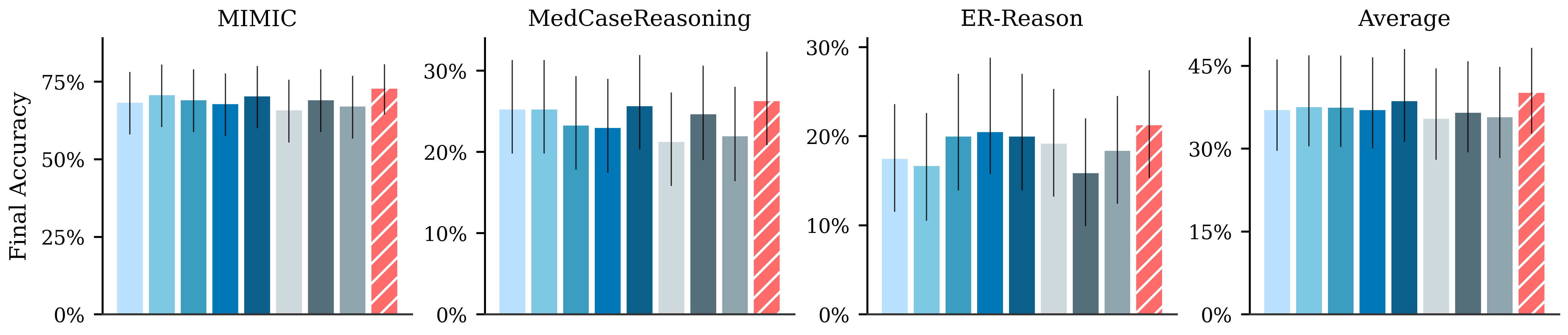}
    \caption{}
    \label{fig:qwen-baseline}
  \end{subfigure}

  \begin{subfigure}[t]{0.75\textwidth}
    \centering
    \includegraphics[width=\linewidth,keepaspectratio]{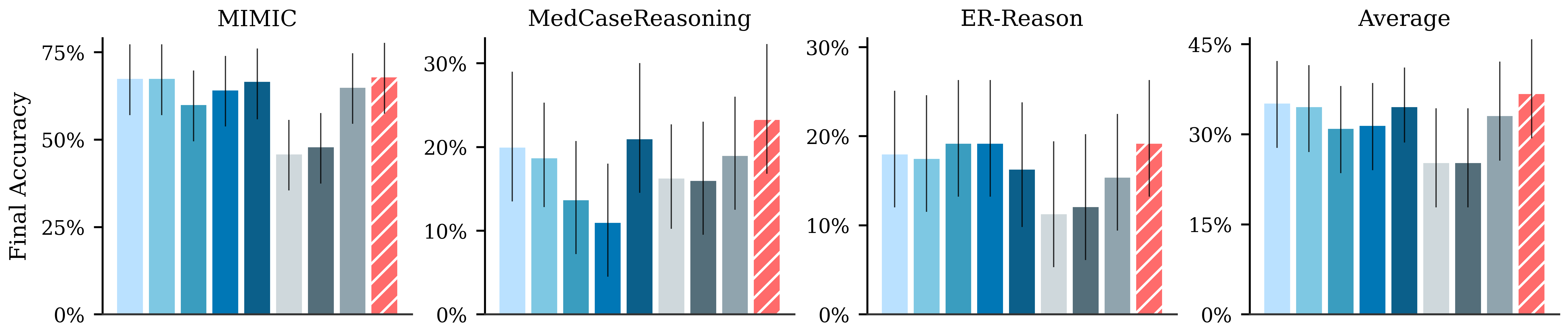}
    \caption{}
    \label{fig:m1-baseline}
  \end{subfigure}

  \begin{subfigure}[t]{0.75\textwidth}
    \centering
    \includegraphics[width=\linewidth,keepaspectratio]{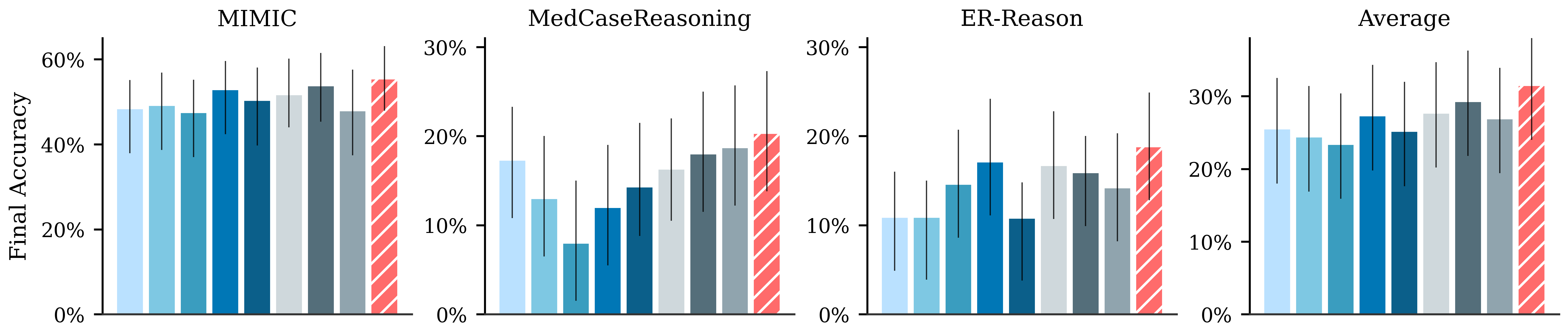}
    \caption{}
    \label{fig:medreason-baseline}
  \end{subfigure}

  \begin{subfigure}[t]{0.75\textwidth}
    \centering
    \includegraphics[width=\linewidth,keepaspectratio]{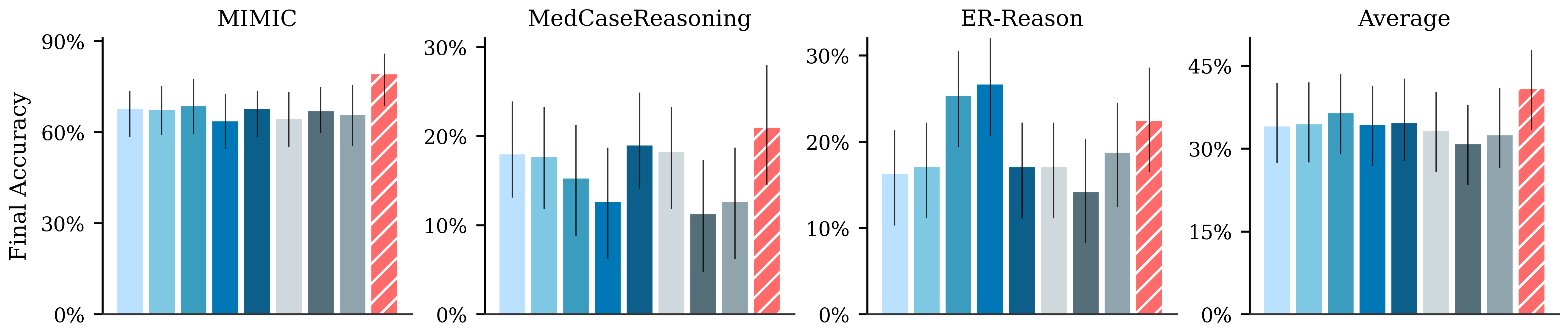}
    \caption{}
    \label{fig:medgemma-baseline}
  \end{subfigure}

  \begin{subfigure}[t]{0.75\textwidth}
    \centering
    \includegraphics[width=\linewidth,keepaspectratio]{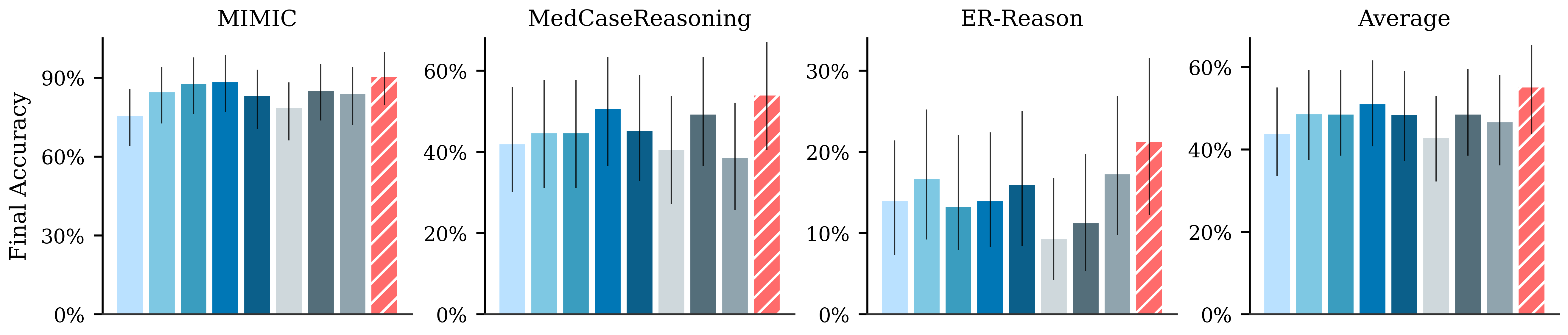}
    \caption{}
    \label{fig:deepseek-baseline}
  \end{subfigure}

  \begin{subfigure}[t]{0.75\textwidth}
    \centering
    \includegraphics[width=\linewidth,keepaspectratio]{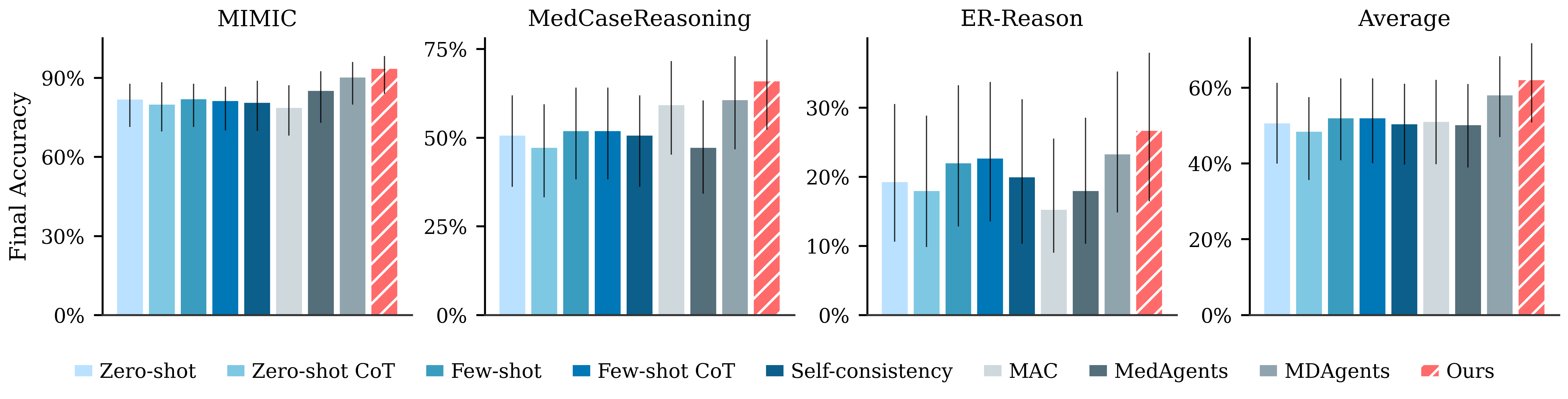}
    \caption{}
    \label{fig:gpt-baseline}
  \end{subfigure}

  \caption{Average diagnostic accuracy of seven LLMs on three datasets, including \textbf{(a)} \llama, \textbf{(b)} \qwen, \textbf{(c)} \mone, \textbf{(d)} \medreason, \textbf{(e)} \medgemma, \textbf{(f)} \deepseek, and \textbf{(g)} \gpt. Bar graphs indicate the accuracy ± 95\% CIs. Numerical results and statistical significance tests are provided in Table~\ref{tab:main-results-std}.}
  \label{fig:baseline-results}
\end{figure*}

To evaluate our proposed system, we use three clinical diagnosis datasets: MIMIC-CDM-FI \cite{hager2024evaluation} (MIMIC), MedCaseReasoning \cite{wu2025medcasereasoning}, and ER-Reason \cite{mehandru2025er}. The MIMIC and ER-Reason datasets are collected from real-world emergency room (ER) cases in hospitals, while MedCaseReasoning is curated from publicly available case reports in PubMedCentral (Appendix~\ref{appx:datasets}). 
These datasets collectively cover diverse clinical scenarios, ranging from real-world emergency presentations to structured case-based reasoning, enabling a comprehensive evaluation of diagnostic performance. Notably, ER-Reason was released after the evaluated open-source models, reducing the risk that their performance reflects exposure to this dataset during training. 
We compare several commonly used prompting baselines, including single-agent prompting strategies (zero-shot, zero-shot CoT, few-shot , few-shot CoT, and self-consistency \cite{wangself}) following prior work \cite{wu2025medcasereasoning, kim2024mdagents}, as well as three advanced multi-agent methods: MAC \cite{chen2025enhancing}, MDAgents \cite{kim2024mdagents}, and MedAgents \cite{tang2024medagents}. For single-agent baselines, few-shot uses five-shot examples, and self-consistency uses three reasoning paths following Zhao et al. \cite{zhao2025confagents}. The selection of multi-agent baselines is informed by the prior benchmarking study MedAgentBoard~\citep{zhu2025medagentboard}, which identifies these methods as strong and representative approaches for medical diagnosis.

We first evaluate the performance of LLMs on diagnosis prediction across the three datasets. Following prior work in MedCaseReasoning \cite{wu2025medcasereasoning}, we select five representative open-source models spanning both general-purpose and medically specialized LLMs: \llama \space (\texttt{Llama}) \cite{llama3-2024}, \qwen \space (\texttt{Qwen}) \cite{yang2025qwen3}, \mone \space (\texttt{m1}) \cite{huang2025m1}, \medreason \space (\texttt{MedReason}) \cite{wu2025medreason}, and \medgemma \space (\texttt{medgemma}). Among these, \texttt{m1}, \texttt{MedReason}, and \texttt{medgemma} are designed for medical reasoning tasks, \texttt{Qwen} is a general-purpose reasoning model, and \texttt{Llama} is a general non-reasoning model. 
This selection enables a systematic comparison of our framework across models with different training objectives and reasoning capabilities.
Detailed descriptions of these models are provided in Appendix~\ref{appx:baseline-models}.

As shown in Figure~\ref{fig:baseline-results}, our method achieves the highest average diagnostic accuracy across all evaluated models and datasets. In particular, \texttt{Llama} and \texttt{MedReason} reach average accuracies of 39.0\% and 31.5\%, respectively, consistently outperforming all baselines. The performance gains are most pronounced for \texttt{Llama}, with a 13.2\% improvement over zero-shot CoT, while gains for \texttt{m1} are more modest (+2.2\%).
These differences reflect variation in model capabilities. Models such as \texttt{Qwen} and \texttt{m1}, which already perform strongly under simple prompting, benefit less from additional structured reasoning. In contrast, models with weaker baseline performance, such as \texttt{Llama} and \texttt{MedReason}, show substantial improvements, indicating that our framework is particularly effective when baseline reasoning is limited.

Across datasets, performance gains remain consistent despite differences in clinical characteristics, with MIMIC focusing on common abdominal conditions and MedCaseReasoning containing more complex and rare disease cases.  

Our approach achieves the highest accuracy on both MIMIC and MedCaseReasoning across all models. Although \texttt{MedReason} shows lower overall accuracy in MedCaseReasoning and MIMIC compared with other smaller LLMs, our approach still achieves the best performance for this model, with an average accuracy of 31.5\%. For \texttt{medgemma}, our method achieves the largest improvement on the MIMIC dataset, reaching an average accuracy of 79.2\% compared with all baselines. 
Overall, these results indicate that our framework provides consistent performance improvements across diverse diagnostic approaches and model types, while remaining effective even when strong single-agent prompting baselines are already competitive.

To further assess scalability, we evaluate our approach on larger models, \texttt{DeepSeek} and \texttt{GPT5mini}. As shown in Figures~\ref{fig:deepseek-baseline}--\ref{fig:gpt-baseline}, our method consistently outperforms all baselines.
For \texttt{Deepseek}, our approach achieves the highest average accuracy, surpassing the second-best method (few-shot CoT) by 4.1\%. \texttt{GPT5mini} achieves higher overall accuracy, reaching 93.6\% on MIMIC, 66.0\% on MedCaseReasoning, and 26.7\% on ER-Reason. Notably, the gains from our approach on the MIMIC dataset are smaller compared to the other two datasets for both models. This suggests that our method are more evident on more challenging diagnostic tasks, such as rare disease cases. A detailed breakdown by disease category and clinical specialty is provided in Section~\ref{sec:mimic-comparison}.

Although \texttt{Deepseek} and \texttt{GPT5mini} achieve the highest average accuracy across datasets (55.2\% for \texttt{Deepseek} and 62.1\% for \texttt{GPT5mini}), on the ER-Reason dataset, smaller LLMs show a notable advantage. Our approach using \texttt{medgemma} achieves 22.5\% accuracy on ER-Reason, outperforming \texttt{Deepseek} (21.3\%). Furthermore, using a standard few-shot CoT setting allows \texttt{medgemma} to reach 26.7\% accuracy, matching the performance of our approach with \texttt{GPT5mini}. We attribute this to the nature of ground-truth diagnoses in ER-Reason, which are broad and symptom-level (e.g., ``fever, unspecified fever cause'') instead of more specific disease names in other two datasets. LLMs such as \texttt{medgemma}, \texttt{m1}, and \texttt{GPT5mini} adapt more effectively to this diagnostic granularity under few-shot settings where example diagnoses reveal the expected diagnostic scope. In contrast, \texttt{Deepseek} struggles to learn from few-shot examples on ER-Reason—few-shot CoT (14.0\%) underperforms zero-shot CoT (16.7\%)—suggesting that \texttt{Deepseek} is less sensitive to in-context learning of label granularity. Despite these dataset-specific limitations, our approach with \texttt{Deepseek} and \texttt{GPT-5 mini} still achieves the highest average accuracy overall, clearly outperforming the best smaller LLM, \texttt{Qwen} (40.2\% average accuracy). These findings show that our system is effective across model scales, while also indicating that performance on ER-Reason is shaped by how well each model adapts to the diagnosis scope defined by the dataset.

\begin{figure*}[!ht]
  \centering

  \begin{subfigure}[t]{0.98\textwidth}
    \centering
    \includegraphics[width=\linewidth,keepaspectratio]{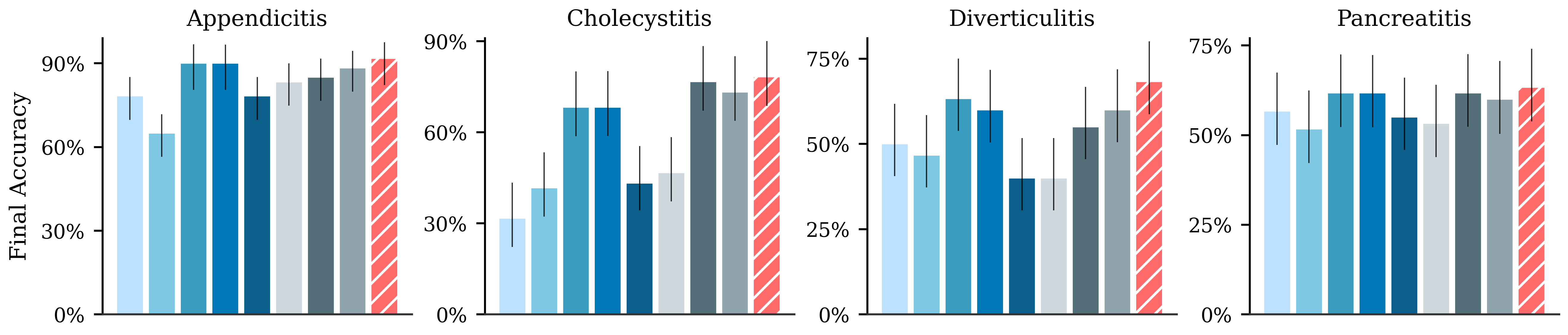}
    \caption{}
    \label{fig:llama-mimic}
  \end{subfigure}

  \begin{subfigure}[t]{0.98\textwidth}
    \centering
    \includegraphics[width=\linewidth,keepaspectratio]{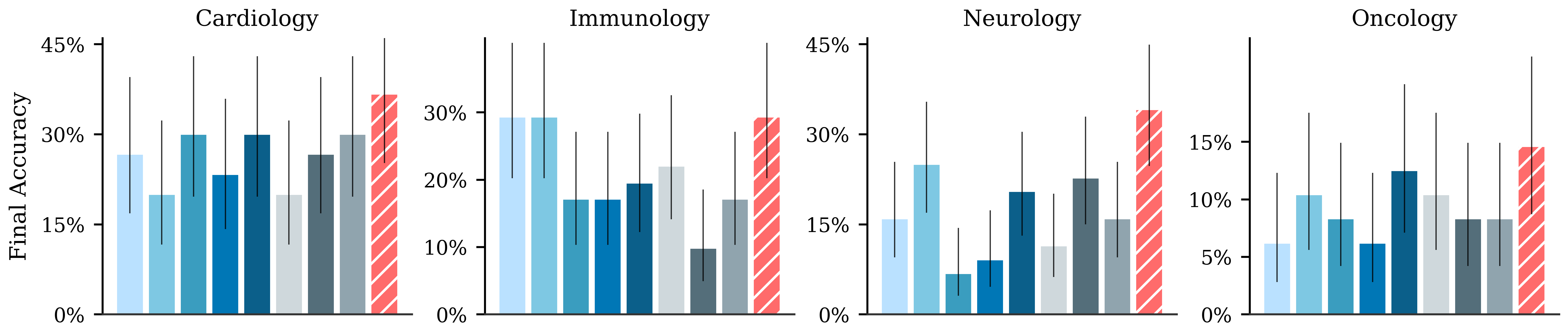}
    \caption{}
    \label{fig:llama-mcr}
  \end{subfigure}

  \begin{subfigure}[t]{0.98\textwidth}
    \centering
    \includegraphics[width=\linewidth,keepaspectratio]{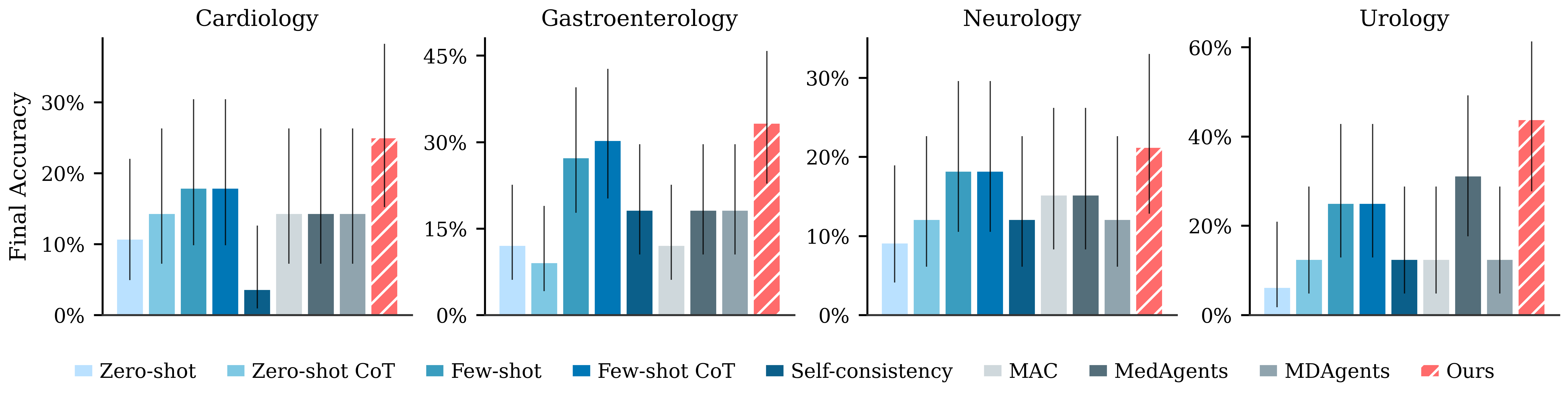}
    \caption{}
    \label{fig:llama-er}
  \end{subfigure}

  \caption{Average diagnostic accuracy of \llama \space for four diseases/specialties on three datasets, including \textbf{(a)} Disease-level accuracy on MIMIC, \textbf{(b)} Specialty-level accuracy on MedCaseReasoning, \textbf{(c)} Specialty-level accuracy on ER-Reason. Following Liu et al.\cite{liu2025generalist}, we categorize the test cases into specialties, and select the Top-4 most frequent specialties with relevant diagnoses from MedCaseReasoning and ER-Reason. Bar graphs indicate the accuracy ± 95\% CIs.}
  \label{fig:mimic-results}
\end{figure*}
\subsection{Performance comparison on different diseases/specialties over three datasets} \label{sec:mimic-comparison}

As illustrated in Figure~\ref{fig:mimic-results}, we report disease/specialty-level diagnostic accuracy of \texttt{Llama} in three datasets. Figure~\ref{fig:llama-mimic} shows four diseases categorized by the original MIMIC dataset \cite{hager2024evaluation}, while Figures~\ref{fig:llama-mcr} --~\ref{fig:llama-er} introduce top-4 specialties categorized using the grouping taxonomy introduced in Liu et al. \cite{liu2025generalist}. For MedCaseReasoning and ER-Reason, we map all ground-truth diagnoses to specialties \cite{liu2025generalist} and report the four most frequent ones. Overall, our method consistently achieves the highest diagnostic accuracy in all diseases/specialties across the datasets. In Figure~\ref{fig:llama-mimic}, appendicitis is generally the easiest condition to identify, and baseline methods already perform strongly. As a result, our method provides only modest improvements by 1.7\% compared with the second-best baseline (few-shot). However, the advantage of our method becomes more apparent on more challenging diseases and specialties. For example, in MedCaseReasoning (Figure~\ref{fig:llama-mcr}), our method substantially outperforms the second-best baselines across Cardiology, Neurology, and Oncology, while zero-shot and zero-shot CoT achieve competitive performance (29.3\%) compared to our approach in Immunology. In ER-Reason, likely due to the nature of the ground-truth diagnoses, the few-shot CoT setting remains competitive across most specialties. Overall, compared to standard prompting methods and existing multi-agent baselines, our approach offers more consistent diagnostic performance across diseases and specialties, particularly in more challenging cases.

\begin{figure*}[!ht]

  \begin{subfigure}[t]{0.23\textwidth}
    \centering
    \includegraphics[width=\linewidth]{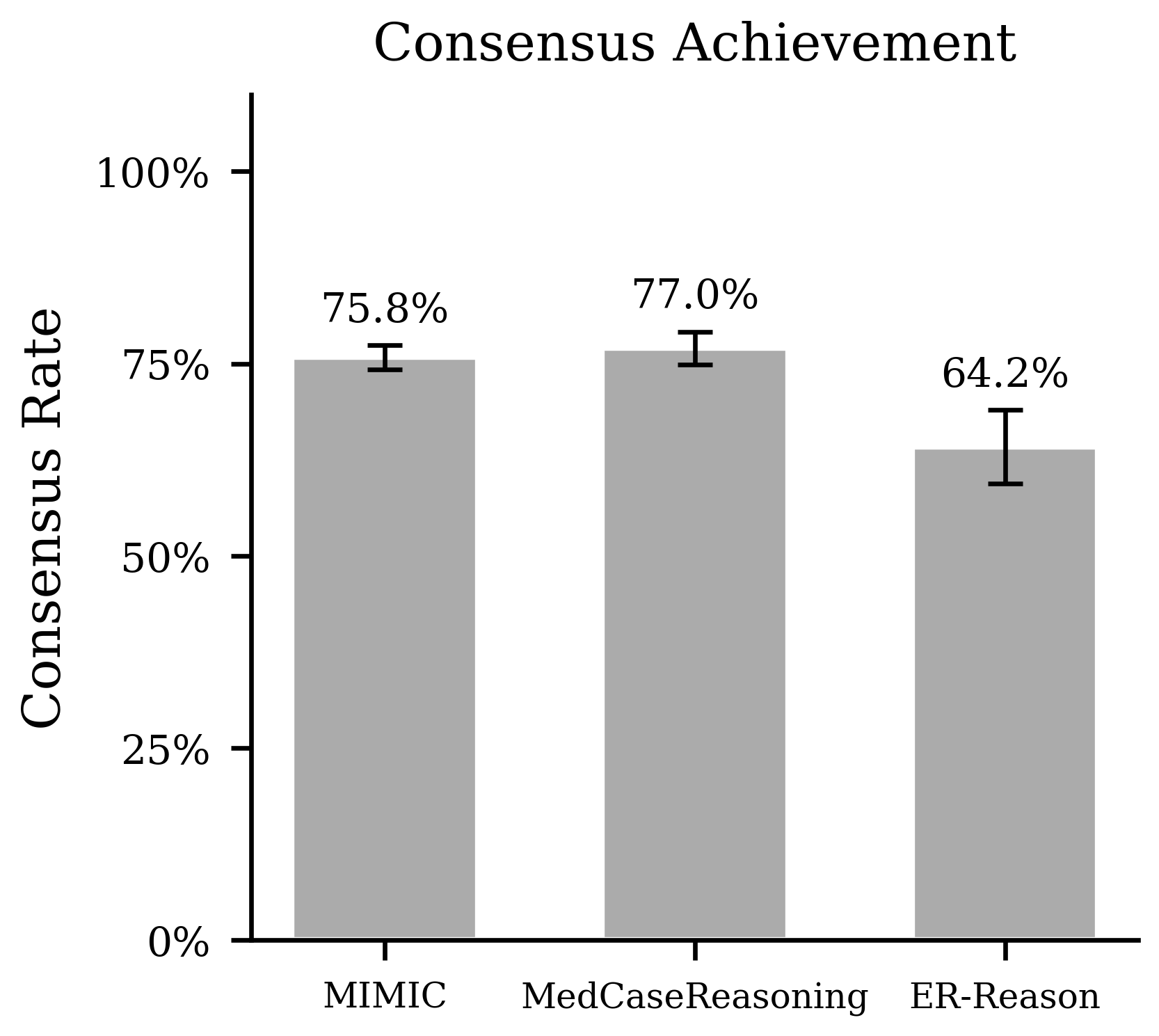}
    \caption{}
    \label{fig:llama-consensus}
  \end{subfigure}
  \hfill
  \begin{subfigure}[t]{0.23\textwidth}
    \centering
    \includegraphics[width=\linewidth]{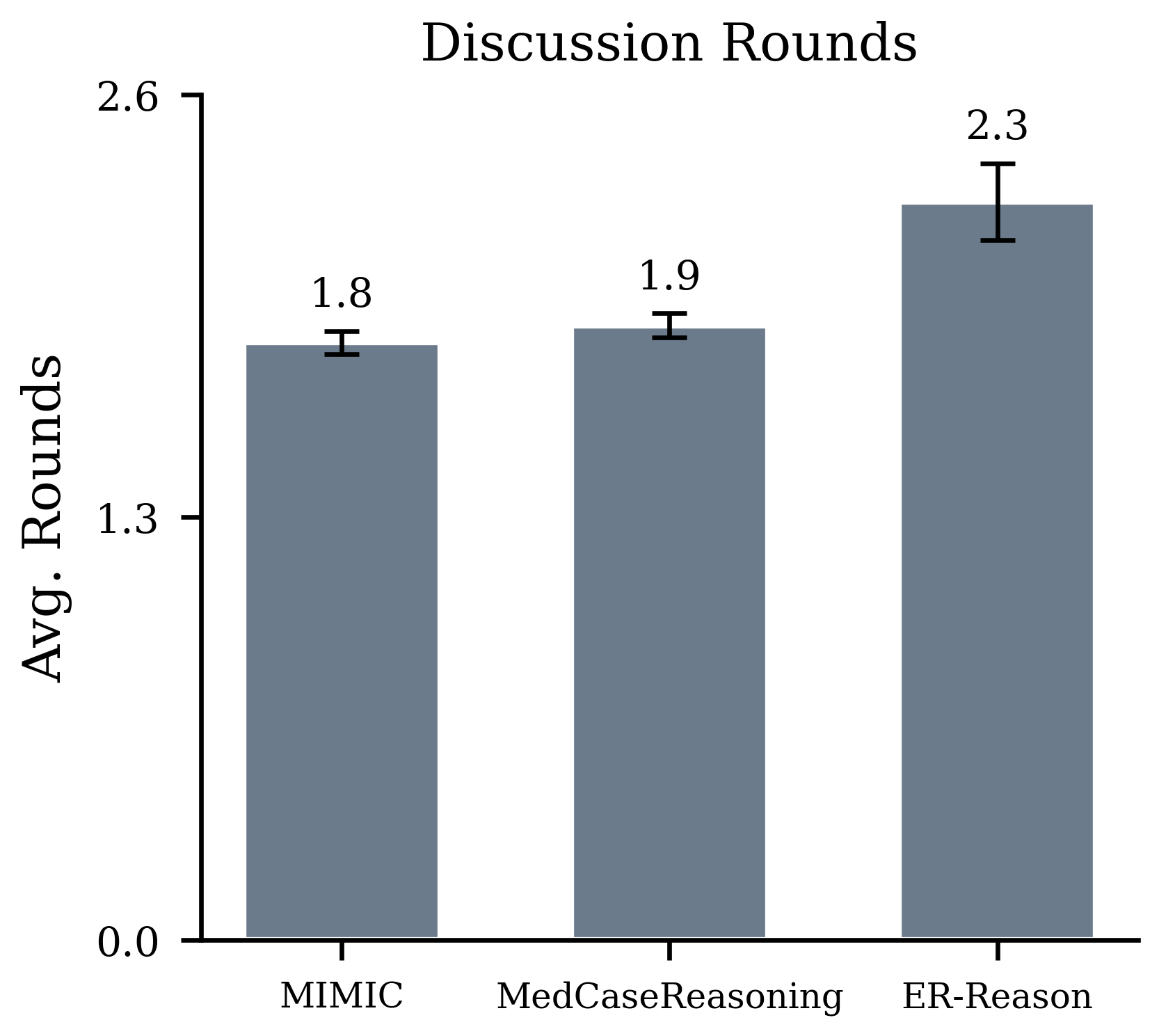}
    \caption{}
    \label{fig:llama-rounds}
  \end{subfigure}
  \hfill
  \begin{subfigure}[t]{0.23\textwidth}
    \centering
    \includegraphics[width=\linewidth]{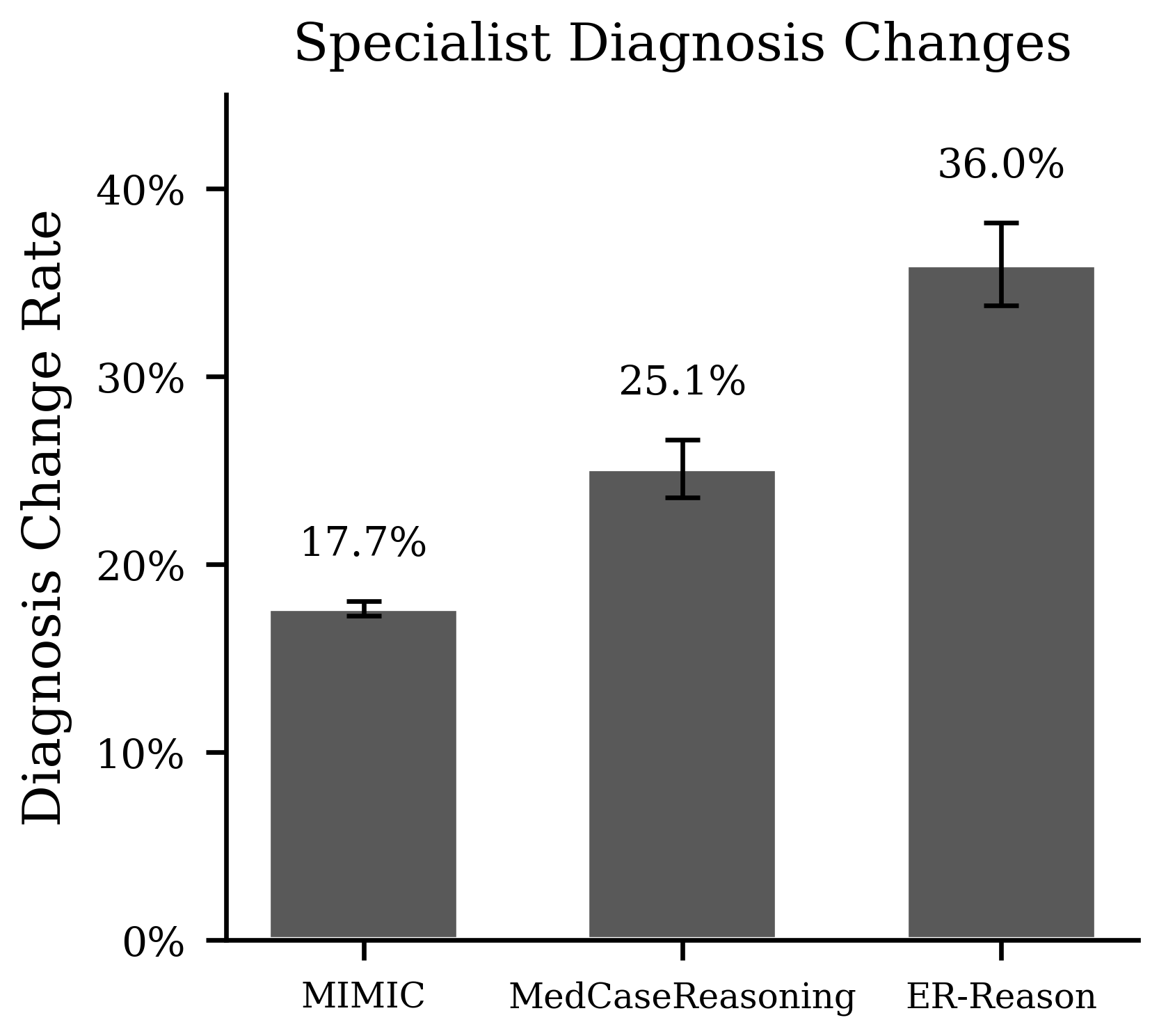}
    \caption{}
    \label{fig:llama-stance-change}
  \end{subfigure}
  \hfill
  \begin{subfigure}[t]{0.28\textwidth}
    \centering
    \includegraphics[width=\linewidth]{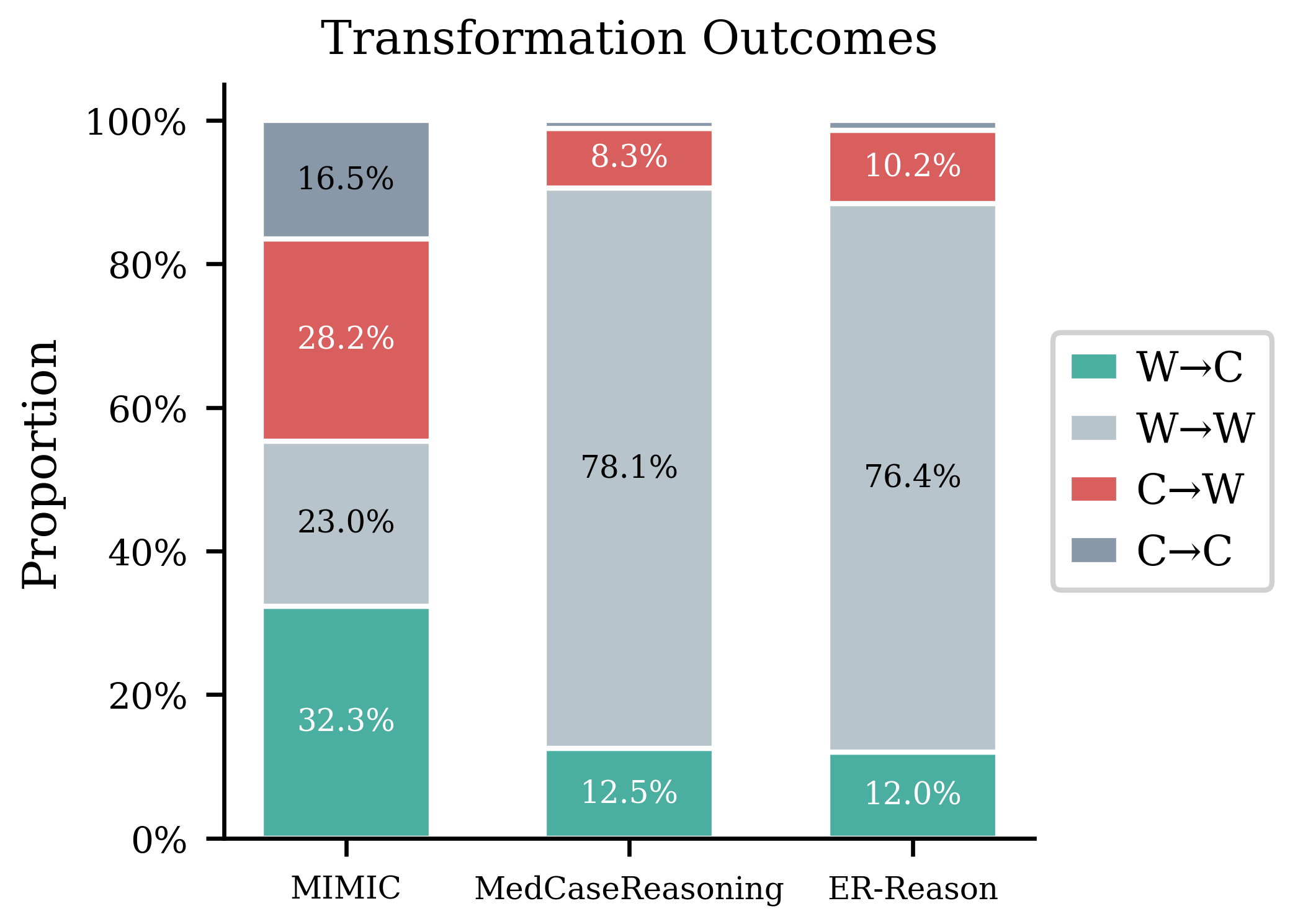}
    \caption{}
    \label{fig:llama-transofmration}
  \end{subfigure}

  \vspace{0.75em}

  \begin{subfigure}[t]{\textwidth}
    \centering
    \begin{subfigure}[t]{0.32\textwidth}
      \centering
      \includegraphics[width=\linewidth]{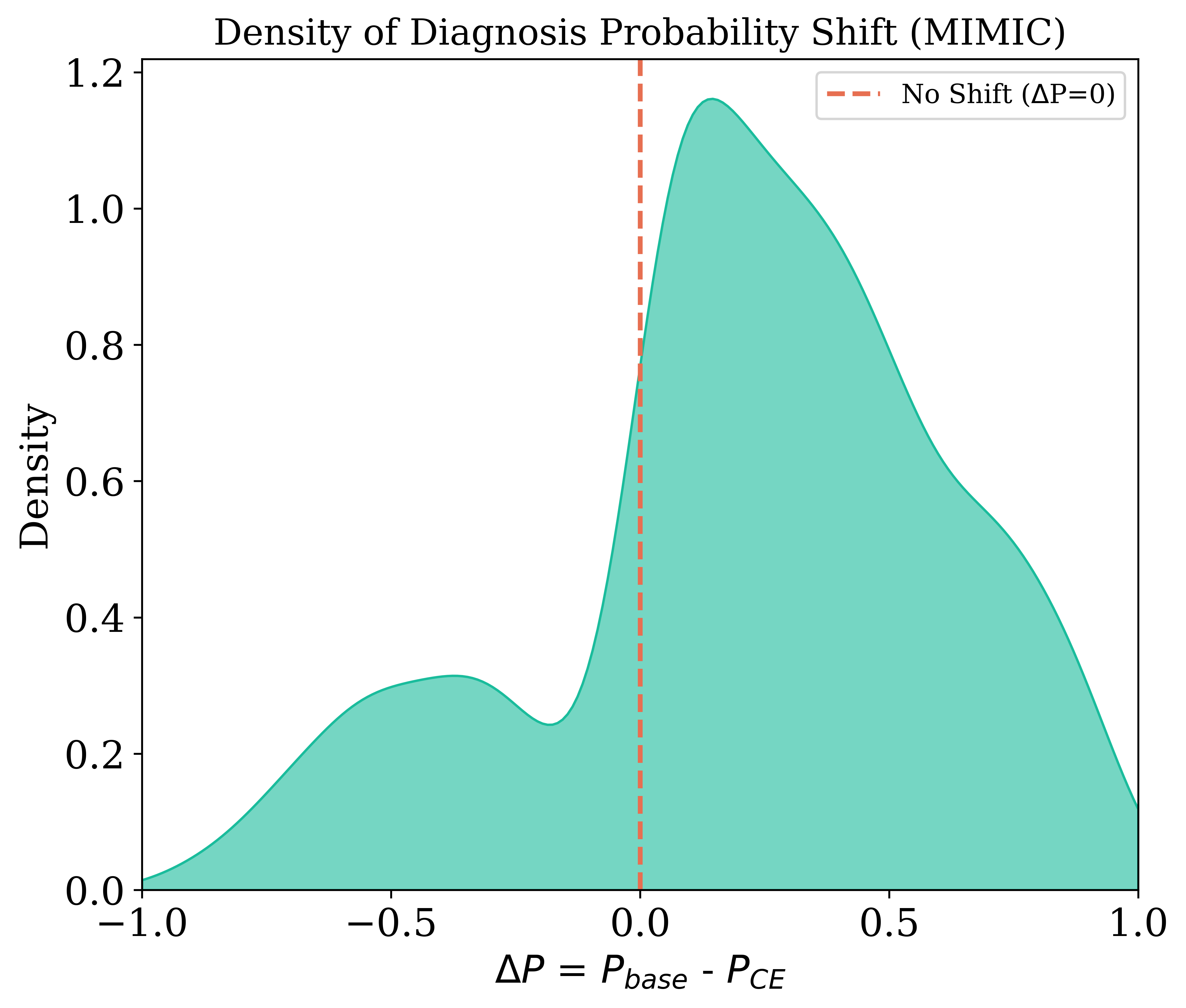}
    \end{subfigure}
    \hfill
    \begin{subfigure}[t]{0.32\textwidth}
      \centering
      \includegraphics[width=\linewidth]{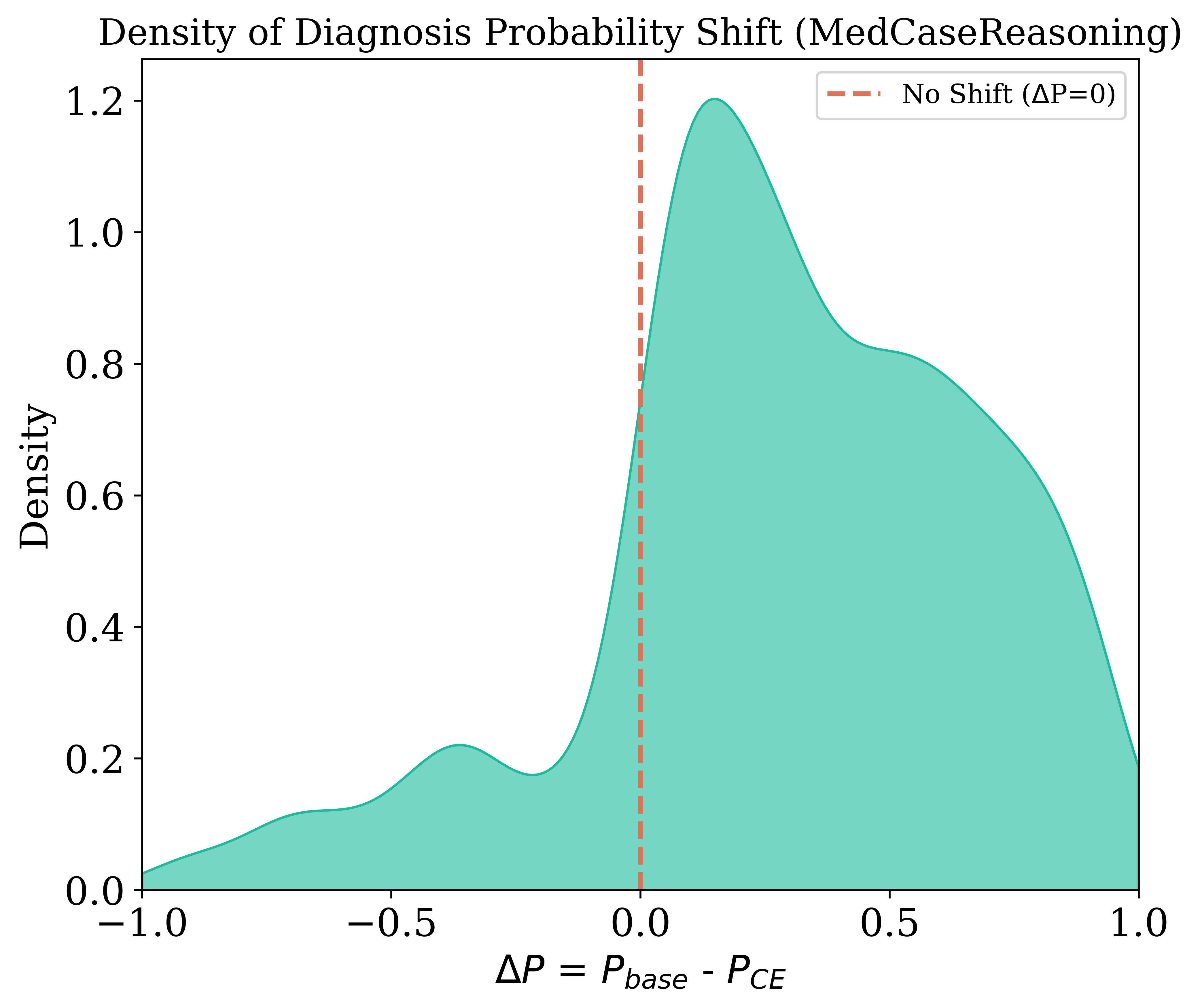}
    \end{subfigure}
    \hfill
    \begin{subfigure}[t]{0.32\textwidth}
      \centering
      \includegraphics[width=\linewidth]{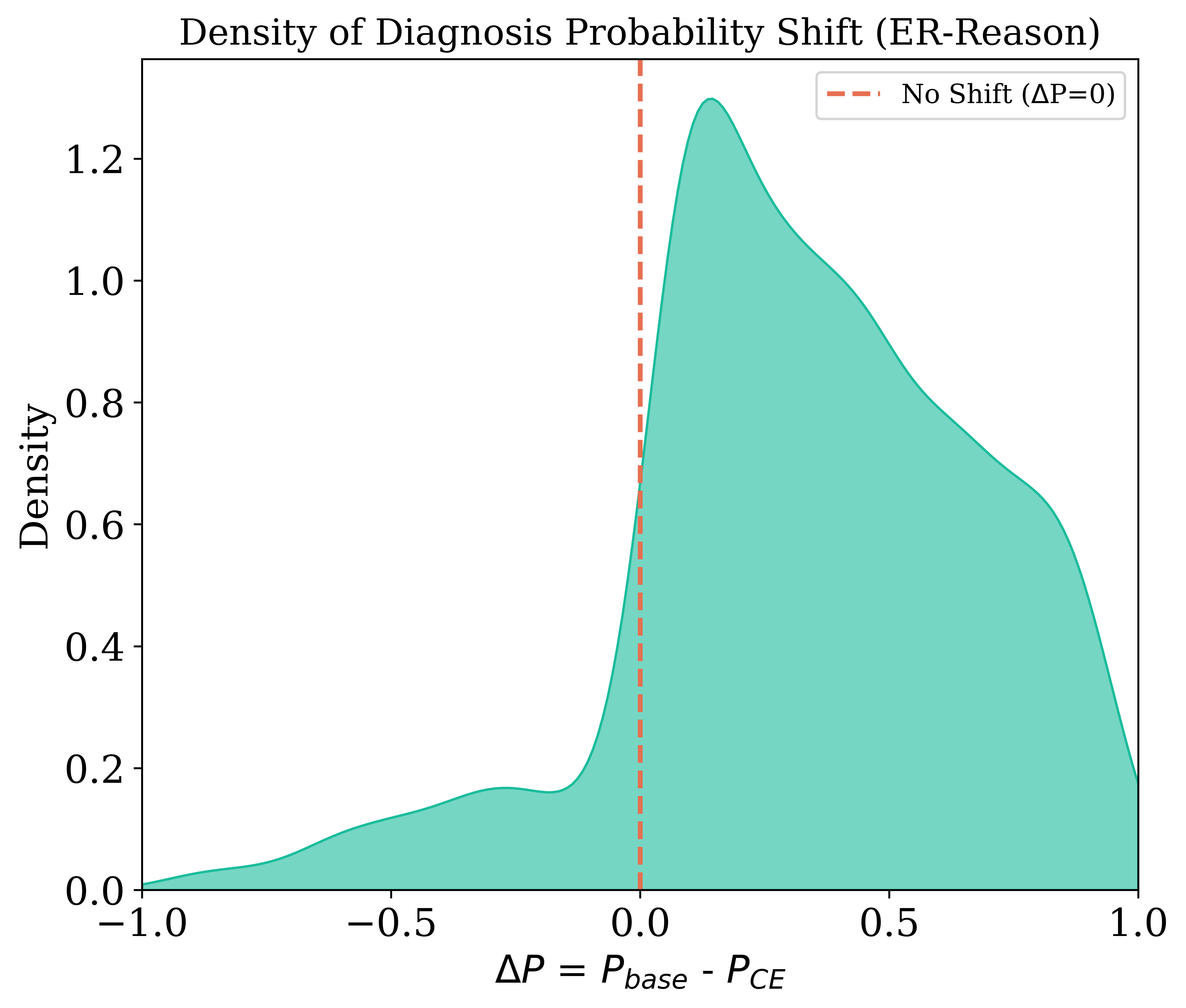}
    \end{subfigure}
    \caption{}
    \label{fig:llama-density}
  \end{subfigure}

  \vspace{0.75em}

  \begin{subfigure}[t]{\textwidth}
    \centering
    \begin{subfigure}[t]{0.32\textwidth}
      \centering
      \includegraphics[width=\linewidth]{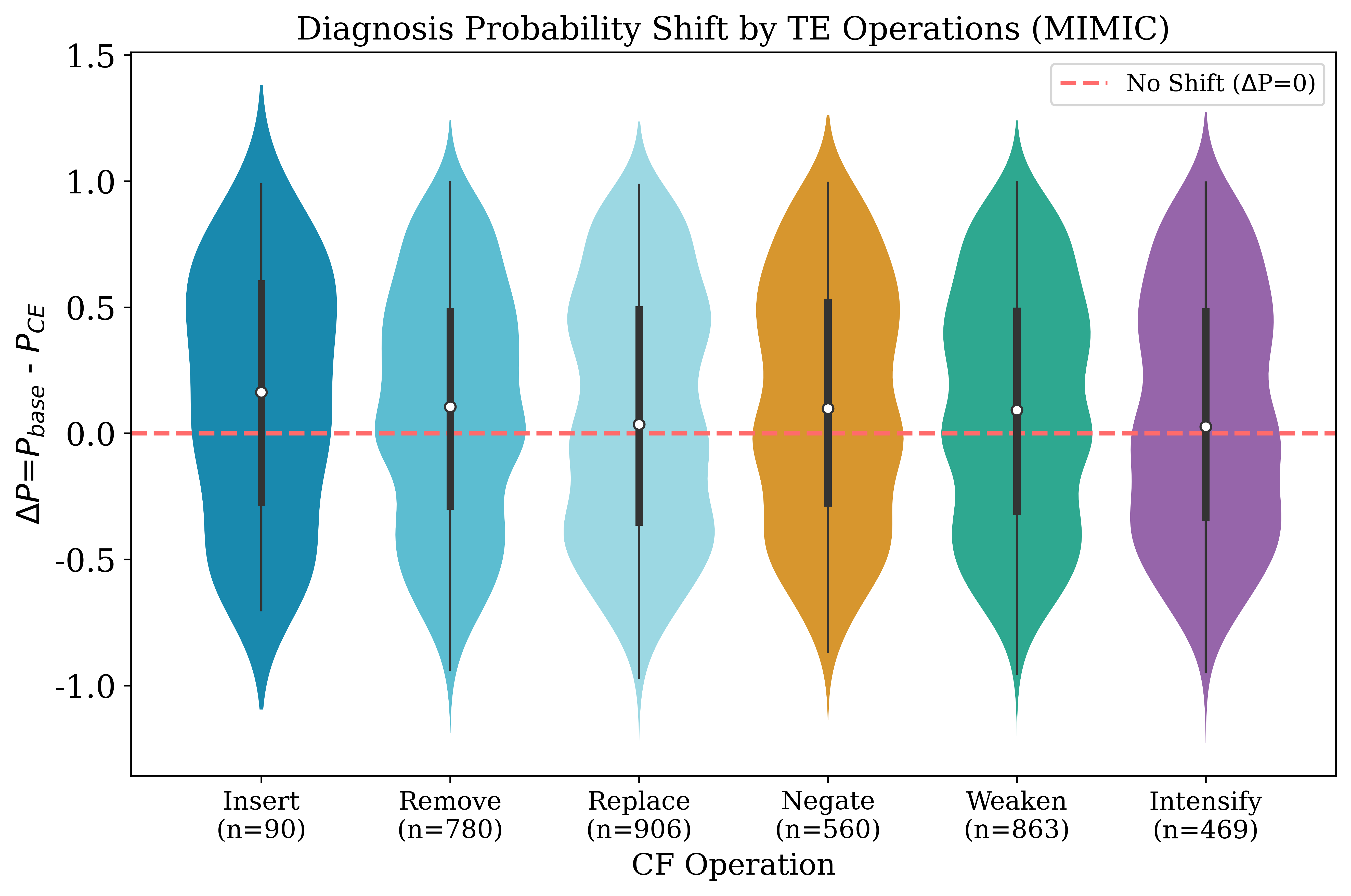}
    \end{subfigure}
    \hfill
    \begin{subfigure}[t]{0.32\textwidth}
      \centering
      \includegraphics[width=\linewidth]{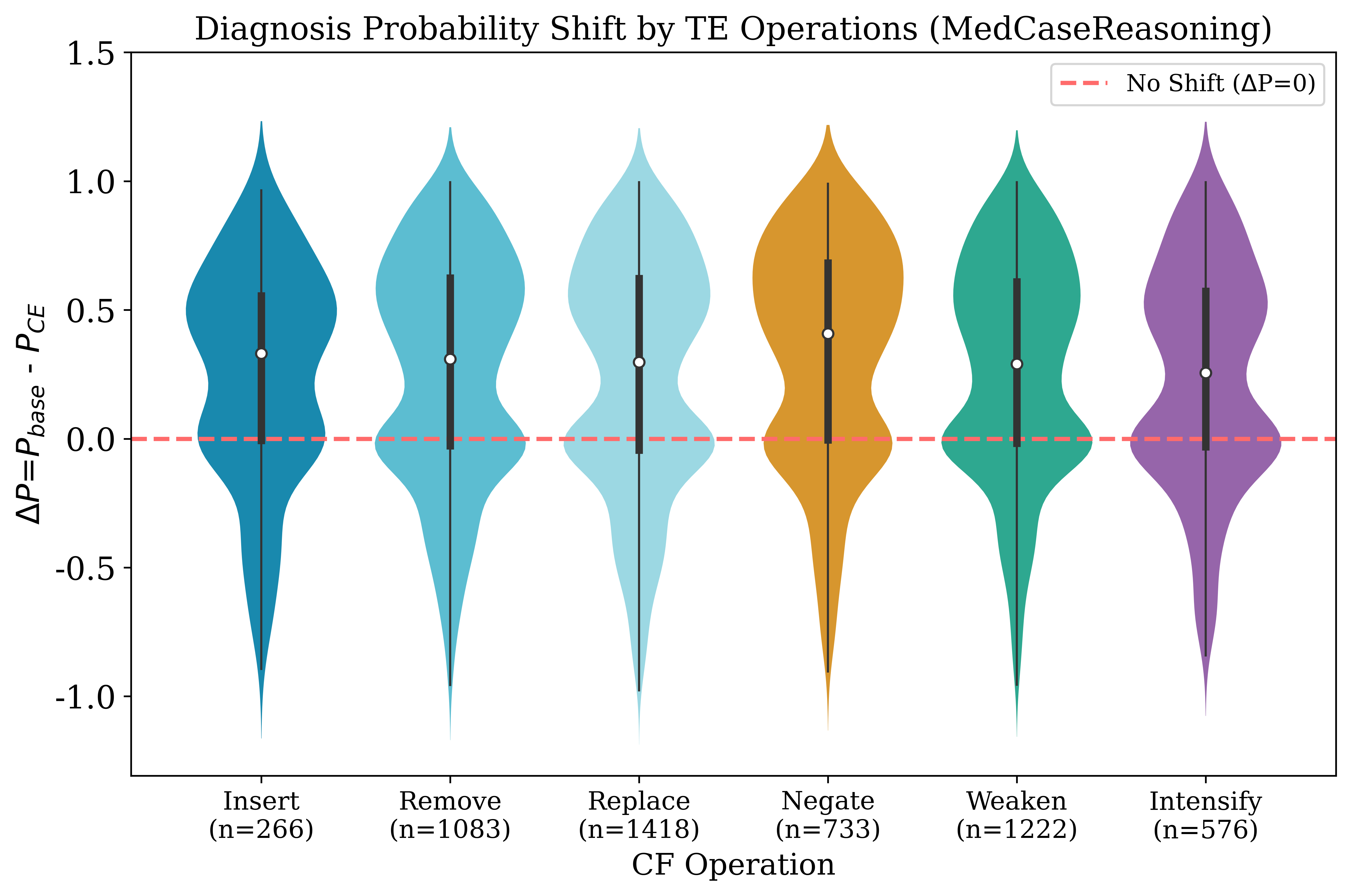}
    \end{subfigure}
    \hfill
    \begin{subfigure}[t]{0.32\textwidth}
      \centering
      \includegraphics[width=\linewidth]{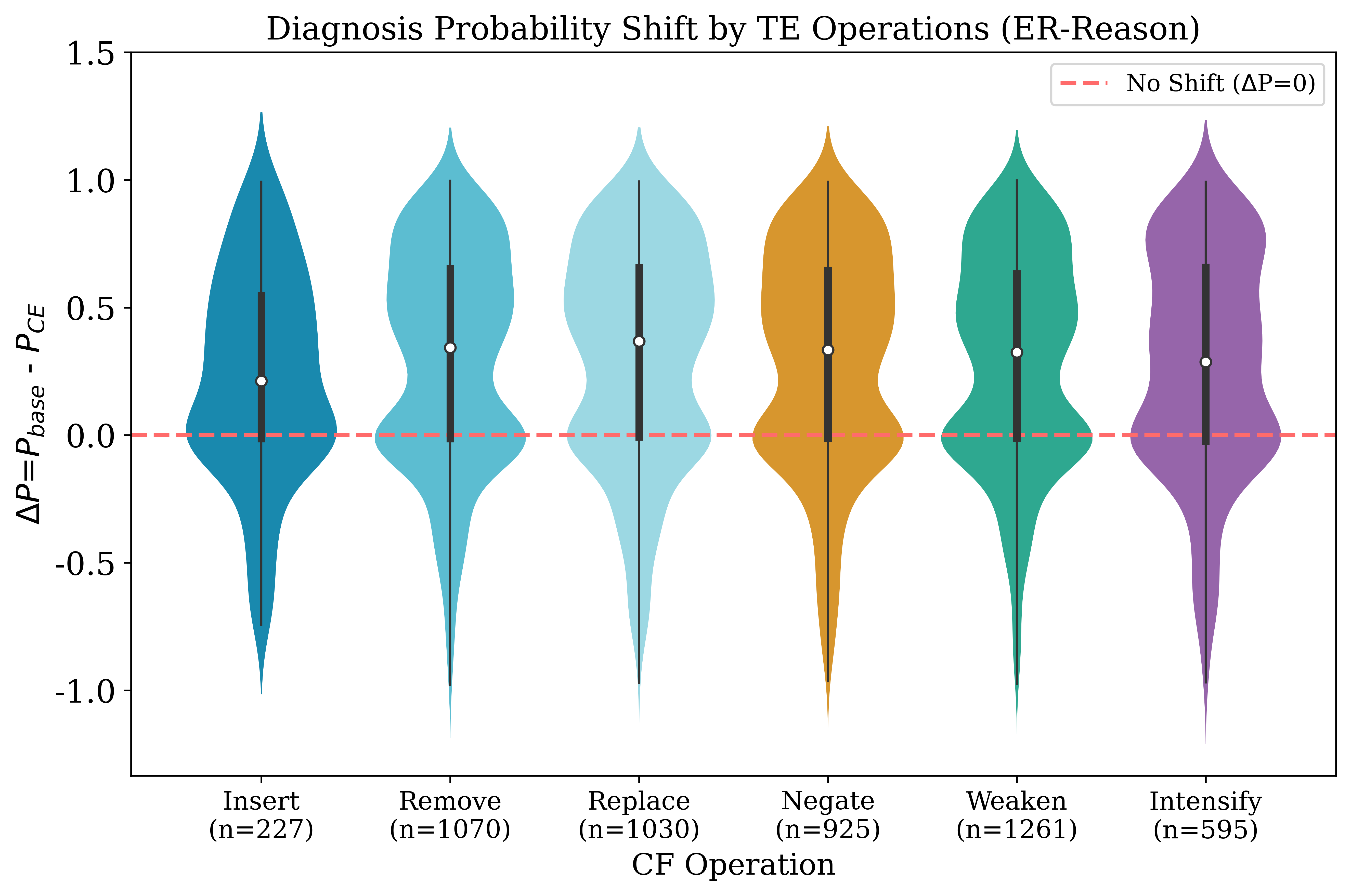}
    \end{subfigure}
    \caption{}
    \label{fig:llama-violin}
  \end{subfigure}
  \caption{Multi-round discussion statistics and the impact of counterfactual case editing on diagnostic confidence using \llama in the multi-round discussion stage. \textbf{(a)}, Consensus rate achieved by the multi-round discussion across datasets. \textbf{(b)}, Average number of the discussion rounds required per case. \textbf{(c)}, Specialist diagnosis-change rate across the three datasets. Error bars indicate the standard deviation across three random seeds. Bar graphs indicate the standard deviation of averaged results across three random seeds. \textbf{(d)}, Outcomes of diagnosis transformations, categorized by the correctness of the initial and final diagnoses relative to the gold standard (W: wrong, C: correct). \textbf{(e)}, Probability density of the diagnostic hypothesis before ($P_{base}$) and after ($P_{CE}$) targeted evidence perturbation during counterfactual case editing. Only cases where the predicted diagnosis remains unchanged before and after CF editing are included. \textbf{(f)}, Distribution of diagnosis probability shifts ($\Delta P$) across different target evidence operations on the three datasets. TE: target evidence. All statistics are calculated by averaging the results of three random seeds.}
  \label{fig:llama-CF-stats}
\end{figure*}

\subsection{Multi-round discussion statistics and the effectiveness of CF case editing} \label{sec:debate-stats}

To better understand the behavior of our proposed multi-agent diagnostic system, we analyze several discussion-related statistics using the \texttt{Llama} model. Specifically, we examine discussion consensus rates, average discussion rounds, specialist diagnosis-change rates, and diagnosis transformation outcomes (Figures~\ref{fig:llama-consensus}--\ref{fig:llama-transofmration}). In addition, to evaluate the effectiveness of our evidence extraction for counterfactual (CF) case editing, we analyze the density of diagnosis probability shifts before and after CF case editing (Figure~\ref{fig:llama-density}). As introduced in Section~\ref{sec:overview}, the multi-round discussion process involves multiple specialists discussing a case until they reach a consensus diagnosis or the maximum round of discussion is reached. On average, consensus is achieved in 75.8\% of cases in MIMIC, 77.0\% in MedCaseReasoning, and 64.2\% in ER-Reason. The ER-Reason dataset requires more discussion rounds (2.3 rounds on average) compared with the other datasets (Figure~\ref{fig:llama-rounds}), indicating that these cases are more challenging and require more extensive reasoning. 

We also analyze how often specialists change their diagnosis during the discussion. In the MIMIC dataset, specialists rarely change their initial diagnosis, suggesting that these cases are relatively straightforward. In contrast, ER-Reason shows the highest rate of diagnosis changes (Figure~\ref{fig:llama-stance-change}), reflecting the greater clinical complexity of rare disease cases. To further examine how these diagnosis changes affect diagnostic outcomes, we compare specialists' predicted diagnoses at the beginning and the end of the discussion. As shown in Figure~\ref{fig:llama-transofmration}, specialists correct more incorrect diagnoses than they convert correct diagnoses into incorrect ones. Specifically, incorrect diagnoses are corrected in 32.3\% of cases in MIMIC, 12.5\% in MedCaseReasoning, and 12.0\% in ER-Reason. These findings indicate that the discussion process helps specialists refine their reasoning and correct initial diagnostic errors, which contributes to improved overall diagnostic accuracy. For instance, in the case shown in Figure~\ref{fig:case-study}, the pulmonologist initially diagnosed ``acute respiratory distress syndrome (ARDS)'' based on prominent respiratory findings. Cardiac evaluation showing concentric left ventricular hypertrophy with preserved contractility and no pulmonary hypertension further supports a non-cardiogenic cause of pulmonary edema. After multi-round discussion with shared counterfactual evidence, however, the pulmonologist revises the diagnosis to the correct diagnosis of ``neurogenic pulmonary edema (NPE).'' This change reflects an effective refinement in reasoning: the agent changes from viewing respiratory findings as evidence of a primary lung disease to interpreting them as downstream consequences of the patient's central nervous system injury. The neurosurgeon and independent clinician further support this shift by emphasizing the patient's hemorrhage history and symptom timeline. Although disagreement remains during discussion, the final judge selects NPE because it provides a more complete clinical explanation of the pulmonary findings and is better supported by the counterfactual evidence. This case illustrates how our system can revise plausible but incomplete clinical internal reasoning through explicit evidence verification and multi-round discussion.

During CF case editing, we first extract a target evidence group from the case based on the predicted diagnosis, then apply target evidence operations to generate an edited version of the case. To evaluate the effectiveness of the extracted evidence, we measure the change in diagnosis probability before and after CF case editing for all specialists during the discussion. Because the predicted diagnosis before and after editing may differ, we focus on cases where the predicted diagnosis remains the same and analyze the corresponding probability shifts. As shown in Figure~\ref{fig:llama-density}, the density distributions of diagnosis probability shifts ($\Delta P = P_{base} - P_{CE}$) across all three datasets provide strong evidence for the effectiveness of our evidence extraction strategy. Across datasets, most probability shifts are positive, indicating that the diagnosis probability decreases after counterfactual editing. This occurs in 78.8\% of cases in MIMIC, 85.3\% in MedCaseReasoning, and 89.5\% in ER-Reason. These results suggest that modifying the extracted evidence reduces the model's confidence in the original diagnosis, demonstrating that the extracted evidence is critical for the diagnostic decision. In contrast, when $\Delta P < 0$, the model becomes more confident in the diagnosis even after the case is edited, indicating potential uncertainty in the original reasoning.

Finally, we analyze which target evidence operations produce the largest probability shifts. Figure~\ref{fig:llama-violin} summarizes the statistical distribution of probability shifts for each editing target evidence operation across the three datasets. In MedCaseReasoning, the \textit{Negate} operation produces the largest disruption to the diagnostic probability, with the highest median shift ($\Delta P = 0.408$), whereas the \textit{Intensity} operation shows the smallest impact ($\Delta P = 0.256$). In ER-Reason, the \textit{Replace} operation leads to the strongest disruption ($\Delta P = 0.3675$), while operations that add new information, such as \textit{Intensify} ($\Delta P = 0.287$) and \textit{Insert} ($\Delta P = 0.212$), produce smaller effects. These findings suggest that explicitly negating or replacing existing clinical findings is particularly effective for challenging diagnostic tasks involving rare diseases.

For the MIMIC dataset, probability shifts are generally smaller across all target evidence operations, indicating that the models are more confident in their predictions for these cases. Interestingly, introducing new clinical variables through the \textit{Insert} operation produces the largest median probability shift ($\Delta P = 0.163$), despite being the least frequently applied intervention ($n = 90$). Other target evidence operations show only minor effects on diagnostic probability. This pattern suggests that, for MIMIC cases, the agents are strongly anchored to the original case presentation, and introducing new information is more effective than modifying existing findings when influencing diagnostic predictions. We report the overall top-10 most frequently assigned specialist roles across the three datasets on \texttt{Llama} in Figure~\ref{fig:slm-specialist-stats}. See Appendix~\ref{appx:llm-debate-analysis} for multi-round discussion analysis of \texttt{Deepseek}.

\subsection{Ablation Study} \label{sec:ablation-study}
\begin{figure*}[!ht]
  \centering
    \begin{subfigure}[t]{\textwidth}
    \centering
    \includegraphics[width=0.6\linewidth,height=6.5cm,keepaspectratio]{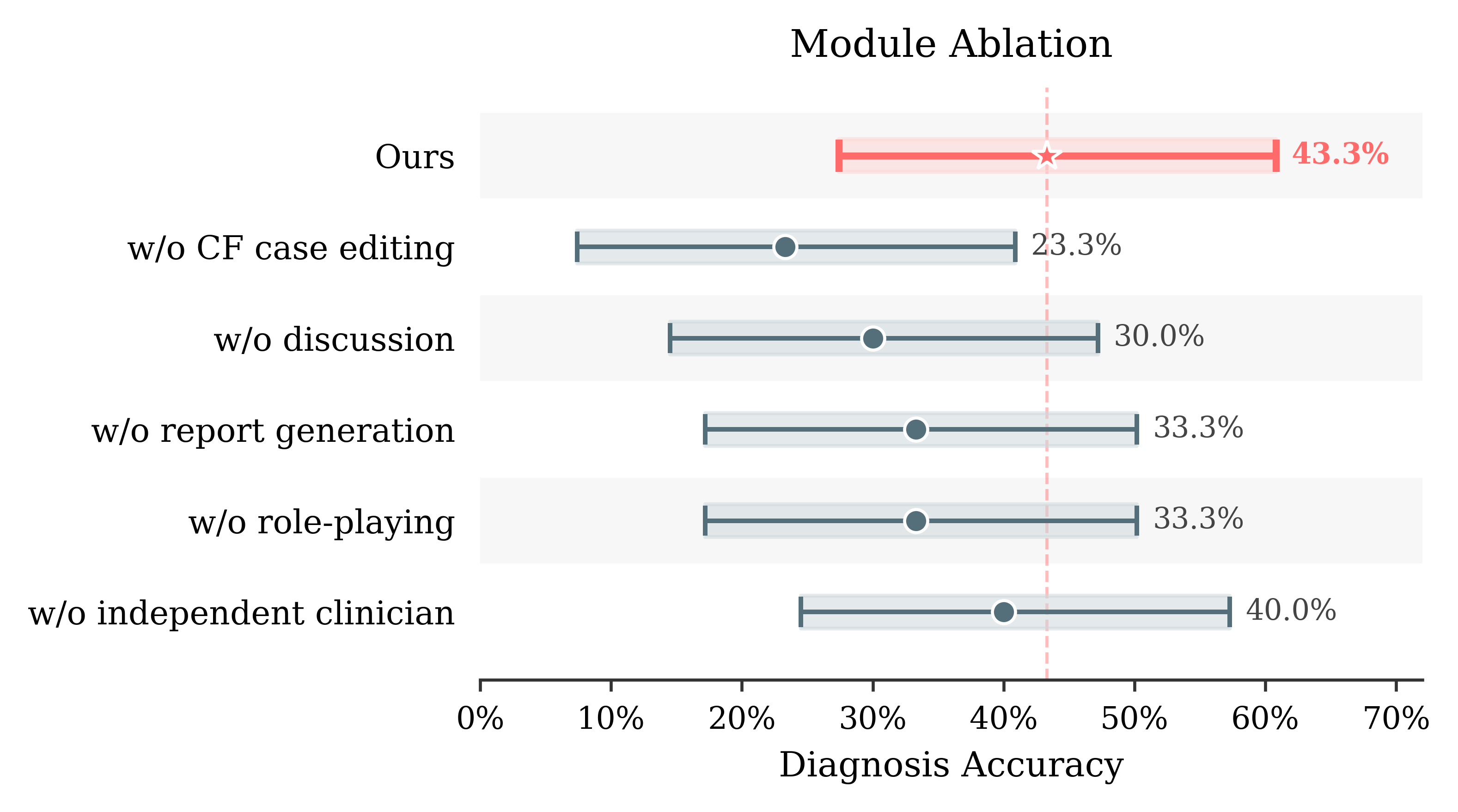}
    \caption{}
    \label{fig:module-ablation}
  \end{subfigure}
  \hfill
   \begin{subfigure}[t]{\textwidth}
    \centering
    \includegraphics[width=\linewidth,height=10.5cm,keepaspectratio]{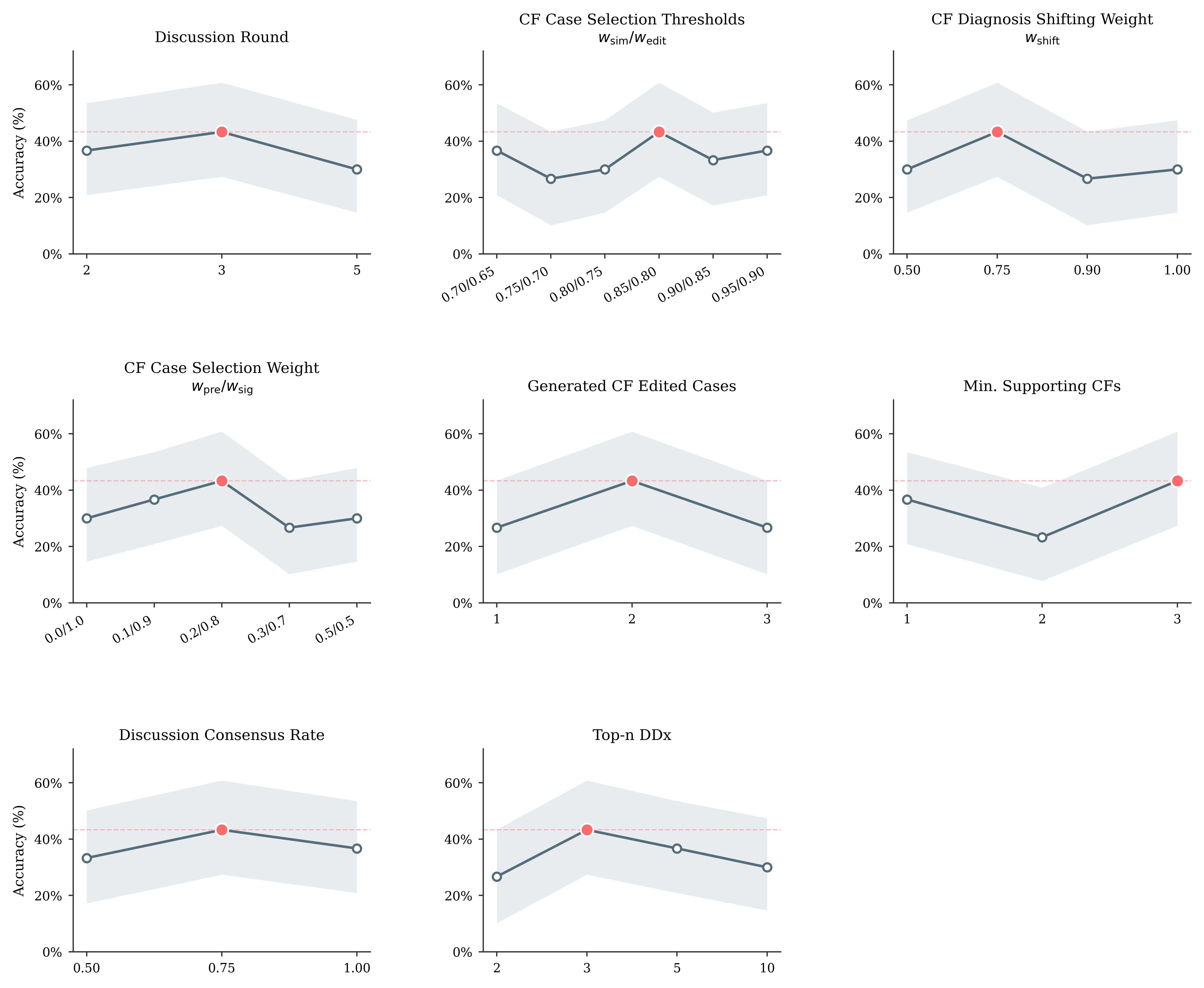}
    \caption{}
    \label{fig:hyperparameter-ablation}
  \end{subfigure}
  \hfill
\caption{Ablation study of \llama \space over different functional modules on MedCaseReasoning. The shaded area represents the 95\% CI. \textbf{(a)} Diagnostic performance with various functional moduels added in our multi-agent diagnostic system. w/o: without; CF: counterfactual. \textbf{(b)} Diagnostic performance with various hyperparameters. DDx: differential diagnosis.}
  \label{fig:ablation-study}
\end{figure*}

We conduct ablation studies by randomly selecting 30 cases from the test set of Medcasereasoning \cite{wu2025medcasereasoning} given the computation limits. The ablation study demonstrates that each functional module in the multi-agent diagnostic system contributes meaningfully to diagnostic performance. As shown in Figure~\ref{fig:module-ablation}, removing CF case editing causes the largest drop in diagnostic accuracy (from 43.3\% to 23.3\%), identifying it as the most critical component. Removing multi-round discussion, specialist report generation, role-playing, and the independent clinician module each also reduce performance, showing that our full multi-agent diagnostic pipeline is necessary for optimal results. For hyperparameter sensitivity (Figure~\ref{fig:hyperparameter-ablation}), most parameters show a clear peak at the selected settings, with performance declining at both lower and higher values, showing the chosen configuration is well-tuned. Notably, accuracy is relatively stable across a moderate range for several parameters, indicating the system is not overly sensitive to small hyperparameter changes. Overall, these results support the robustness and design validity of the proposed system, where collaborative reasoning and CF case editing are vital for better diagnostic performance.

\subsection{Human Evaluation} \label{sec:human-eval}

\begin{figure*}[!ht]
    \begin{subfigure}[t]{\textwidth}
  \begin{subfigure}[t]{0.24\textwidth}
    \centering
    \includegraphics[width=\linewidth]{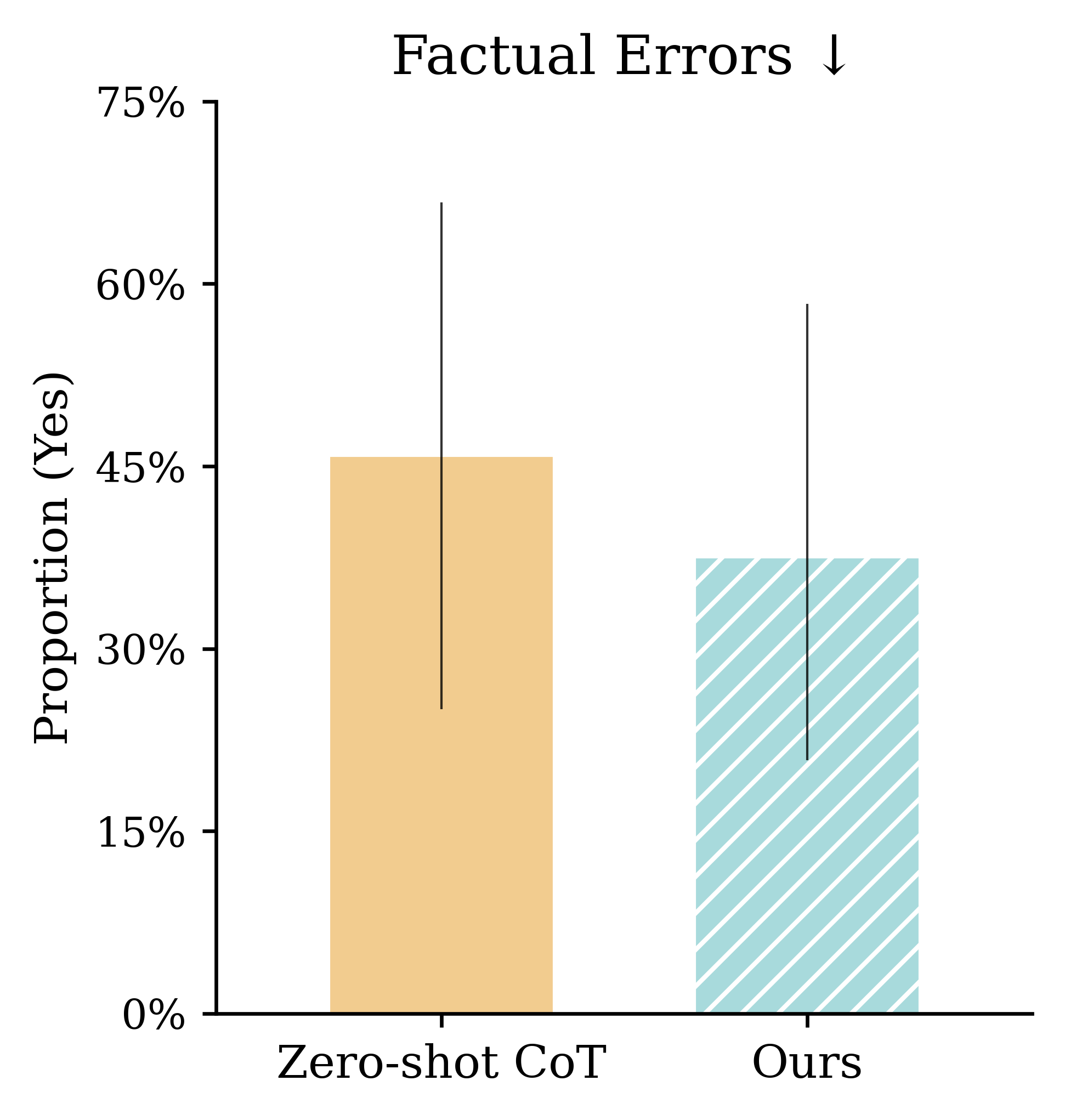}
  \end{subfigure}
  \hfill
  \begin{subfigure}[t]{0.24\textwidth}
    \centering
    \includegraphics[width=\linewidth]{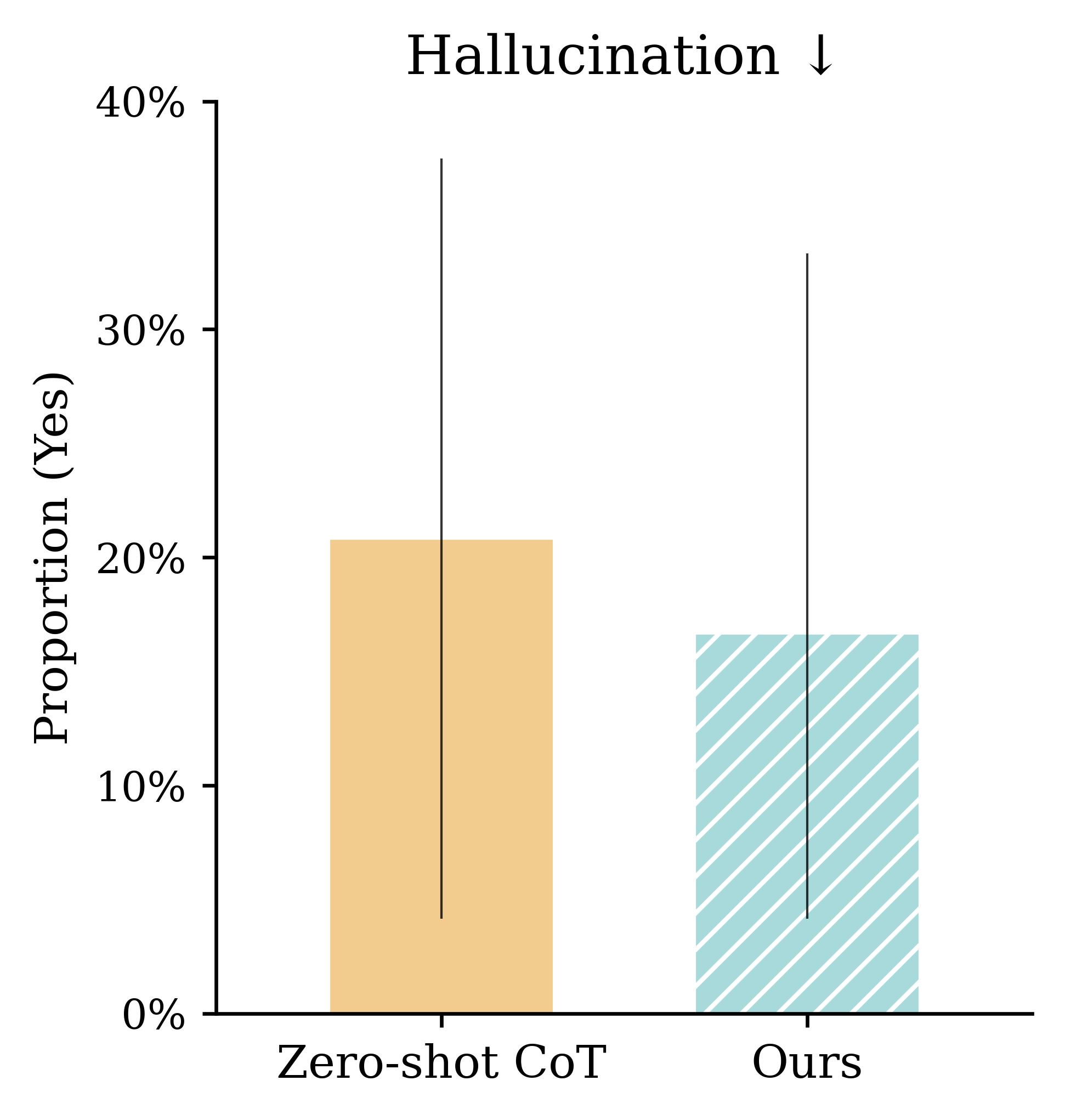}
  \end{subfigure}
  \hfill
  \begin{subfigure}[t]{0.24\textwidth}
    \centering
    \includegraphics[width=\linewidth]{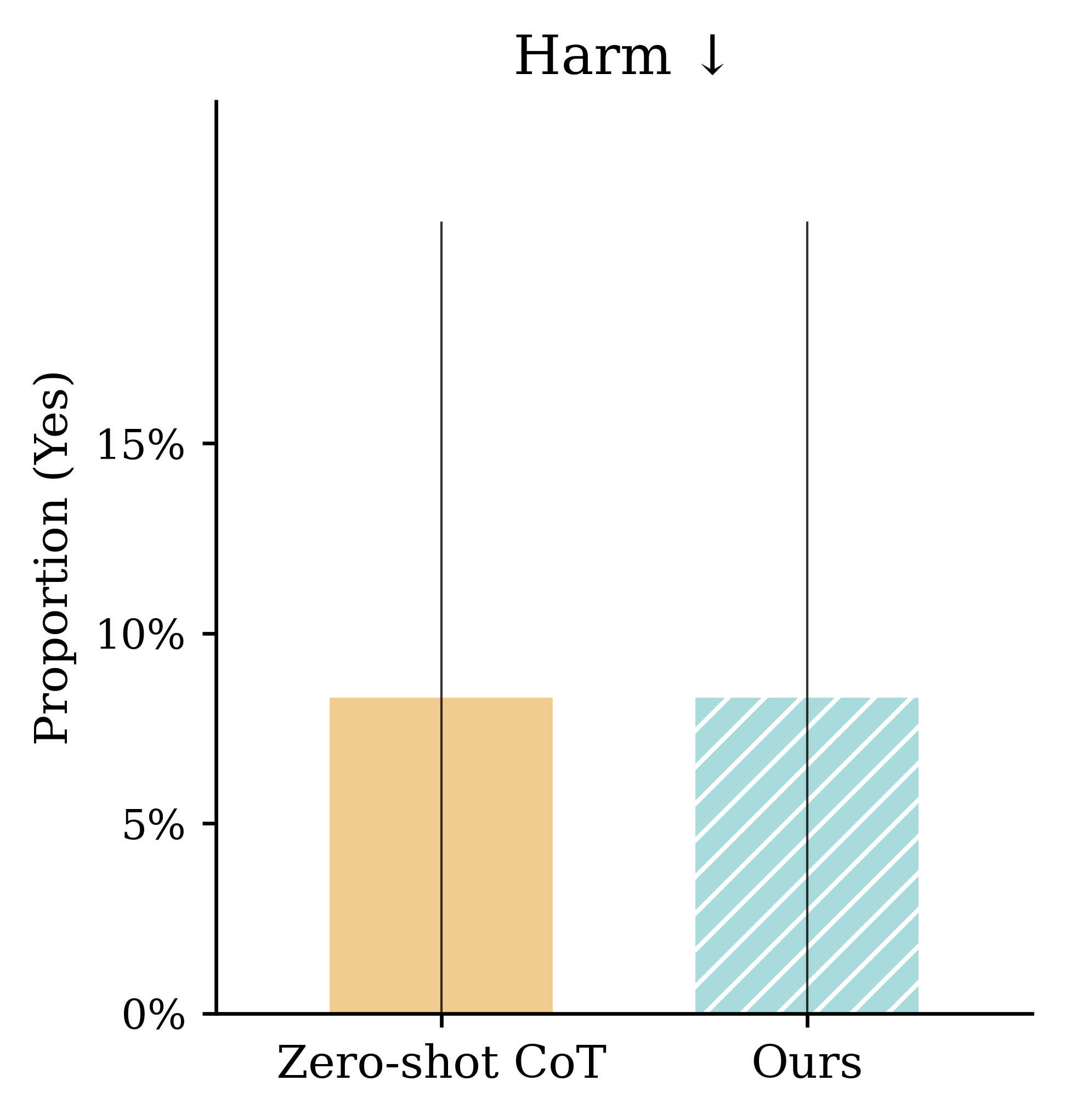}
  \end{subfigure}
  \hfill
  \begin{subfigure}[t]{0.24\textwidth}
    \centering
    \includegraphics[width=\linewidth]{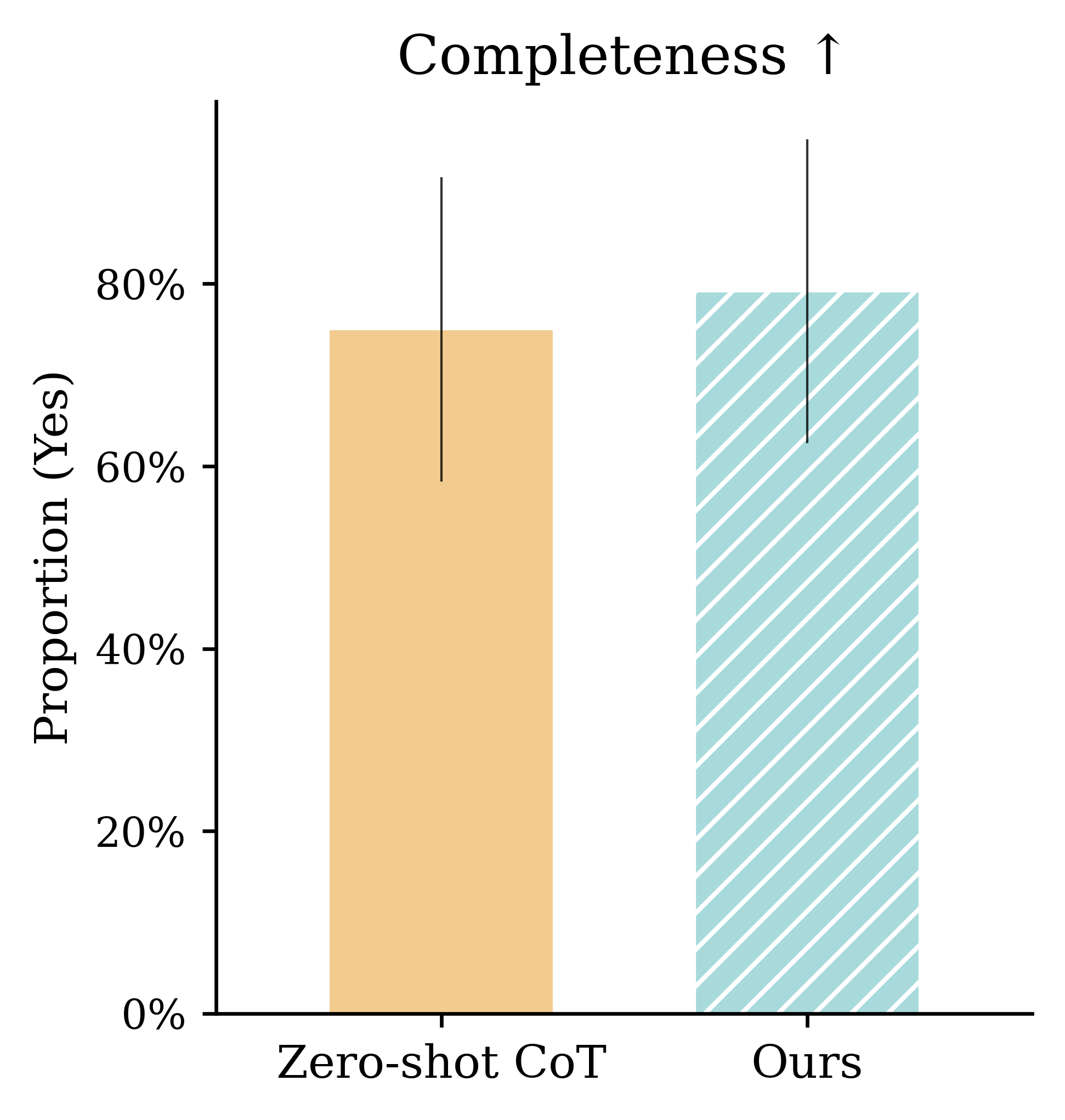}
    \end{subfigure}
  \caption{}
  \label{fig:binary}
  \end{subfigure}

  \vspace{0.75em}

  \begin{subfigure}[t]{\textwidth}
    \centering
      \includegraphics[width=0.8\linewidth]{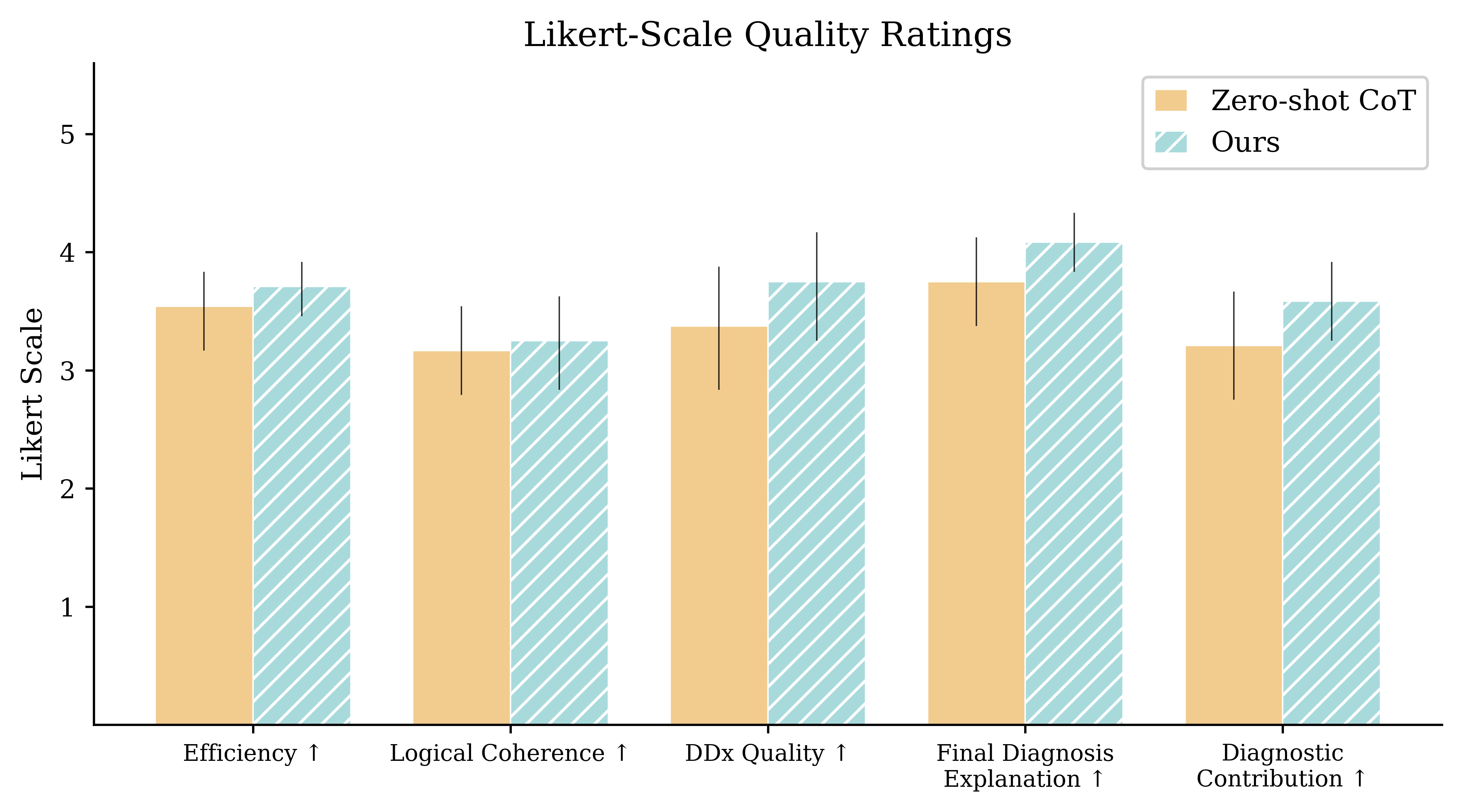}
    \caption{}
    \label{fig:likert}
  \end{subfigure}

  \vspace{0.75em}

  \begin{subfigure}[t]{\textwidth}
    \centering
      \includegraphics[width=0.82\linewidth]{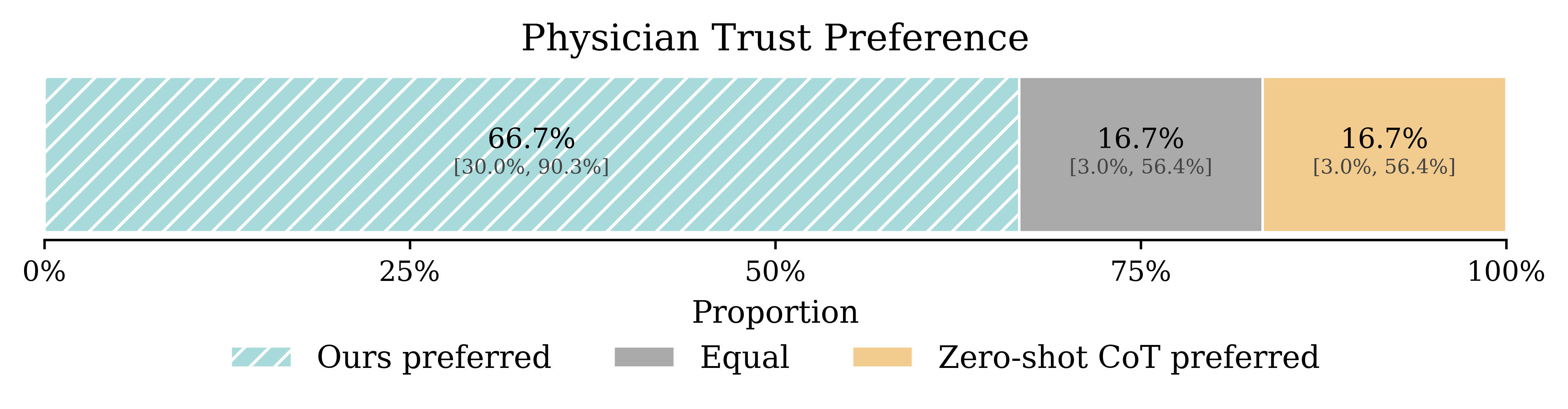}
      \caption{}
    \label{fig:trust}
  \end{subfigure}
  \caption{Human evaluation of clinical reasoning quality comparing our method with zero-shot CoT across three datasets. \textbf{(a)}, Error and Safety Assessment (ESA) metrics annotated with binary yes/no labels. \textbf{(b)}, Reasoning Quality Assessment (RQA) and Clinical Contribution (CC) evaluated on a 1–5 Likert scale (presented in Figure~\ref{fig:interface4} and Figure~\ref{fig:interface5}). \textbf{(c)}, Overall trust preference between the two methods. Bar graphs indicate the mean ± 95\% CIs.}
  \label{fig:human-eval}
\end{figure*}

Evaluating model-generated clinical reasoning traces requires moving beyond simple accuracy metrics toward human assessment of the logic, safety, and justification underlying a diagnosis. For example, an LLM may produce the correct diagnosis for the wrong reasons, such as relying on spurious correlations. Previous studies mainly focus on evaluating the correctness of model-predicted diagnoses \cite{chen2025enhancing} or overall diagnostic performance \cite{liu2025generalist}, which do not fully capture the quality of the underlying reasoning process. To address this limitation, we conduct a human evaluation to assess the quality of model-generated clinical reasoning traces. Two licensed physicians (S.D. and G.E.B.) independently evaluate the reasoning traces produced by our method and the zero-shot CoT method.

Most prior evaluation protocols emphasize diagnostic accuracy or similarity to ground-truth reasoning, which may overlook the clinical quality of the diagnostic process. In our framework, we summarize the intermediate reasoning chains produced by each diagnostic agent and present a final reasoning trace for evaluation. We adopt evaluation principles inspired by prior frameworks such as Medical Decision Making textbook \cite{sox2024medical}, CLEVER (CLinical EValuation for Effective Reasoning in Diagnosis) \cite{liu2025generalist} and Revised-IDEA \cite{schaye2022development}, which are designed to assess the quality of LLM-generated clinical reasoning through expert review.

For the evaluation, we randomly select 12 pairs of reasoning traces generated by zero-shot CoT and our proposed method. Based on the results in Figure~\ref{fig:baseline-results}, we observe that \texttt{Llama} shows the largest performance improvement when comparing our method with zero-shot CoT, whereas \texttt{m1} shows relatively small improvements. Therefore, for each model, we sample two pairs of reasoning traces: one case where both methods produce the correct diagnosis and one case where both methods produce an incorrect diagnosis. Across the three datasets, we sample six pairs for each model, resulting in a total of 12 reasoning pairs for human evaluation. A detailed description of the evaluation protocol is provided in Appendix~\ref{appx:human-eval}.

As shown in Figure~\ref{fig:human-eval}, we design three evaluation dimensions to assess the quality of the reasoning traces. Two physicians independently annotate the 12 reasoning pairs. The first dimension, \textit{Error and Safety Assessment (ESA)}, examines whether the reasoning traces contain factual errors, hallucinations, potential clinical harm, and whether the reasoning process is complete. Our method shows lower error rates and higher completeness compared with zero-shot CoT. Specifically, our approach produces 37.5\% factual errors, 16.7\% hallucinations, 8.3\% critical harm, and 79.2\% completeness, outperforming the zero-shot CoT baseline (Figure~\ref{fig:binary}).

We further evaluate reasoning quality using Likert-scale ratings across two dimensions: \textit{Reasoning Quality Assessment (RQA)} and \textit{Clinical Contribution (CC)}. The RQA dimension measures the logical coherence and efficiency of the reasoning process, while the CC dimension evaluates differential diagnosis consideration, quality of final diagnosis explanation, and potential usefulness for clinicians. All criteria are rated on a 1--5 Likert scale. As shown in Figure~\ref{fig:likert}, our method consistently achieves higher ratings across all criteria compared with zero-shot CoT, indicating improved reasoning quality and stronger clinical relevance.

Additionally, we assess overall trust in the reasoning outputs. Physicians are asked to determine which reasoning trace within each pair inspires greater confidence in the diagnostic process. Our method is preferred in 66.7\% of cases, outperforming the single-prompt zero-shot CoT approach (16.7\%) (Figure~\ref{fig:trust}). Together, these results suggest that our proposed method not only improves diagnostic accuracy but also produces reasoning traces that are more reliable, coherent, and clinically useful compared with standard prompting.

Finally, as illustrated in Figure~\ref{fig:interface6}, physicians are able to highlight potential biases in the reasoning traces during the evaluation. Overall, three types of anchoring-related biases are identified in both our method and the baseline approach. For omission bias, physicians identify four instances in our method and three instances in the baseline. Our method shows one case of overinvestigation, whereas no such cases are identified in the baseline. In contrast, the baseline contains more incorrect causal attributions, with seven instances compared to four in our method. Beyond these specific bias categories, both physicians note qualitative differences in the reasoning traces. They observe that our method discusses DDx and ruling-out justifications in more detail while maintaining a more concise and logically structured explanation. One physician explicitly indicates a preference for the explanation style produced by our method. In comparison, the baseline reasoning traces are described as overly verbose and repetitive, and they contain more factual and logical errors.

\section{Discussion} \label{sec:discussion}

In this study, we develop a multi-agent diagnostic system with counterfactual case editing for complex clinical diagnosis. Accurate clinical diagnosis cannot rely on a single forward reasoning \cite{chen2025reverse} through a language model. Instead, it requires exploring differential hypotheses, testing each hypothesis against available evidence, and revising conclusions when contradictions arise. This process mirrors how experienced clinicians reason in real-world practice \cite{sox2024medical}, yet many existing LLM-based approaches do not explicitly incorporate this reasoning pattern. To improve diagnostic interpretations, prior work models diagnosis using explicit causal graphs to understand how necessary or sufficient each disease is as a cause of the patient's symptoms \cite{richens2020improving}. Although this line of work highlights the importance of structured clinical reasoning, many current LLM-based diagnostic methods fail to explicitly test competing hypotheses during inference. As a result, standard prompting, even when combined with chain-of-thought reasoning, often commits to a leading hypothesis early and rarely revisits alternative explanations \cite{wu2025medcasereasoning}. To investigate how LLMs produce internal reasoning, Chen et al. \cite{chen2025reverse} force the LLMs to produce explicit bidirectional reasoning traces and then check their mutual consistency, showing how an answer is justified from question $\rightarrow$ answer and answer $\rightarrow$ question. Existing multi-agent systems partially address this issue by introducing specialist roles and collaborative discussion. However, these approaches typically guide agents toward consensus through open-ended dialogue without providing a structured mechanism to verify whether their reasoning is grounded in clinical evidence \cite{tang2024medagents, kim2024mdagents}. More recently, AURA \cite{fathi2025aura} introduces an agent-driven precision-guided image editing of pathologies to generate counterfactual images to test hypotheses, improve visual--linguistic explanations, and perform self-evaluation of edits in tasks like chest X-ray explanation. While these advancements demonstrate the value of counterfactual editing in the visual domain, the role of counterfactual reasoning as an explicit evidence-checking mechanism for text-based clinical diagnosis remains underexplored.

Our approach addresses this gap in two ways. First, we introduce counterfactual case editing as a structured form of evidence verification. Specialist agents modify targeted clinical findings, such as negating, removing, replacing, weakening, intensifying, or inserting evidence, and then observe how the predicted diagnosis probability changes. Second, we quantify this change through the Counterfactual Probability Gap (CPG), which provides each agent with an interpretable signal for identifying the most critical clinical evidence supporting the decision. This mechanism makes the reasoning process more interpretable. Instead of producing a plausible narrative alone, the agent must demonstrate which clinical features directly influence its diagnostic conclusion. Our design also addresses several practical challenges observed in previous multi-agent approaches. Many prior systems rely on fixed agent teams, long discussion rounds, or computationally expensive LLMs \cite{chen2025enhancing, tang2024medagents, kim2024mdagents}. In contrast, our method is designed to operate with smaller open-source language models that can be deployed locally. We introduce a summarizer agent to condense discussion rounds and maintain manageable token lengths, and we replace the requirement for full consensus with a judge agent that resolves disagreements when specialists cannot converge. These design choices make the system more scalable and accessible than approaches that rely heavily on closed-source APIs. More broadly, our work contributes a simple but clinically meaningful idea: counterfactual case editing (``what if not?'') can serve as an explicit evidence-checking mechanism within language model--based diagnostic pipelines. As a result, our method provides a training-free and model-agnostic approach that can be applied to existing language models to improve reasoning reliability and diagnostic accuracy.

We evaluate six open-source LLMs and one frontier close-source LLM across three diagnostic datasets that differ in source, difficulty, and disease scope. Compared with both standard prompting methods and multi-agent baselines, our method achieves the highest average diagnostic accuracy across all LLMs (Figure~\ref{fig:baseline-results}), although the improvement over the second-best baselines is smaller for \texttt{Qwen} and \texttt{m1}. The largest improvement is observed for \texttt{Llama}, which shows a 13.2\% gain over zero-shot CoT. This result suggests that even without introducing additional medical knowledge, our method can substantially improve the clinical reasoning ability of general-purpose models. Performance on \texttt{Deepseek} and \texttt{GPT5mini} further indicates that our proposed multi-agent diagnostic framework benefits models across different scales. In addition, the module ablation study (Figure~\ref{fig:module-ablation}) shows that removing the counterfactual case editing component causes the largest accuracy drop (from 43.3\% to 23.3\%), highlighting that the counterfactual verification mechanism is the critical component of our method.

\begin{figure*}[!ht]
  \centering
  \makebox[\textwidth][c]{%
    \includegraphics[width=1.25\textwidth]{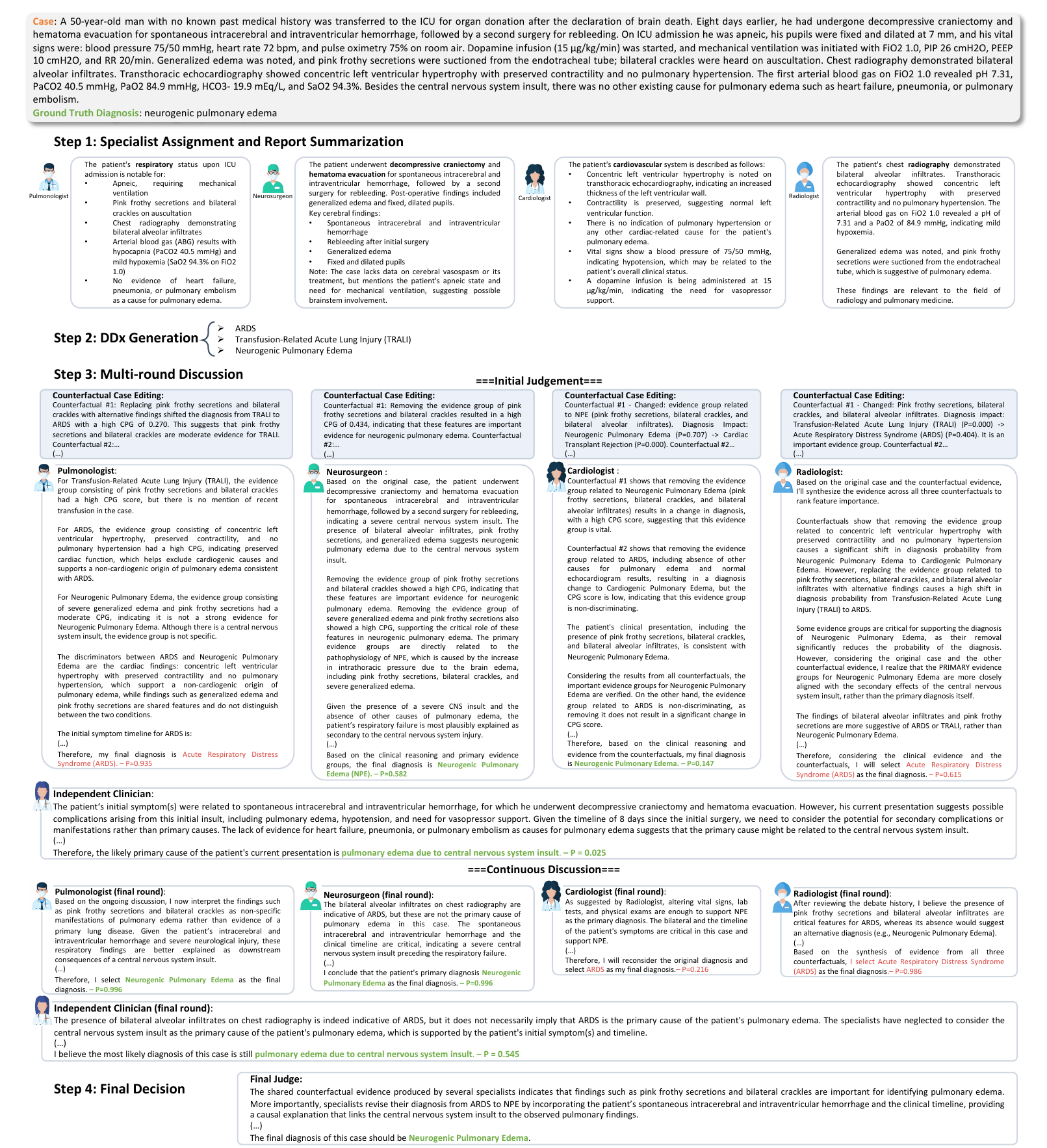}
  }
  \caption{Example of diagnostic rationales generated by our multi-agent diagnostic framework using \llama \space model on a case from the MedCaseReasoning dataset.}
  \label{fig:case-study}
\end{figure*}

To better understand why the system improves diagnostic performance, we further analyze the multi-round discussion behavior of the best-performing model, \texttt{Llama} (Figure~\ref{fig:llama-CF-stats}). Specialists reach consensus in most cases, although the consensus rate varies across datasets. Importantly, the discussion process corrects more incorrect diagnoses than it introduces new errors, demonstrating that the discussion process helps specialist agents refine their diagnostic reasoning. A key driver of these corrections is the discussion process across specialties. As illustrated in the case study (Figure~\ref{fig:case-study}), a pulmonologist initially anchors on respiratory findings to favor acute respiratory distress syndrome as the final diagnosis, but revises the diagnosis to neurogenic pulmonary edema in later rounds after the neurosurgeon and independent clinician highlight the central nervous system injury timeline. This example suggests that cross-specialty discussion is particularly helpful when a prominent finding is actually a downstream consequence of an underlying cause, rather than the primary diagnosis itself. Moreover, counterfactual evidence appears to be effective at prompting diagnosis changes when it targets findings that are specific to one hypothesis but not others, helping agents distinguish primary evidence from incidental associations. Analysis of the CPG scores shows that removing critical evidence typically leads to a drop in diagnostic probability, indicating that the system successfully identifies the clinical features that most strongly influence the diagnostic decision. Finally, human evaluation by two licensed physicians (SD and GEB) shows that our method produces reasoning traces with stronger logical coherence, fewer factual errors, and higher overall clinical trust compared with the baseline approach (Figure~\ref{fig:human-eval}).

Overall, our method is designed to address a critical limitation in clinical reasoning: the difficulty of verifying whether available evidence truly supports one diagnosis over competing alternatives. By explicitly introducing counterfactual evidence, the framework enables clinical agents to refine their internal reasoning, test the validity of diagnostic evidence, and rule out differentials in a more structured way. As shown in the case study (Figure~\ref{fig:case-study}), the framework can interpret laboratory and radiological findings, identifies key evidence, and supports diagnostic decision making through multi-round specialist discussion. The process begins with targeted specialist assignment based on the case, followed by domain-specific reports. During discussion, the framework decomposes the case into major findings, examines how each finding changes the probability of candidate diagnoses, and provides explicit justification for both selecting the most likely diagnosis and excluding alternatives. This design improves the final diagnostic accuracy and makes clinical reasoning more interpretable. More broadly, a framework that explicitly checks which evidence supports each diagnosis and flags when evidence is weak or non-discriminating could support safer clinical decision-making, especially in settings where rare diseases or complex presentations increase the risk of diagnostic error.

Despite these promising results, several limitations remain. First, due to the computational cost of running the multi-agent diagnostic system across multiple models and random seeds, our evaluation uses relatively small test sets (80 cases for MIMIC and ER-Reason and 100 cases for MedCaseReasoning), which is common in prior studies \cite{kim2024mdagents, hager2024evaluation}. Although we report results across three random seeds for all LLMs, larger-scale evaluations would further strengthen the reliability of the findings. Second, our human evaluation includes 12 pairs of reasoning traces annotated by two physicians. While this provides an initial assessment of reasoning quality, it does not capture the full diversity of clinical presentations or diagnostic complexity. Future work could expand the evaluation to a larger panel of clinicians and a broader set of cases. Third, the CPG score requires access to model log probabilities, which limits the method to models that expose token-level probability outputs. Many closed-source API models do not provide such access, and alternative confidence estimation methods may be needed for broader deployment. Finally, our counterfactual case editing module is currently designed for text-based clinical case reports. In real-world clinical practice, diagnosis often relies on multi-modal information, including imaging data and time-series vital signs. Extending the framework to support multi-modal clinical inputs would be an important direction for future research.

Several future directions may further strengthen this line of work. Extending counterfactual case editing to multi-modal clinical inputs, such as radiology reports paired with images or time-series laboratory data, would broaden the range of clinical tasks the system can support. It would also be valuable to explore how to incorporate structured clinical knowledge to guide agents during complex reasoning, as the current approach encourages agents to test clinical hypotheses but cannot guarantee the correctness of reasoning without external knowledge support. More broadly, counterfactual reasoning provides a promising strategy for making LLM decision processes more transparent and testable in high-stakes domains. We hope this work encourages further research on integrating counterfactual thinking into language model reasoning systems.

\section{Methods}
\label{sec:method}
The design of our counterfactual case editing-based multi-agent diagnostic system (Figure~\ref{fig:system}) has four main components: (1) case triage and top-n differential diagnosis generation, (2) assignment of specialized agent roles and case summarization, (3) multi-round discussion enhanced by counterfactual case editing among the agents, and (4) final diagnosis selection via a judgment agent or consensus and reasoning traces summarization. These components implement LLM-based clinical reasoning through counterfactual reasoning, comparison across hypotheses, and iterative discussion.

\subsection{Triage and Differential Generation} \label{triage-ddx}
\paragraph{Triage Agent} We first prompt an LLM to read the raw case presentation and identify the key clinical problem and context. Unlike previous work that uses a fixed number of specialist agents for all cases \cite{chen2025enhancing}, we introduce a triage agent that automatically decides how many and which specialists should participate in the discussion. Given a case, the triage agent selects a team of experts with diverse specialties. The selected specialists are chosen to be most relevant to the case (e.g., cardiologist for chest pain, neurologist for focal deficits). Zhou et al.~\cite{zhou2025large} introduce a comprehensive clinical specialty taxonomy with 12 various specialties. Following prior studies \cite{chen2025enhancing, tang2024medagents, kim2024mdagents, zhou2025large}, we aggregate a domain-specific specialist pool by adapting roles from prior studies, resulting in 43 specialist roles.

\paragraph{Differential Generator Agent} A differential diagnosis (DDx) generation agent then synthesizes the patient case and proposes the top-n most likely DDx $\mathcal{D}=\{d_1,d_2,d_3\}$ with brief justifications.
Each specialist agent $r$ keeps a current diagnosis $s_r \in \mathcal{D}$. For example, given a case with appendix dilation, the DDx generator may list ``acute appendicitis, appendiceal mucocele, and mesenteric adenitis'' as the top hypotheses, each supported by specific findings. These top-n hypotheses serve as the diagnosis candidates that the assigned specialists will examine and discuss in later steps. Identifying multiple plausible diagnoses at the start helps explore more diagnostic possibilities to avoid locking onto one diagnosis too early.

\subsection{Report Summarization}
Following Kim et al.\ \cite{kim2024mdagents}, we summarize the input case presentation into a specialist report for each agent, tailored to that agent’s domain knowledge. This step aims to align each specialist’s view of the case while still allowing different emphasis across specialties. For example, a cardiologist may highlight hemodynamic findings, while an infectious disease specialist may focus on exposure history and laboratory results. Each summarized report concentrates the main findings that are most relevant to that specialist.

During the multi-round discussion, each specialist receives: (1) the original case, (2) all specialists' reports, (3) top-n DDx, and (4) top-k counterfactual cases. This shared but structured context encourages agents to ground their reasoning in the same clinical evidence while still contributing distinct perspectives through internal counterfactual reasoning.

\begin{figure*}[!h]
  \centering
    \includegraphics[width=\linewidth,height=9.5cm,keepaspectratio]{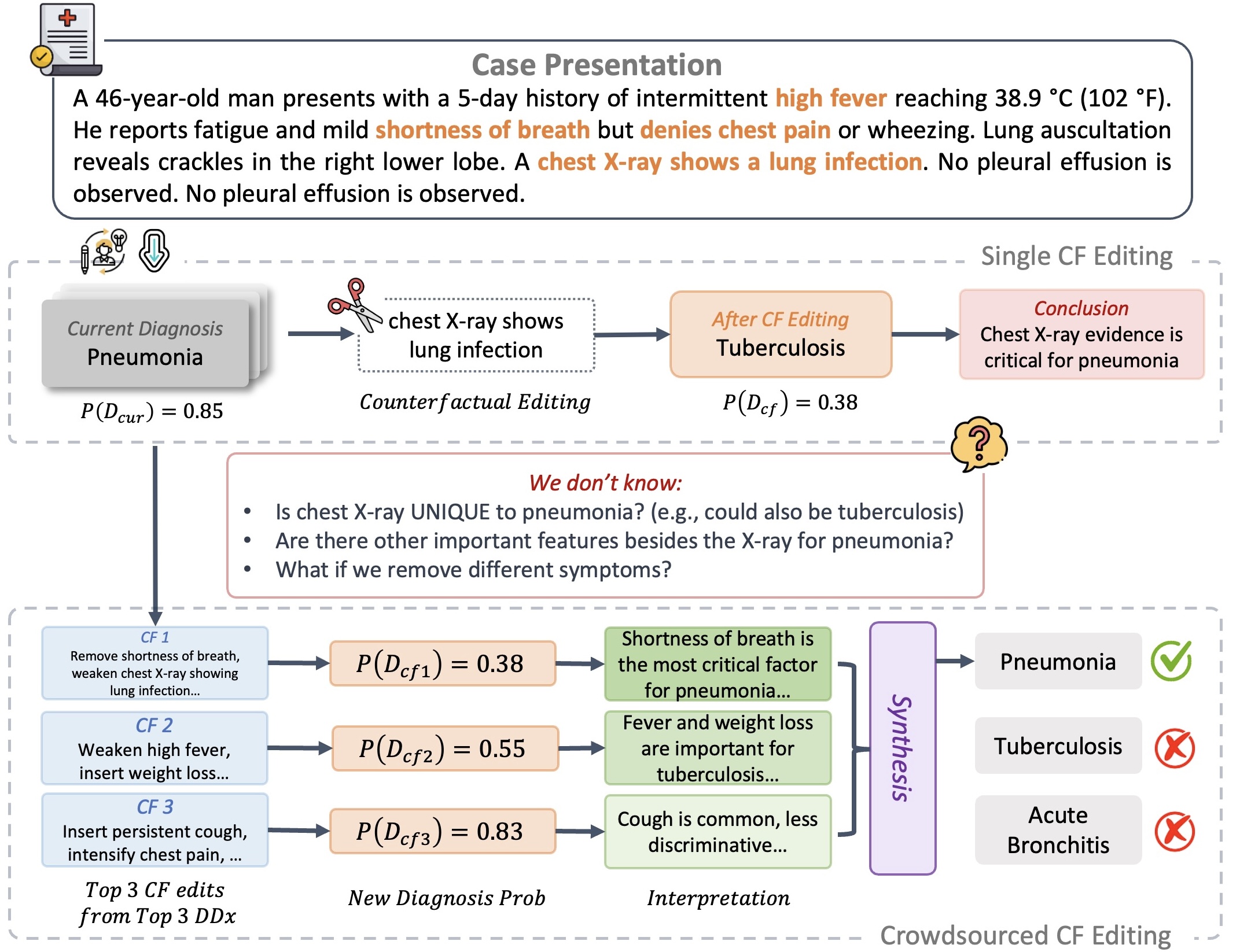}
  \caption{Example of our counterfactual case editing approach. CF: counterfactual. DDx: differential diagnosis. }
  \label{fig:CF-editing}
\end{figure*}

\subsection{Counterfactual Case Editing} \label{CF-editing}
To encourage reverse thinking during clinical reasoning, we introduce a counterfactual strategy named \textit{Counterfactual Case Editing (CE)}. CE prompts specialist agents $r$ to think about how changes in key findings would affect the diagnosis, so that they consider not only the current symptoms but also the clinical importance of each evidence (Figure~\ref{fig:CF-editing}). 

As introduced in Section~\ref{triage-ddx},the DDx agent first generates $\mathcal{D}$. For each diagnosis $d$, we generate $N$ counterfactual variants $\{C'_{1}, C'_{2}, \dots, C'_{N}\}$ given the original case $C_{\text{orig}}$ using a set of target evidence operations adapted from Hua et al.\ \cite{hua2024propagation}:
\begin{itemize}
  \item \textbf{Negate}: change a positive finding to a negative one (e.g., ``fever present'' $\rightarrow$ ``no fever'').
  \item \textbf{Remove}: delete a specific span while keeping the surrounding context.
  \item \textbf{Replace}: substitute a finding with a different but plausible value.
  \item \textbf{Weaken}: reduce the severity (e.g., ``severe pain'' $\rightarrow$ ``mild discomfort'').
  \item \textbf{Intensify}: increase the severity of a finding.
  \item \textbf{Insert}: add a new, clinically plausible finding (used sparingly).
\end{itemize}

$r$ will first extract an evidence group that support each $d$. Then, editing only those spans (i.e., negate/remove/replace/weaken/intensify/insert) while keeping the rest of the case text unchanged. These edited cases act as counterfactual probes of $C_{\text{orig}}$. For each candidate diagnosis, the specialist evaluates how the predicted diagnosis changes across the counterfactual variants, and whether those changes are consistent with medical knowledge (e.g., removing a core symptom should reduce confidence in a diagnosis that depends on it). We then apply a filtering strategy to select the most informative counterfactuals, those that cause large shifts in diagnostic confidence or reveal contradictions. These selected counterfactual cases are passed into the multi-round discussion phase to help each specialist decide which clinical features are truly essential for the final diagnosis. In this way, CE module as a counterfactual evidence verifier that steers the agents’ reasoning toward more interpretable diagnostic conclusions.

\subsection{Internal Counterfactual Reasoning} \label{counterfactual-reasoning}
\paragraph{Step 1: Case preservation scoring}
For CE, we need to keep each edit minimal that only critical clinical evidence is changed. Therefore, we propose a CE quality evaluation approach. We first compute a semantic similarity score between the original case $C_{\text{orig}}$ and each edited case $C'$ using SentenceTransformers \cite{reimers-2019-sentence-bert} embeddings \footnote{The embedding model we use for semantic similarity calculation: \url{https://huggingface.co/sentence-transformers/all-MiniLM-L6-v2}}:
\begin{equation}
\text{SemSim}(C_{\text{orig}},C')= \text{clip}_{[0,1]}\left(\frac{\cos((C_{\text{orig}}),(C_{\text{N}}'))+1}{2}\right),
\label{eq:semsim}
\end{equation}
and a heuristic string-level sequence similarity ratio $\text{EditSim}(C_{\text{orig}},C_{\text{N}}') \in [0,1]$.

\paragraph{Step 2: Counterfactual probability gap}
For each candidate edited case $C'$, we first retrieve a baseline diagnosis log probabilities (logprobs) $P_{\text{base}} = P(d\mid C_{\text{orig}})$ given the original case. We also compute the logprobs of the diagnosis that is predicted using the edited case $\hat d$ and define as
$P_{\text{CE}} = P(\hat d\mid C')$. Then, we measure how much the edit changed the specialist's prediction confidence using the Counterfactual Probability Gap (CPG), defined as:
\begin{equation}
\text{CPG}(C_{\text{orig}}, C') = \left|\, P_{\text{base}} - P_{\text{CE}} \,\right|,
\label{eq:cpg}
\end{equation}
where $P_{\text{base}}$ is the logprobs of the baseline diagnosis (from the previous discussion round). The CPG score quantifies how much the probability of the baseline diagnosis changes when a specific evidence group is removed or modified (i.e., ``\textit{If I remove/modify this evidence group, does the agent become less confident or switch to a different diagnosis?}''). 

\paragraph{Step 3: Counterfactual Selection}
To quantify the quality of CE, we filter counterfactual candidates using a Semantic and Identity Preservation (SIP) score:
\begin{equation}
\text{SIP}(C_{\text{orig}},C_{\text{N}}') = w_{sim} \times \text{SemSim}(C_{\text{orig}},C_{\text{N}}') + w_{edit} \times \text{EditSim}(C_{\text{orig}},C_{\text{N}}'),
\end{equation}

where $w_{sim}$ and $w_{edit}$ are pre-defined weights of these two scores. We require $\text{SIP} \geq 0.85$ and edit similarity $\geq 0.80$ to ensure counterfactuals represent minimal, realistic modifications. We also compute a diagnosis shift score to measure semantic distance between diagnoses predicted through $C_{\text{orig}}$ and $C'$:
\begin{equation}
\text{DiagShift}(d, \hat d) = 1 - \text{SemSim}(d, \hat d).
\end{equation}

The final combined score for selecting informative counterfactuals is:

\begin{equation}
\text{CombinedScore} = w_{sig} \times \max(\text{CPG}, w_{shift} \times \text{DiagShift}) + w_{pre} \times \text{SIP},
\label{eq:combined}
\end{equation}

where $w_{sig}$, $w_{shift}$, and $w_{pre}$ represent the weights of informativeness of the CE, diagnosis shifting, and case preservation. This algorithm prioritizes counterfactuals with high diagnostic impact (e.g., CPG score) while maintaining a preference for minimal edits (SIP). For each round, we select the top-k counterfactuals by Eq.~\eqref{eq:combined} to present to each specialist.

\subsection{Multi-round Discussion} \label{debate}
\paragraph{Initial round} After DDx generation and case summarization, each specialist begins the discussion by synthesizing all available information. In the first round, each specialist will (1) review their domain-specific report and $\mathcal{D}$; (2) conduct counterfactual reasoning (Section~\ref{counterfactual-reasoning}) for all DDx; (3) evaluates the CPG scores to identify the diagnosis relevant evidence groups; and (4) proposes an initial diagnosis selected from $\mathcal{D}$. By examining how each specialist's confidence shifts through the CPG calculation (Section~\ref{counterfactual-reasoning}), each specialist will obtain a verification signal about which clinical findings are truly discriminatory. A large drop in an specialist's confidence for a diagnosis when a finding is altered indicates that finding is crucial evidence for that diagnosis. To provide an independent viewpoint, we also include an \textit{Independent Clinician} agent. This agent audits the discussion process by performing the same diagnostic analysis but without generating counterfactuals, focusing instead on the initial symptoms, timeline, and internal consistency of the case.

\paragraph{Following rounds} In subsequent rounds, each specialist will: (1) review the ongoing discussion summary (produced by a summarizer agent) and other specialists' latest diagnoses; (2) continue generating counterfactuals to test the current leading hypotheses; (3) updates their reasoning based on counterfactual evidence and peer feedback; (4) answering questions directed to them from previous rounds asked by other specialists; (5) updates or maintains their diagnosis with justification. 

\paragraph{Consensus detection} After each round, the system checks for consensus among specialists. Consensus is reached when the majority of specialists agree on the same diagnosis:

\begin{equation}
\text{Consensus} = \begin{cases}
\text{True}, & \text{if } \frac{|\{r \in R : d = d^*\}|}{|R|} \geq 0.75 \\
\text{False}, & \text{otherwise}
\end{cases},
\end{equation}

where $R$ is the set of specialists, $d$ is specialist $r$'s diagnosis, and $d^*$ is the most common diagnosis. If consensus is reached, we adopt the consensus diagnosis directly. If no consensus is achieved after $R$ rounds, we employ a judge agent $j$ to select the final diagnosis given the discussion history. Note the judge must select from the $\mathcal{D}$ provided at the start to ensure diagnostic validity. The prompt templates of all agents are presented in Appendix~\ref{appx:prompts}.

\subsection{Code Availability}
The underlying code for this study is available in GitHub and can be accessed via this link:  \url{https://github.com/FAIRHealth/clinical-counterfactual-reasoning}.

\section{Data Availability}
The data supporting the findings of this study are available from Hugging Face and PhysioNet; however, restrictions apply to the availability of these data, which were used under license for the current study and are therefore not publicly available.

\section{Research Funding}
Yue Guo, National Science Foundation, This work used the Delta GPU at NCSA through allocation [CIS240504] from the Advanced Cyberinfrastructure Coordination Ecosystem: Services \& Support (ACCESS) program, which is supported by U.S. National Science Foundation grants \#2138259, \#2138286, \#2138307, \#2137603, and \#2138296.

\bibliographystyle{unsrt}
\bibliography{references}

\clearpage
\appendix

\section{Experiment Details}

\subsection{Datasets} \label{appx:datasets}
We evaluate our proposed diagnostic system on three datasets: MIMIC-CDM-FI \cite{hager2024evaluation}, MedCaseReasoning \cite{wu2025medcasereasoning}, and ER-REASON \cite{mehandru2025er}. All these datasets provide patient's case presentation for clinical diagnostic tasks. \textbf{MIMIC-CDM-FI} consists of 2,400 real-world cases from the emergency department (ED) across four common abdominal pathologies (i.e., appendicitis, cholecystitis, diverticulitis, and pancreatitis) built upon MIMIC-IV \cite{johnson2020mimic}. It provides all diagnosis-relevant information (e.g., history of present illness, physical exam, abdominal imaging, and guideline-selected laboratory data). Following the experiment design of MIMIC-CDM-FI, we randomly select 80 cases (20 for each disease) for all the experiments in this study. \textbf{MedCaseReasoning} is an open-access dataset of 14,489 real diagnostic QA cases derived from PubMedCentral\footnote{\url{https://pmc.ncbi.nlm.nih.gov/tools/openftlist/}} case reports, each paired with clinician-authored diagnostic rationales. Each example is summarized into a short case presentation, an enumerated reasoning statement, and a final diagnosis. \textbf{ER-REASON} contains 25,174 longitudinal clinical notes from 3,984 patients in the emergency room (ER). It is designed to evaluate LLM-based reasoning across the end-to-end emergency room workflow, from triage intake to final disposition. Due to the input length limitation of five open-source LLMs (\texttt{Llama}, \texttt{Qwen}, \texttt{m1}, \texttt{MedReason}, and \texttt{medgemma}), we summarize the case presentations of MIMIC-CDM-FI and ER-REASON following the same prompting protocol (i.e., Prompt 1) of Wu et al. \cite{wu2025medcasereasoning}. Due to the high computational cost and time-intensive nature of the clinical evaluation, we randomly sample 100 cases from MedCaseReasoning for experiments. For the ablation study, we randomly select a representative stratified subset of 30 cases from MedCaseReasoning.

\subsection{Dataset Statistics}

As introduced in Section~\ref{sec:main-results}, we evaluate our method on three datasets: MIMIC-CDM-FI (MIMIC) \cite{hager2024evaluation}, MedCaseReasoning \cite{wu2025medcasereasoning}, and ER-Reason \cite{mehandru2025er}. Dataset statistics are summarized in Table~\ref{tab:dataset_stats}. Given the design of our multi-agent diagnostic system and the input length limitations of LLMs, we summarize the input case presentations for MIMIC and ER-Reason using the same summarization prompt template proposed in MedCaseReasoning \cite{wu2025medcasereasoning}. For ER-Reason, we extract the fields used in Task 4 (Final Diagnosis Task) and include them in the summarization process. Specifically, the input fields include patient age, sex, chief complaint, past medical history, and ER clinical presentation. This preprocessing step standardizes the case format across datasets while keeping the most clinically relevant information for diagnostic reasoning.

\begin{table}[!ht]
\footnotesize
\begin{tabular*}{\textwidth}{@{\extracolsep\fill}cp{1.1cm}p{1.1cm}p{1.1cm}p{1.1cm}p{3cm}}
\toprule
                                 & \centering Dataset size   & \centering Testing size   & \centering Original  Length & \centering Summarized Length & \centering Source \tabularnewline
\midrule
MIMIC-CDM-FI \cite{hager2024evaluation}                   &  \centering 2,400  &     \centering 80    &   \centering 1,549  &  \centering 1,046  &  \centering ICU database from Beth Israel Deaconess Medical Center \tabularnewline
\midrule
MedCaseReasoning \cite{wu2025medcasereasoning}                    &  \centering 14,489  &     \centering 100    &   \centering 360  & \centering -  &  \centering PubMed Central \tabularnewline
\midrule
ER-REASON \cite{mehandru2025er}           &     \centering 25,174           &       \centering 80         &      \centering 1,830  &  \centering 647  &  \centering Emergency department data from UCSF Medical Center \tabularnewline
\bottomrule
\end{tabular*}
\caption{Comparison of three clinical diagnosis datasets. We calculate the average case length in token level using \texttt{GPT-4o} through the OpenAI tokenizer.}
\label{tab:dataset_stats}
\end{table}

\subsection{Baseline Models} \label{appx:baseline-models}

In this study, we evaluate diagnostic prediction performance using six leading open-source LLMs and one close-source LLM.

\paragraph{DeepSeek R1 (\texttt{Deepseek}).}
\texttt{DeepSeek R1} \cite{guo2025deepseek} is an open-source reasoning-focused LLM with over 600B parameters that can generate explicit intermediate reasoning traces. In our experiments, we use the official \texttt{DeepSeek-R1-0528}\footnote{\url{https://huggingface.co/deepseek-ai/DeepSeek-R1-0528}} weights released on Hugging Face and deploy the model through the Microsoft Azure AI platform\footnote{\url{https://azure.microsoft.com/en-us/solutions/ai}}.

\paragraph{GPT-5 mini (\texttt{GPT5mini}).}
\texttt{GPT-5 mini}\footnote{\url{https://developers.openai.com/api/docs/models/gpt-5-mini}} is a cost-efficient OpenAI reasoning model designed to perform well on tasks such as mathematics, programming, and complex reasoning while maintaining lower latency and computational cost. In this study, we use the version \texttt{gpt-5-mini-2025-08-07} and access the model through the Microsoft Azure AI platform.

\paragraph{Llama 3.1 8B Instruct (\texttt{Llama}).}
\texttt{Llama-3.1-8B-Instruct}\footnote{\url{https://huggingface.co/meta-llama/Llama-3.1-8B-Instruct}} \cite{llama3-2024} is an autoregressive language model developed by Meta that adopts an optimized Transformer architecture and is designed for instruction-following tasks.

\paragraph{Qwen 3 8B (\texttt{Qwen}).}
\texttt{Qwen 3 8B}\footnote{\url{https://huggingface.co/Qwen/Qwen3-8B}} is an 8.2-billion-parameter mixture-of-experts (MoE) language model from Alibaba’s Qwen3 series \cite{yang2025qwen3}. The model is designed to balance strong reasoning ability with efficient inference on a single GPU. It supports both ``thinking'' and ``non-thinking'' modes; we use the ``thinking'' mode in all experiments.

\paragraph{m1-7b-23k (\texttt{m1}).}
\texttt{m1-7b-23k} is a medical reasoning model fine-tuned from \texttt{Qwen2.5-7B-Instruct} \cite{qwen2.5} using 23,000 carefully curated medical reasoning examples. This fine-tuning process improves reasoning efficiency and diagnostic accuracy in medical tasks \cite{huang2025m1}.

\paragraph{MedReason (\texttt{MedReason}).}
\texttt{MedReason-8B}\footnote{\url{https://huggingface.co/UCSC-VLAA/MedReason-8B}} \cite{wu2025medreason} is a lightweight language model trained on the MedReason dataset\footnote{\url{https://huggingface.co/datasets/UCSC-VLAA/MedReason}}, a large-scale medical reasoning dataset designed to support faithful and explainable medical problem-solving with LLMs.

\paragraph{MedGemma 1.5 4B (\texttt{medgemma}).}
\texttt{medgemma-1.5-4b-it}\footnote{\url{https://huggingface.co/google/medgemma-1.5-4b-it}} is a decoder-only Transformer model from Google’s MedGemma 1.5 family\footnote{\url{https://research.google/blog/next-generation-medical-image-interpretation-with-medgemma-15-and-medical-speech-to-text-with-medasr/}}. The model is instruction-tuned for medical applications and can process both text and images, although it generates text-only outputs in our experiments.

\paragraph{Other Multi-Agent Baselines}
We also compare our method with several state-of-the-art multi-agent clinical reasoning frameworks, including MAC \cite{chen2025enhancing}, MDAgents \cite{kim2024mdagents}, and MedAgents \cite{tang2024medagents}.

\subsection{Evaluation} \label{appx:evaluation}

Because the model-predicted diagnoses are generated in free-form text, simple exact or partial string matching may fail to correctly identify semantically equivalent diagnoses. Therefore, following prior work \cite{wu2025medcasereasoning}, we adopt the same LLM-as-a-judge prompt to evaluate the correctness of model-predicted diagnoses (see Appendix~\ref{appx:llm-judge-prompt} for the full prompt). Specifically, the predicted diagnosis is compared with the ground-truth diagnosis using a structured grading prompt that determines whether the prediction is clinically correct. Based on the comprehensive evaluation reported in HealthBench \cite{arora2025healthbench}, \texttt{GPT-4.1} has been shown to achieve expert-level agreement as a model-based grader across multiple healthcare evaluation tasks. Therefore, in all experiments we use \texttt{GPT-4.1}\footnote{\url{https://developers.openai.com/api/docs/models/gpt-4.1}} (version: \texttt{gpt-4.1-2025-04-14}) as the evaluation model and set the temperature to 0 to ensure deterministic grading.

\subsection{Implementation Details}

All agents in our proposed system are implemented using the same underlying language model for a given experiment, following the hyperparameters recommended in the model's official documentation or previous studies \cite{wu2025medcasereasoning}. Diagnosis log-probabilities are derived directly from token-level probabilities in the open-source LLMs outputs using structured tags, enabling us to compute continuous probability estimates rather than relying on discrete confidence scores. For \texttt{DeepSeek} and \texttt{GPT5mini}, we use prompt-based probability self-reports when ``logprobs'' are unavailable. During counterfactual case editing, candidate edits are filtered and ranked using a combined scoring function. To improve computational efficiency, we cache probability computations during inference.

Across all experiments, we set the temperature to 1 for \texttt{DeepSeek} and \texttt{GPT5mini}. For the other open-source LLMs, we follow the temperature and top-p settings used in MedCaseReasoning \cite{wu2025medcasereasoning}. The only exception is \texttt{Qwen}, for which we use the recommended decoding settings: temperature=0.6, top-p=0.95, and top-k=20. We use the vLLM library \cite{kwon2023efficient} to enable efficient inference with reduced memory usage and improved computational throughput. For experiments with LLMs, we set the hyperparameters as the same as the previous study \cite{wu2025medcasereasoning}. All experiments for each model in this study are evaluated with three random seeds on each dataset. Due to the higher computational cost of running large models (\texttt{DeepSeek} and \texttt{GPT5mini}), we randomly sample 50 cases for MIMIC and ER-Reason, and 52 cases for MIMIC (13 cases for each disease) and report the evaluation results in Figure~\ref{fig:deepseek-baseline}--~\ref{fig:gpt-baseline}. For case presentation summarization, we use \texttt{GPT-5}\footnote{\url{https://developers.openai.com/api/docs/models/gpt-5}} (version \texttt{gpt-5-2025-08-07}) through the Azure API service. The reasoning effort is set to ``medium,'' and the default temperature is set to 1. The prompt template used for case summarization is provided in Appendix~\ref{appx:case-summarization}.

\subsection{Statistical Tests} \label{appx:stat-tests}

To assess the statistical significance of performance differences between our proposed approach and other experimental settings, we compare binary evaluation outcomes using the exact McNemar’s test. This test is appropriate for paired binary outcomes and evaluates whether two methods differ significantly in their prediction accuracy on the same set of cases. To control the family-wise error rate (FWER) across multiple comparisons in the ablation study, raw p-values are adjusted using the Holm–Bonferroni correction method.

For the human evaluation of model-generated reasoning quality, we further analyze inter-annotator agreement (IAA) between the two physician reviewers. Specifically, we compute Cohen’s kappa ($\kappa$) for the binary labels used in the ``Error and Safety Assessment'' criteria, and weighted Cohen’s kappa ($\kappa_w$) for the Likert-scale ratings used in the ``Reasoning Quality Assessment'' and ``Clinical Contribution'' dimensions.

\begin{figure*}[!ht]
  \centering

  \begin{subfigure}[t]{0.22\textwidth}
    \centering
    \includegraphics[width=\linewidth,height=6.5cm,keepaspectratio]{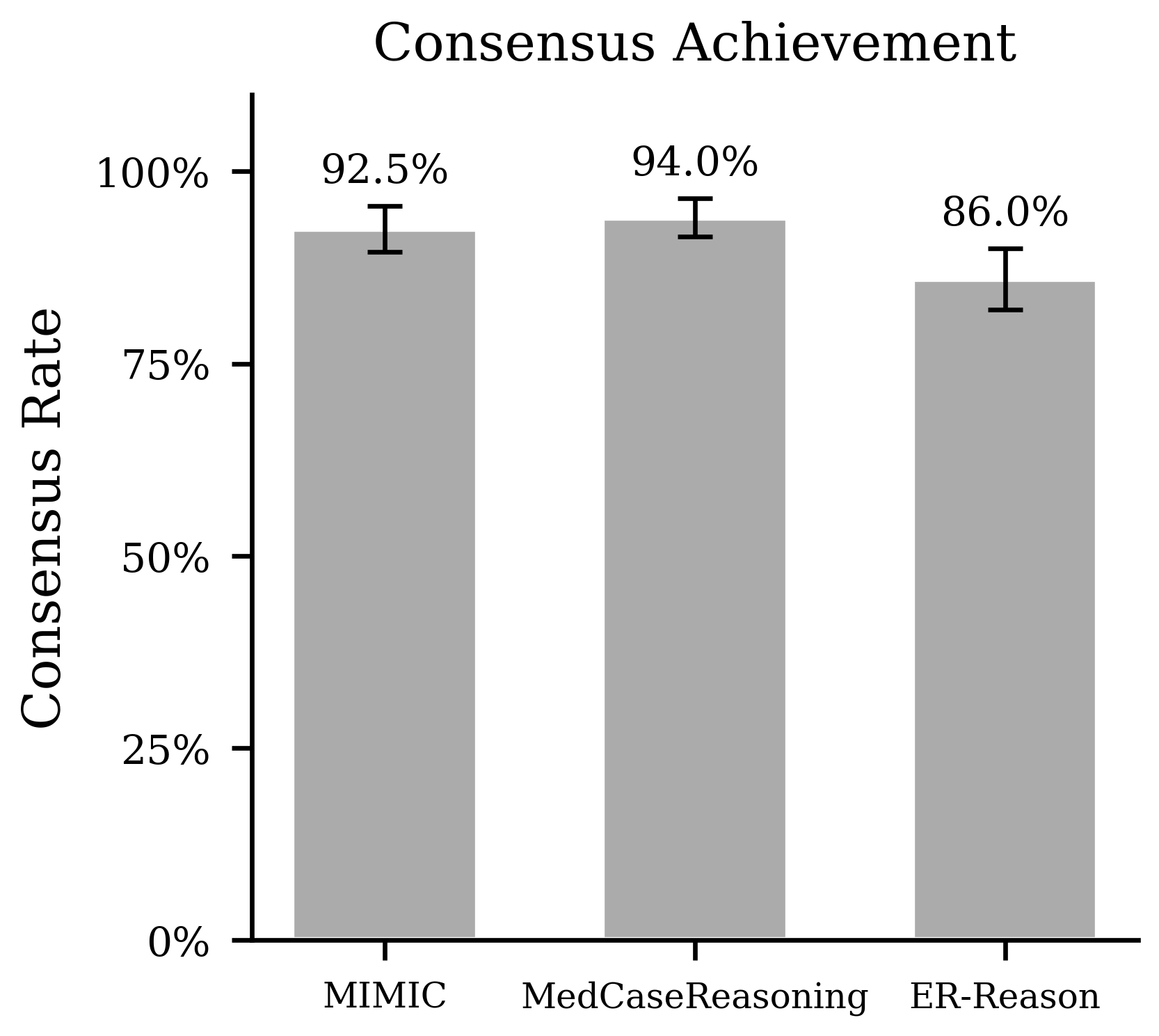}
    \caption{}
    \label{fig:llm-consensus-rate}
  \end{subfigure}
  \hfill
   \begin{subfigure}[t]{0.22\textwidth}
    \centering
    \includegraphics[width=\linewidth,height=6.5cm,keepaspectratio]{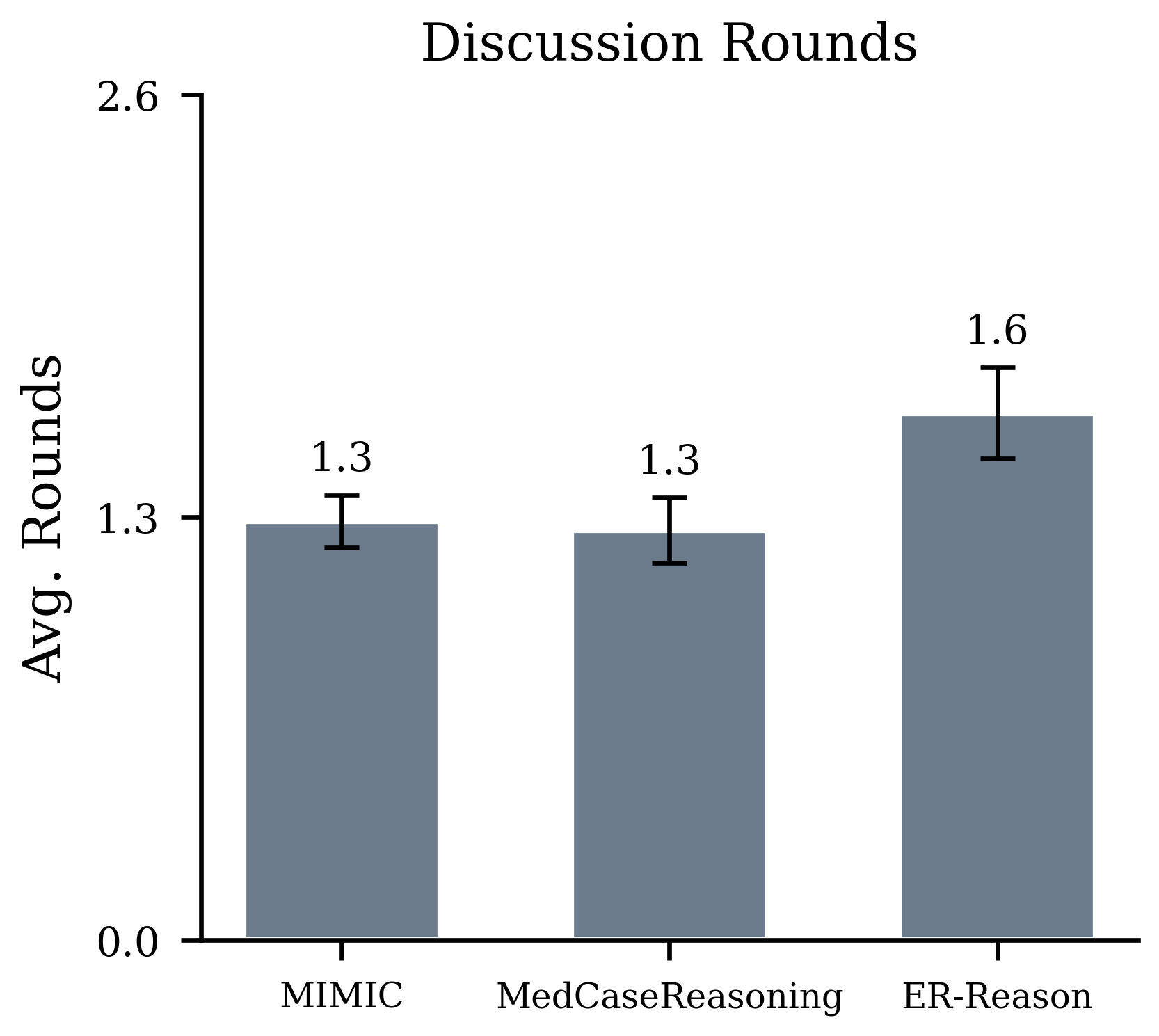}
    \caption{}
    \label{fig:llm-debate-rounds}
  \end{subfigure}
  \hfill
  \begin{subfigure}[t]{0.22\textwidth}
    \centering
    \includegraphics[width=\linewidth,height=6.5cm,keepaspectratio]{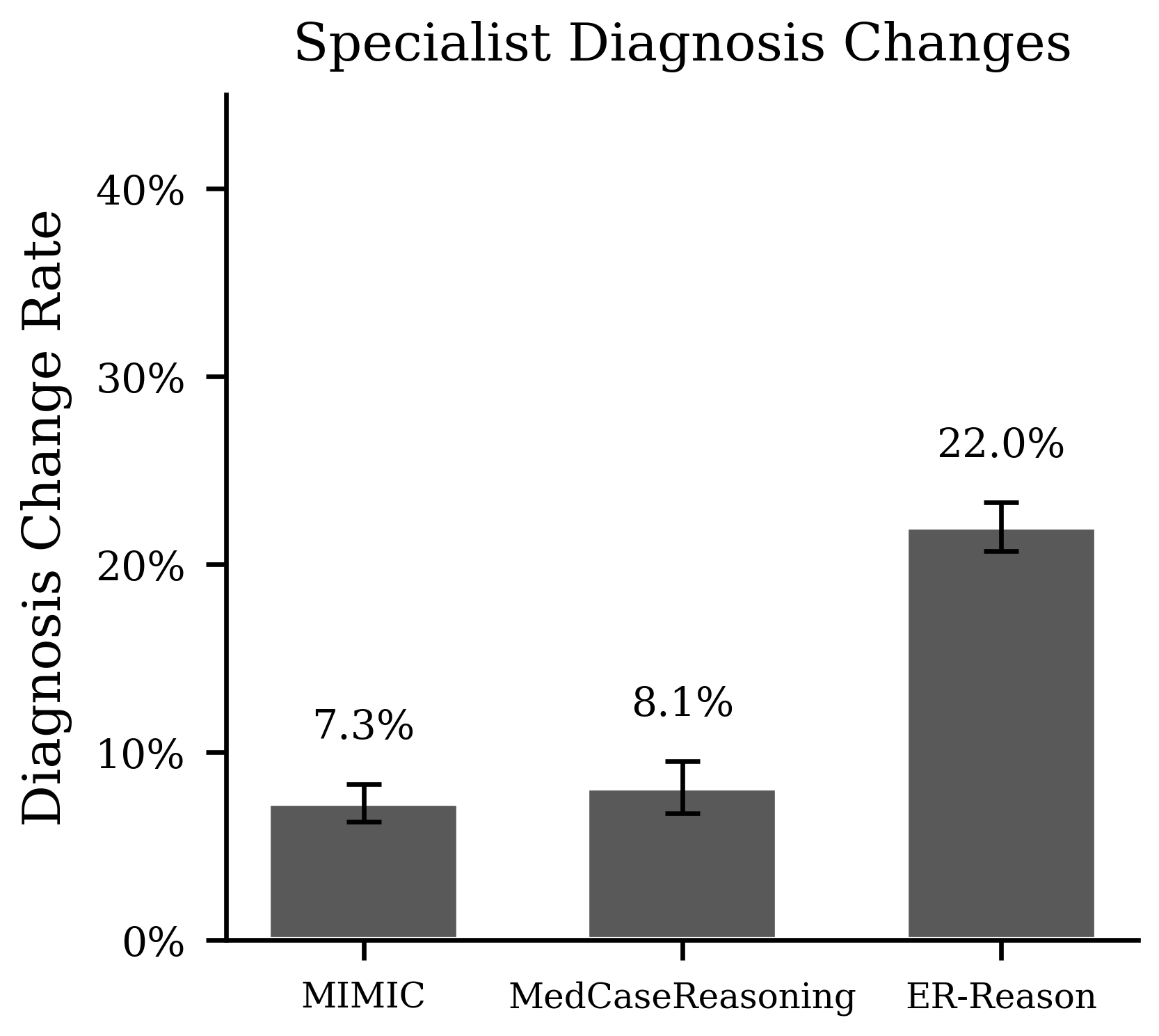}
    \caption{}
    \label{fig:llm-stance-change}
  \end{subfigure}
  \hfill
  \begin{subfigure}[t]{0.28\textwidth}
    \centering
    \includegraphics[width=\linewidth,height=6.5cm,keepaspectratio]{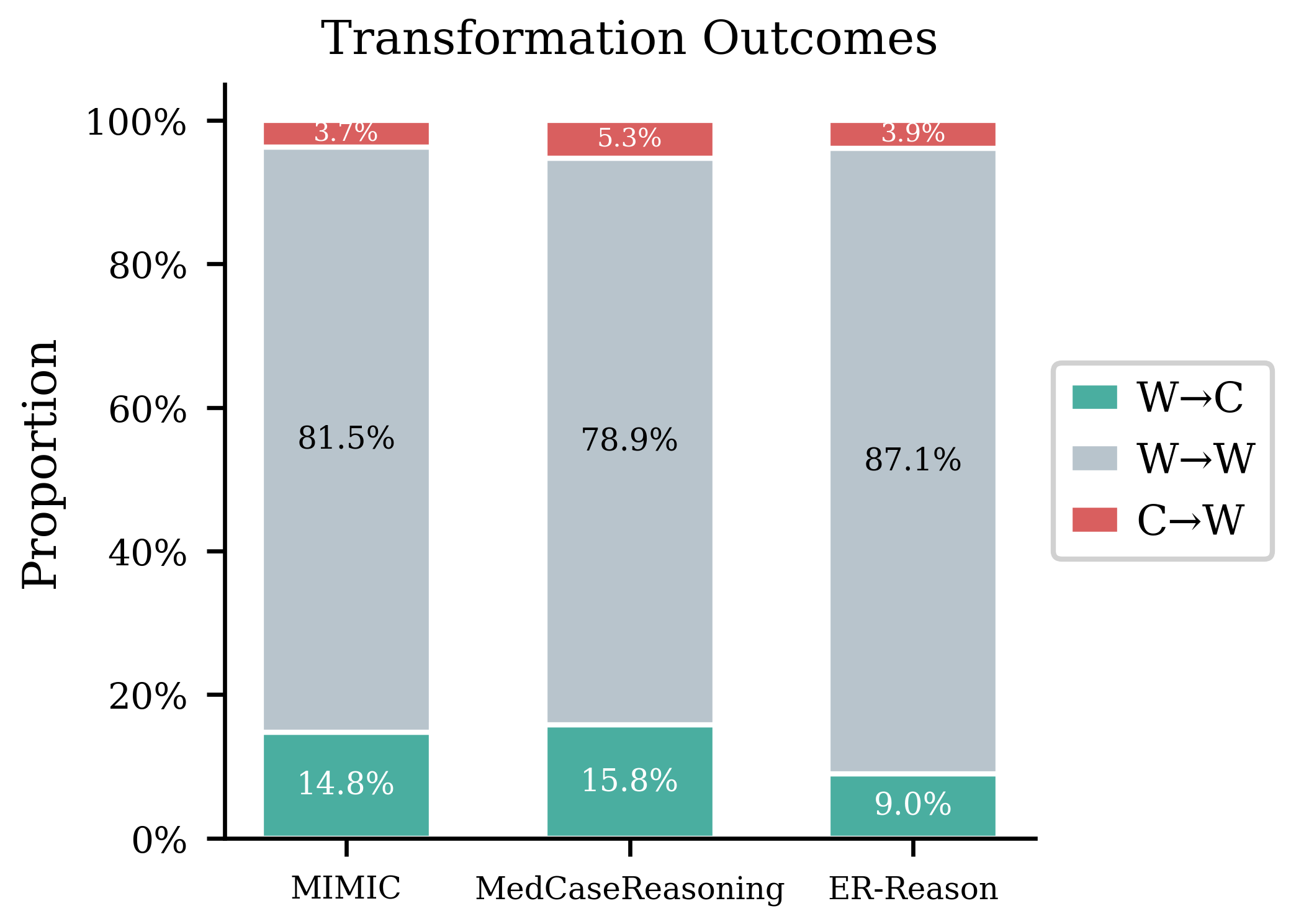}
    \caption{}
    \label{fig:llm-stance-transformation}
  \end{subfigure}

  \vspace{0.75em}

  \begin{subfigure}[t]{\textwidth}
    \centering
    \begin{subfigure}[t]{0.32\textwidth}
      \centering
      \includegraphics[width=\linewidth]{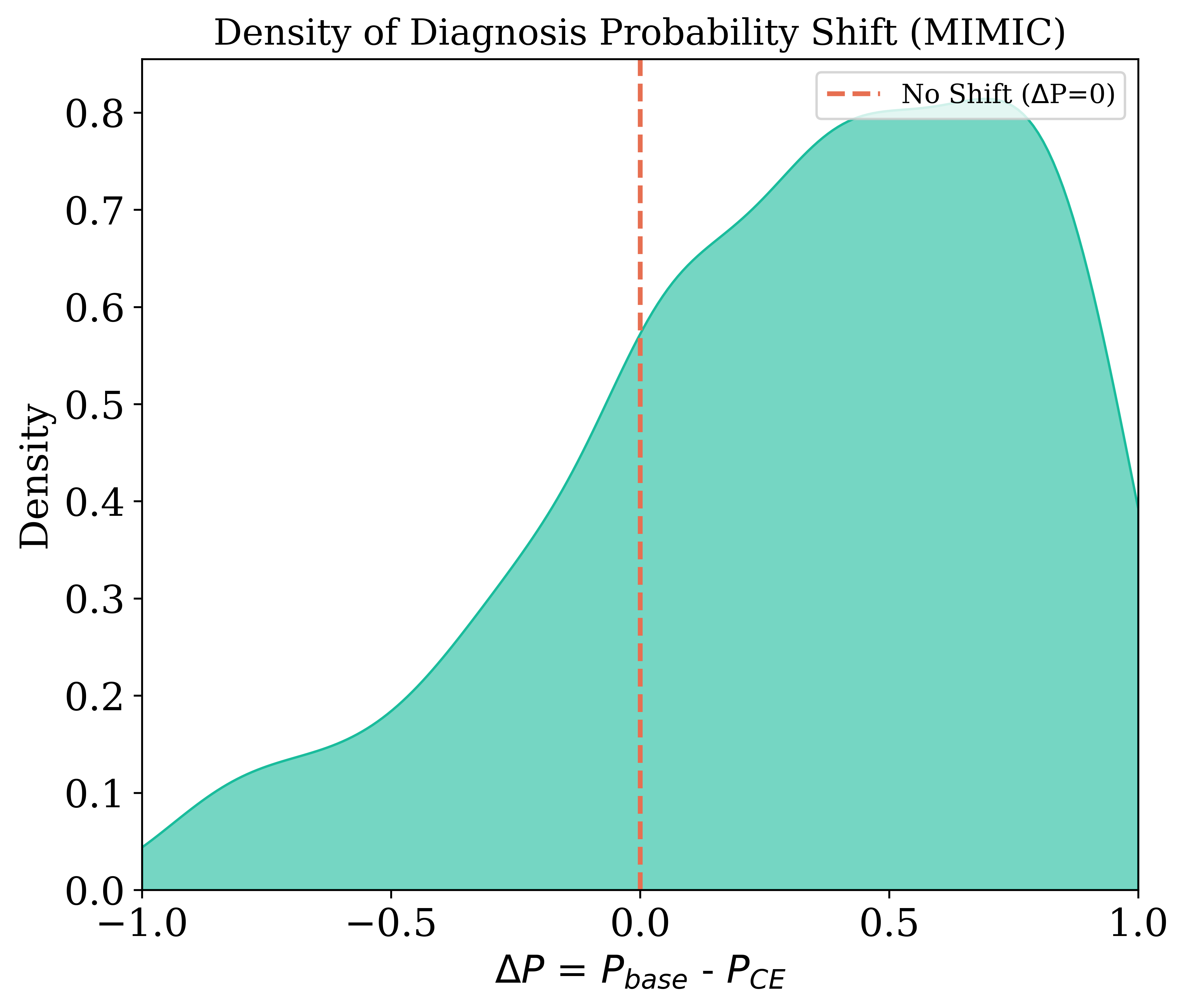}
    \end{subfigure}
    \hfill
    \begin{subfigure}[t]{0.32\textwidth}
      \centering
      \includegraphics[width=\linewidth]{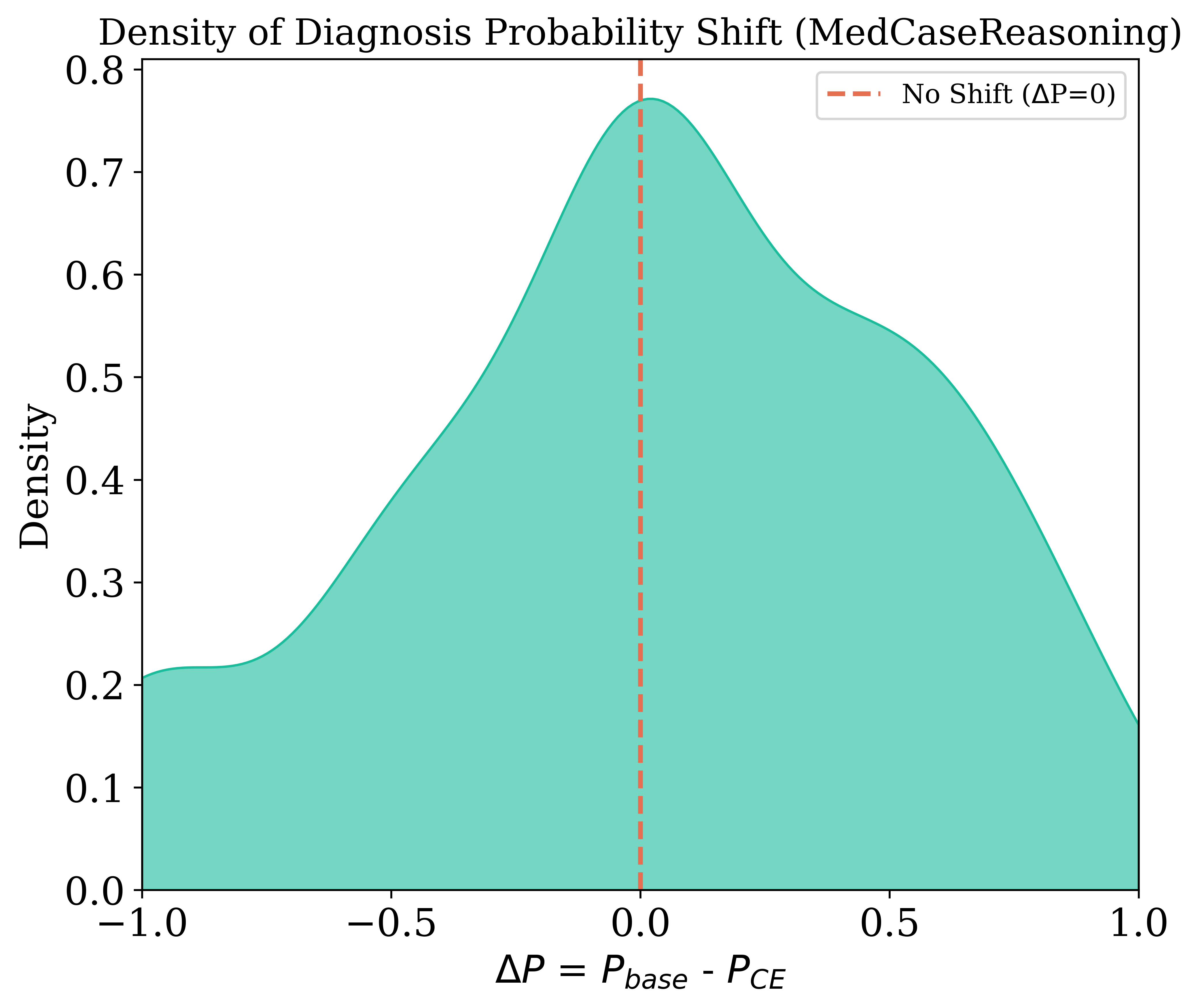}
    \end{subfigure}
    \hfill
    \begin{subfigure}[t]{0.32\textwidth}
      \centering
      \includegraphics[width=\linewidth]{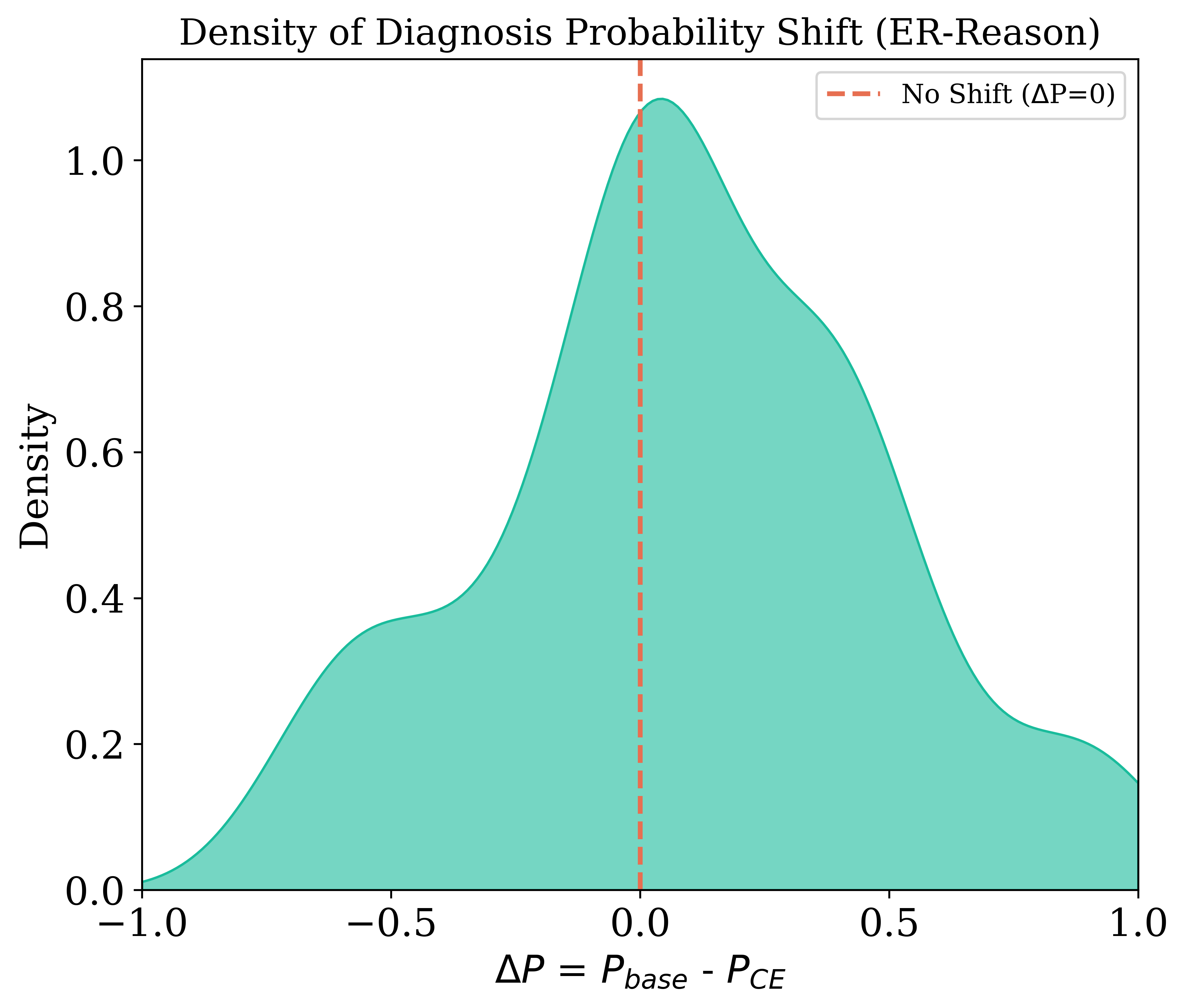}
    \end{subfigure}
    \caption{}
    \label{fig:deepseek-density}
  \end{subfigure}

  \vspace{0.75em}
  \begin{subfigure}[t]{\textwidth}
    \centering
    \begin{subfigure}[t]{0.32\textwidth}
      \centering
      \includegraphics[width=\linewidth]{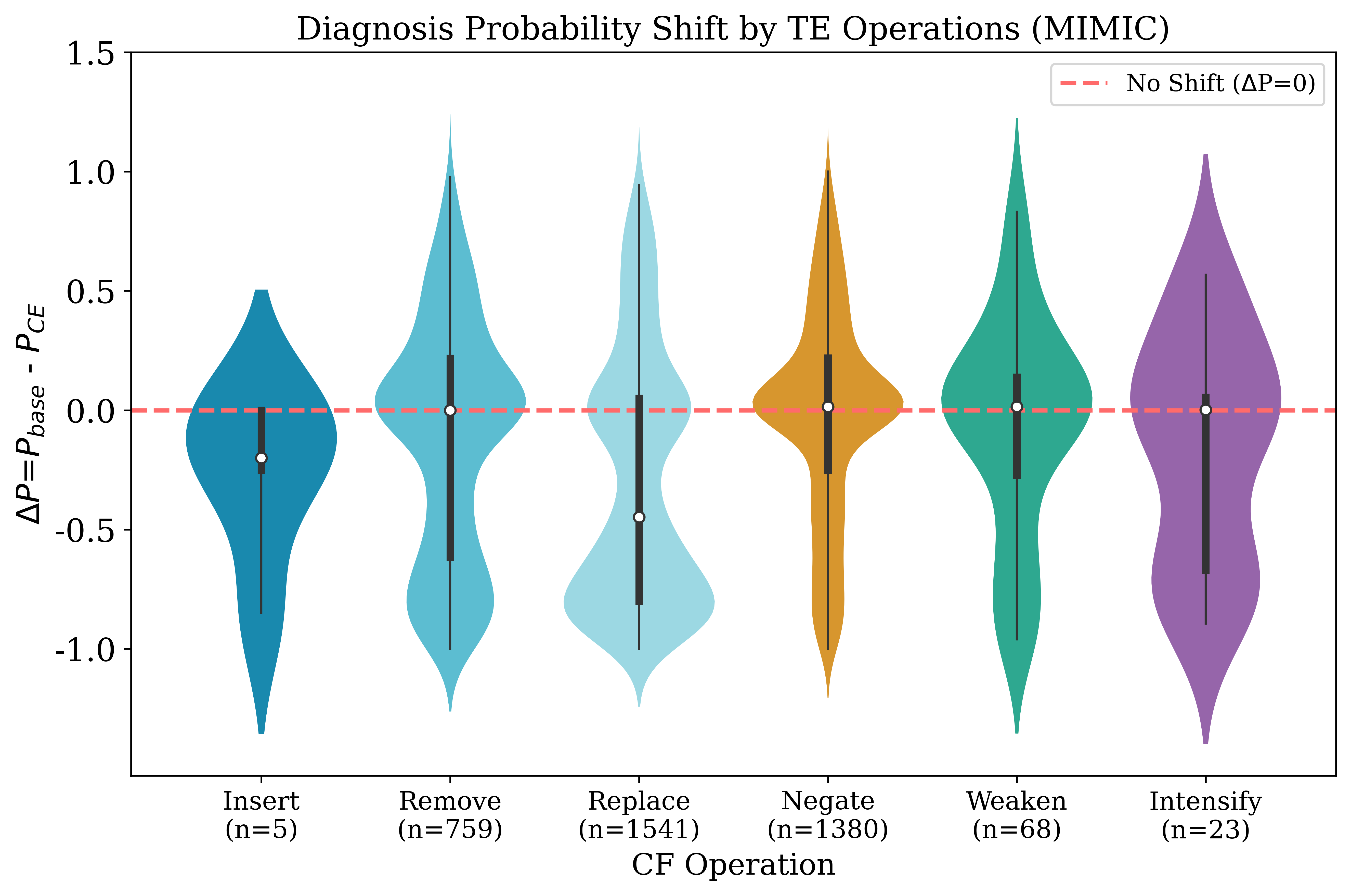}
    \end{subfigure}
    \hfill
    \begin{subfigure}[t]{0.32\textwidth}
      \centering
      \includegraphics[width=\linewidth]{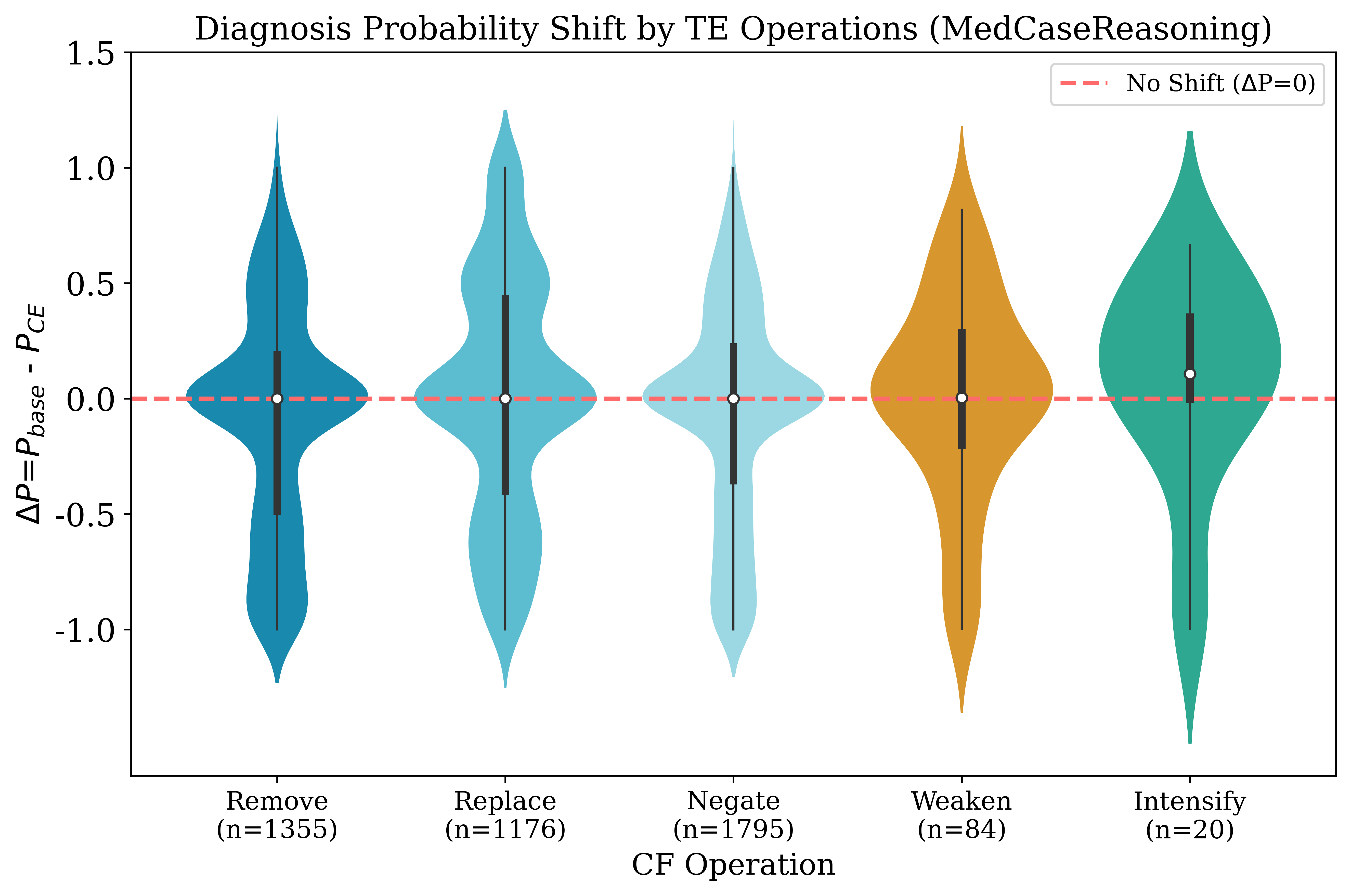}
    \end{subfigure}
    \hfill
    \begin{subfigure}[t]{0.32\textwidth}
      \centering
      \includegraphics[width=\linewidth]{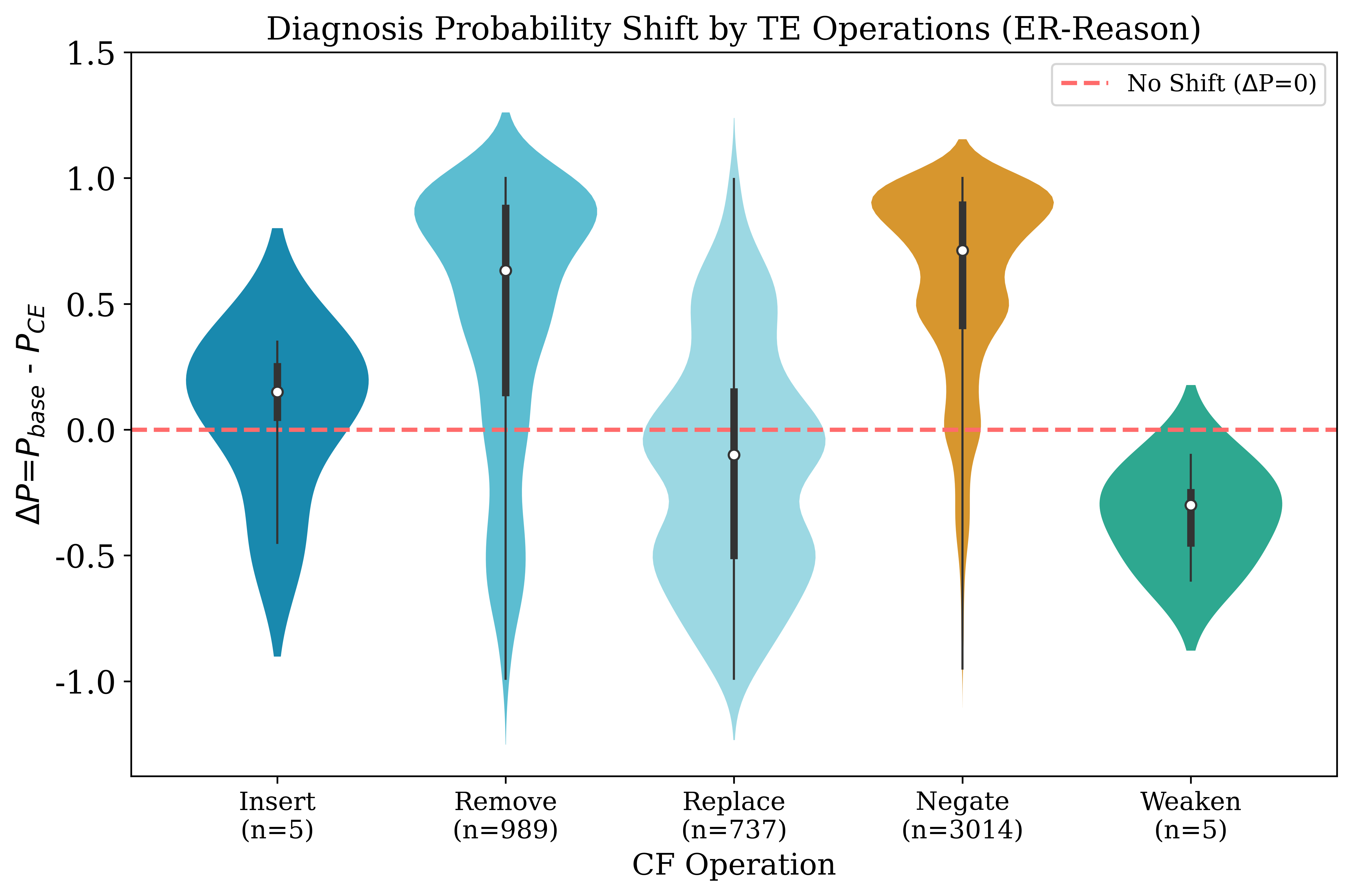}
    \end{subfigure}
    \caption{}
    \label{fig:deepseek-violin}
  \end{subfigure}
\caption{Performance of \deepseek \space for the diagnosis of case presentation on three datasets. \textbf{(a)}, Consensus rate achieved by the multi-round discussion process across datasets. \textbf{(b)}, Average number of the discussion rounds required per case. \textbf{(c)}, Specialist diagnosis-change rate across the three datasets. Error bars indicate the standard deviation across three random seeds. Bar graphs indicate the standard deviation of averaged results across three random seeds. \textbf{(d)}, Outcomes of diagnosis transformations, categorized by the correctness of the initial and final diagnoses relative to the gold standard (W: wrong, C: correct). Note that there is no C $\rightarrow$ C category in LLM-based diagnosis transformation outcomes. \textbf{(e)}, Probability density of the diagnostic hypothesis before ($P_{base}$) and after ($P_{CE}$) targeted evidence perturbation during counterfactual case editing. Only cases where the predicted diagnosis remains unchanged before and after CF editing are included. \textbf{(f)}, Distribution of diagnosis probability shifts ($\Delta P$) across different target evidence operations on the three datasets. TE: target evidence. All statistics are calculated by averaging the results of three random seeds.}
  \label{fig:LLM-debate-stats}
\end{figure*}

\section{Supplementary Results} \label{appx:supp-results}

\subsection{Main Results with Standard Deviation} \label{appx:std-results}
\begin{table}[t]
\scriptsize
\setlength{\tabcolsep}{3pt} 
\centering

\begin{tabularx}{\columnwidth}{l l l >{\centering\arraybackslash}X >{\centering\arraybackslash}X >{\centering\arraybackslash}X >{\centering\arraybackslash}X}
\toprule
\textbf{Model} & \textbf{Setting} & \textbf{Method} & \textbf{MIMIC} & \textbf{MedCaseReasoning} & \textbf{ER-Reason} & \textbf{Average Acc.} \\
\midrule
\multirow{10}{*}{\textbf{Llama}}
 & \multirow{5}{*}{\textit{Prompting}} 
 & ZS     & 54.2 (1.6) & 17.3 (0.5) & 10.5 (2.4) & 27.3* (1.5) \\
 & & ZS CoT & 50.9 (3.9) & 18.0 (4.1) & 8.4 (2.6)  & 25.8* (3.5) \\
 & & FS      & 71.3 (2.0) & 17.7 (3.1) & 17.1 (3.6) & 35.4 (2.9)  \\
 & & FS CoT  & 70.0 (3.7) & 14.3 (2.5) & 17.9 (1.5) & 34.1 (2.6)  \\
 & & SC  & 54.2 (2.6) & 19.3 (6.5) & 7.9 (0.7) & 27.1* (3.3)  \\
\cmidrule(l){2-7}
 & \multirow{3}{*}{\textit{Agents}} 
 & MAC       & 55.9 (2.6) & 17.3 (3.4) & 13.8 (2.7) & 29.0* (2.9) \\
 & & MDAgents  & 70.3 (5.7) & 18.7 (0.5) & 15.9 (1.5) & 35.0 (2.6)  \\
 & & MedAgents & 69.6 (1.2) & 16.3 (3.1) & 12.9 (2.6) & 32.9 (2.3)  \\
\cmidrule(l){2-7}
& & \cellcolor[HTML]{D4EAD1}\textbf{Ours}
& \cellcolor[HTML]{D4EAD1}\textbf{75.4} (1.6)
& \cellcolor[HTML]{D4EAD1}\textbf{22.7} (0.9)
& \cellcolor[HTML]{D4EAD1}\textbf{18.8} (1.5)
& \cellcolor[HTML]{D4EAD1}\textbf{39.0} (1.3) \\
\midrule
\midrule

 & \multirow{5}{*}{\textit{Prompting}} 
 & ZS     & 68.4 (5.6) & 25.3 (3.4) & 17.5 (2.0) & 37.1 (3.7) \\
 & & ZS CoT & 70.8 (3.1) & 25.3 (2.9) & 16.7 (1.6) & 37.6 (2.5) \\
 & & FS      & 69.2 (1.6) & 23.3 (1.7) & 20.0 (2.0) & 37.5 (1.8) \\
 & & FS CoT  & 67.9 (2.6) & 23.0 (2.2) & 20.5 (3.1) & 37.1 (2.6) \\
  & & SC  & 70.4 (0.7) & 25.7 (0.5) & 20.0 (2.2) & 38.7 (1.1)  \\
\cmidrule(l){2-7}
 & \multirow{3}{*}{\textit{Agents}} 
 & MAC       & 65.9 (1.5) & 21.3 (5.2) & 19.2 (3.3) & 35.5 (3.3) \\
 & & MDAgents  & 67.1 (3.3) & 22.0 (1.6) & 18.4 (2.1) & 35.8 (2.3) \\
 & & MedAgents & 69.2 (2.4) & 24.7 (2.4) & 15.9 (1.5) & 36.6 (2.1) \\
\cmidrule(l){2-7}
\rowcolor[HTML]{D4EAD1} \cellcolor{white}\multirow{-10}{*}{\textbf{Qwen}} 
 &  \cellcolor{white} & \textbf{Ours} & \textbf{72.9} (2.6) & \textbf{26.3} (4.2) & \textbf{21.3} (1.8) & \textbf{40.2} (2.9) \\
\midrule
\midrule

 & \multirow{5}{*}{\textit{Prompting}} 
 & ZS     & 67.5 (1.7) & 20.0 (2.9) & 18.0 (6.2) & 35.2 (3.6) \\
 & & ZS CoT & 67.5 (2.0) & 18.7 (2.1) & 17.5 (0.0) & 34.6 (1.4) \\
 & & FS      & 60.0 (4.4) & 13.7 (0.9) & \textbf{19.2} (2.6) & 31.0* (2.6) \\
 & & FS CoT  & 64.2 (4.7) & 11.0 (4.2) & \textbf{19.2} (2.6) & 31.5* (3.8) \\
 & & SC  & 66.7 (1.9) & 21.0 (3.6) & 16.3 (3.3) & 34.6 (2.9)  \\
\cmidrule(l){2-7}
 & \multirow{3}{*}{\textit{Agents}} 
 & MAC       & 45.9 (6.7) & 16.3 (2.1) & 11.3 (2.7) & 24.5* (3.8) \\
 & & MDAgents  & 65.0 (3.7) & 19.0 (3.7) & 15.4 (1.5) & 33.1 (3.0)  \\
 & & MedAgents & 47.9 (2.6) & 16.0 (2.2) & 12.1 (2.1) & 25.3* (2.3) \\
\cmidrule(l){2-7}
\rowcolor[HTML]{D4EAD1} \cellcolor{white}\multirow{-10}{*}{\textbf{m1-7B-23k}} 
 & \cellcolor{white}  & \textbf{Ours} & \textbf{67.9} (2.6) & \textbf{23.3} (5.4) & \textbf{19.2} (1.6) & \textbf{36.8} (3.2) \\
\midrule
\midrule

 & \multirow{5}{*}{\textit{Prompting}} 
 & ZS     & 48.4 (6.8) & 17.3 (4.8) & 10.9 (0.6) & 25.5 (4.1)  \\
 & & ZS CoT & 49.2 (2.6) & 13.0 (3.6) & 10.9 (2.1) & 24.4* (2.8) \\
 & & FS      & 47.5 (2.7) & 8.0 (1.6)  & 14.6 (1.6) & 23.4* (2.0) \\
 & & FS CoT  & 52.9 (4.8) & 12.0 (1.6) & 17.1 (4.6) & 27.3 (3.7)  \\
  & & SC  & 50.4 (2.6) & 14.3 (0.6) & 10.8 (1.9) & 25.2 (1.7)  \\
\cmidrule(l){2-7}
 & \multirow{3}{*}{\textit{Agents}} 
 & MAC       & 51.7 (2.6) & 16.3 (3.4) & 15.0 (2.7) & 27.7 (2.9) \\
 & & MDAgents  & 47.9 (3.6) & 18.7 (4.5) & 14.2 (3.3) & 26.9 (3.8) \\
 & & MedAgents & 53.8 (5.7) & 18.0 (5.7) & 15.9 (2.1) & 29.3 (4.5) \\
\cmidrule(l){2-7}
\rowcolor[HTML]{D4EAD1} \cellcolor{white}\multirow{-10}{*}{\textbf{MedReason}} 
 &  \cellcolor{white} & \textbf{Ours} & \textbf{55.4} (5.8) & \textbf{20.3} (4.0) & \textbf{18.8} (2.0) & \textbf{31.5} (3.9) \\
\midrule
\midrule

 & \multirow{5}{*}{\textit{Prompting}} 
 & ZS     & 67.9 (0.6) & 18.0 (2.2) & 16.3 (2.0) & 34.1 (1.6) \\
 & & ZS CoT & 67.5 (1.0) & 17.7 (1.7) & 18.4 (2.1) & 34.5 (1.6) \\
 & & FS      & 68.8 (2.7) & 15.3 (0.9) & 25.4 (2.1) & 36.5 (1.9) \\
 & & FS CoT  & 63.8 (2.7) & 12.7 (3.1) & \textbf{26.7} (2.1) & 34.4 (2.6) \\
 & & SC  & 67.9 (6.2) & 19.0 (1.0) & 17.1 (0.7) & 34.7 (2.6)  \\
\cmidrule(l){2-7}
 & \multirow{3}{*}{\textit{Agents}} 
 & MAC       & 64.6 (3.1) & 18.3 (5.6) & 17.1 (2.6) & 33.3* (3.8) \\
 & & MDAgents  & 65.9 (2.6) & 12.7 (2.1) & 18.8 (1.8) & 32.5* (2.2) \\
 & & MedAgents & 67.1 (3.0) & 11.3 (0.5) & 14.2 (1.6) & 30.9* (1.7) \\
\cmidrule(l){2-7}
\rowcolor[HTML]{D4EAD1} \cellcolor{white}\multirow{-10}{*}{\textbf{medgemma}} 
 &  \cellcolor{white} & \textbf{Ours} & \textbf{79.2} (2.1) & \textbf{21.0} (1.4) & 22.5 (2.0) & \textbf{40.9} (1.8) \\
\midrule
\midrule

 & \multirow{5}{*}{\textit{Prompting}} 
 & ZS     & 75.6 (2.9) & 42.0 (5.3) & 14.0 (4.0) & 43.9* (4.1) \\
 & & ZS CoT & 84.6 (5.8) & 44.7 (4.2) & 16.7 (3.1) & 48.7 (4.4) \\
 & & FS      & 87.8 (3.0) & 44.7 (3.1) & 13.3 (2.3) & 48.6 (2.8) \\
 & & FS CoT  & 88.5 (3.9) & 50.7 (2.3) & 14.0 (4.0) & 51.1 (3.4) \\
  & & SC  & 83.3 (2.2) & 45.3 (3.1) & 16.0 (5.3) & 48.5 (3.5) \\
\cmidrule(l){2-7}
 & \multirow{3}{*}{\textit{Agents}} 
 & MAC       & 78.8 (2.0) & 40.7 (6.4) & 9.3 (2.3) & 42.9* (3.6) \\
 & & MDAgents  & 84.0 (2.9) & 38.7 (7.0) & 17.3 (3.1) & 46.7 (4.3) \\
 & & MedAgents & 85.2 (1.1)& 49.3 (5.0) & 11.3 (2.3) & 48.6 (2.8) \\
\cmidrule(l){2-7}
\rowcolor[HTML]{D4EAD1} \cellcolor{white}\multirow{-10}{*}{\textbf{Deepseek R1}} 
 &  \cellcolor{white} & \textbf{Ours} & \textbf{90.4} (1.9) & \textbf{54.0} (4.0) & \textbf{21.3} (2.3) & \textbf{55.2} (2.7) \\
\midrule
\midrule
  \multirow{10}{*}{\textbf{GPT-5 mini}}
 & \multirow{5}{*}{\textit{Prompting}} 
 & ZS     & 82.0 (3.0) & 50.7 (4.2) & 19.3 (1.2) & 50.7 (2.8) \\
 & & ZS CoT & 80.1 (3.0) & 47.3 (3.1) & 18.0 (2.0) & 48.5* (2.7) \\
 & & FS      & 82.1 (2.2) & 52.0 (3.5) & 22.0 (4.0) & 52.0 (3.2) \\
 & & FS CoT  & 81.4 (2.9) & 52.0 (2.0) & 22.7 (3.1) & 52.0 (2.7) \\
  & & SC  & 80.7 (1.9) & 50.7 (3.1) & 20.0 (5.0) & 50.5 (3.3) \\
\cmidrule(l){2-7}
 & \multirow{3}{*}{\textit{Agents}} 
 & MAC       & 78.8 (3.9) & 59.3 (3.1) & 15.3 (2.3) & 51.1 (3.1) \\
 & & MDAgents  & 90.4 (1.9) & 60.7 (2.3) & 23.3 (1.2) & 58.1 (1.8) \\
 & & MedAgents & 85.3 (3.0) & 47.3 (4.6) & 18.0 (2.0) & 50.2 (3.2) \\
\cmidrule(l){2-7}
& & \cellcolor[HTML]{D4EAD1}\textbf{Ours}
& \cellcolor[HTML]{D4EAD1}\textbf{93.6} (1.1)
& \cellcolor[HTML]{D4EAD1}\textbf{66.0} (2.0)
& \cellcolor[HTML]{D4EAD1}\textbf{26.7} (3.1)
& \cellcolor[HTML]{D4EAD1}\textbf{62.1} (2.1) \\
\bottomrule
\end{tabularx} 
\caption{Evaluation results of final diagnosis prediction accuracy on three datasets. We report the standard deviation (std.) in the brackets. ZS: zero-shot; CoT: chain-of-thought; FS: few-shot; SC: self-consistency. *$P$ < 0.05 indicates our method shows significant difference than the baselines.}
\label{tab:main-results-std}
\end{table}
As reported in Figure~\ref{fig:baseline-results}, we compare the model diagnostic performance on three datasets across LLMs. Unlike other studies using supervised fine-tuning or reinforcement learning \cite{you-etal-2024-sciprompt, zhang2025med}, our method is a training-free multi-agent diagnostic framework. Given the model generation uncertainty, we run each method through three random seeds for each dataset and report the standard deviation in Table~\ref{tab:main-results-std}.

\begin{figure*}[!ht]
  \centering
    \begin{subfigure}[t]{\textwidth}
    \centering
    \includegraphics[width=\linewidth,height=6.5cm,keepaspectratio]{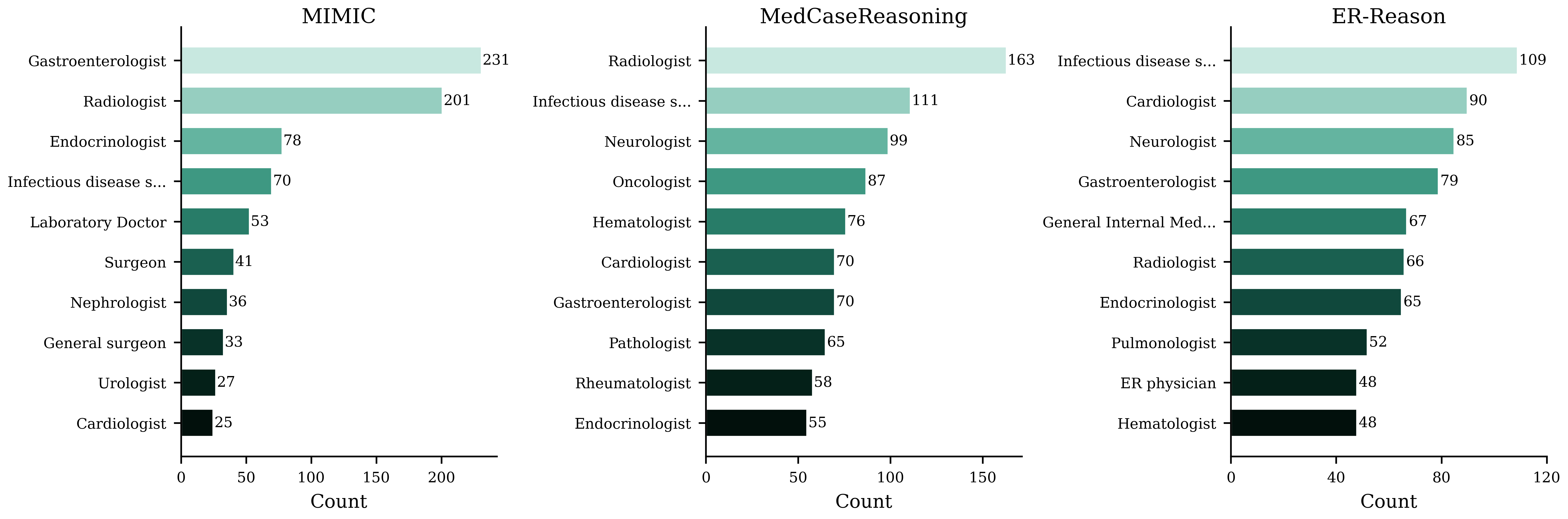}
    \caption{}
    \label{fig:slm-specialist-stats}
  \end{subfigure}
  \hfill
   \begin{subfigure}[t]{\textwidth}
    \centering
    \includegraphics[width=\linewidth,height=6.5cm,keepaspectratio]{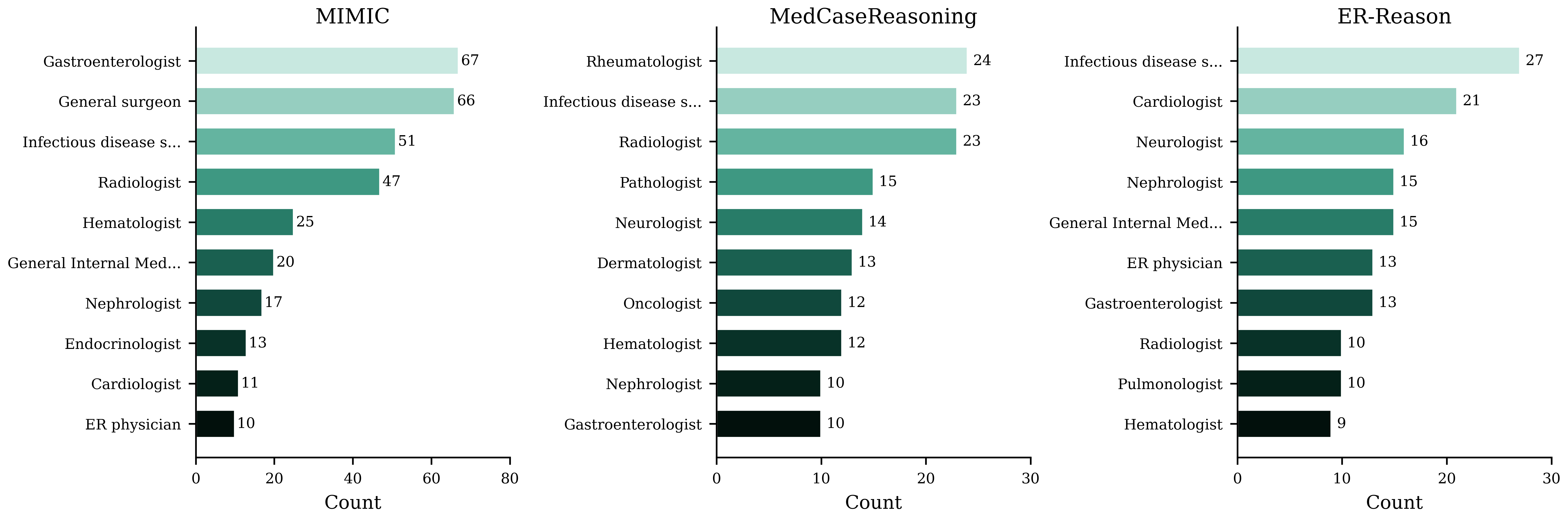}
    \caption{}
    \label{fig:deepseek-specialist-stats}
  \end{subfigure}
  \hfill

\caption{Distribution of the Top-10 most frequently assigned specialists by the triage agent for each dataset on \textbf{(a)} \llama \space and \textbf{(b)} \deepseek. }
  \label{fig:slm-specialist-stats}
\end{figure*}

\subsection{Deepseek Multi-round Discussion Analysis} \label{appx:llm-debate-analysis}

As shown in Figure~\ref{fig:LLM-debate-stats}, we report the discussion statistics of \texttt{Deepseek} on three datasets. Compared with \texttt{Llama} (Figure~\ref{fig:llama-CF-stats}), \texttt{Deepseek} shows a higher consensus rate in all three datasets, while the specialist diagnosis change rates decrease to 7.3\% for MIMIC, 8.1\% for MedCaseReasoning, and 22.0\% for ER-Reason, showing that the larger-scale LLM hardly refines the diagnosis during the discussion. Notably, \texttt{Deepseek} shows lower successful diagnosis transformation rates in MIMIC and ER-Reason compared with \texttt{Llama} (Figure~\ref{fig:llama-transofmration}) during the discussion. Similar to Figure~\ref{fig:llama-density}, we report the density distributions of diagnosis probability shifts ($\Delta P = P_{base} - P_{CE}$) on \texttt{Deepseek} across all three datasets, providing strong evidence for the effectiveness of our evidence extraction strategy. Across datasets, most probability shifts $\Delta P$ are positive, indicating that the diagnosis probability decreases after counterfactual editing. This occurs in 77.7\% of cases in MIMIC, 50.4\% in MedCaseReasoning, and 57.8\% in ER-Reason. In contrast to \texttt{Llama}, \texttt{Deepseek} shows higher uncertainty in the original diagnosis prediction in MedCaseReasoning and ER-Reason datasets.

\subsection{Specialists Assignment Distribution} \label{appx:specialist-analysis}

In additional to the discussion analysis of \texttt{Llama} model in Figures~\ref{fig:llama-consensus}--\ref{fig:llama-transofmration}, we also report the mostly assigned specialists on three datasets in Figure~\ref{fig:slm-specialist-stats}. Comparing with \texttt{Llama}, \texttt{Deepseek} assigns fewer specialists during the discussion, while the assigned specialist roles are generally matched.

\section{Human Evaluation} \label{appx:human-eval}

To evaluate the quality of LLM-generated clinical reasoning traces, we conduct a human evaluation comparing reasoning outputs from our method and the zero-shot CoT baseline. Specifically, we randomly sample 12 pairs of reasoning traces from three datasets (MIMIC, MedCaseReasoning, and ER-Reason). Within each pair, both methods produce either correct diagnoses or incorrect diagnoses for the same case, allowing evaluators to focus on differences in reasoning quality rather than diagnostic correctness. Based on the results shown in Figure~\ref{fig:baseline-results}, we observe that \texttt{Llama} shows the largest improvement when comparing our method with zero-shot CoT, whereas \texttt{m1} shows the smallest performance improvement. To ensure diversity in model behavior, we include reasoning traces from both models in the evaluation. For each model, we sample six reasoning pairs, resulting in a total of 12 evaluation pairs. Two licensed physicians independently review the reasoning traces and provide pairwise preferences while remaining blind to the underlying method and model throughout the evaluation process. To improve evaluation efficiency and consistency, we develop a custom annotation interface that allows physicians to easily review reasoning traces and identify potential reasoning issues (see Appendix~\ref{appx:annotation-interface}).

\subsection{Evaluation Protocol} \label{appx:human-eval-protocol}

We design a structured evaluation protocol to assess the quality of model-generated clinical reasoning traces, drawing inspiration from prior frameworks such as CLEVER \cite{liu2025generalist} and Revised-IDEA \cite{schaye2022development}. The protocol evaluates multiple dimensions of clinical reasoning quality, including factual correctness, reasoning completeness, logical coherence, and clinical usefulness. The evaluation criteria are summarized in Table~\ref{tab:error_safety}, Table~\ref{tab:reasoning_quality}, and Table~\ref{tab:bias_categories}.

\begin{table}[!ht]
\centering
\footnotesize
\renewcommand{\arraystretch}{1.3} 
\begin{tabular}{@{}p{0.20\textwidth} p{0.55\textwidth} p{0.15\textwidth}@{}}
\toprule
\textbf{Metric} & \textbf{Question} & \textbf{Options} \\
\midrule
\textbf{Factual Errors} & Does the reasoning contain any false/incorrect medical information? & Yes / No \\
\textbf{Hallucination} & Does the reasoning cite findings not present in the case? & Yes / No \\
\textbf{Harm} & Does this reasoning trace contain a  ``Critical Failure?'' (e.g., ignoring contraindication, recommending fatal drug dosage, ruling out life-threatening emergency without cause) & Yes / No \\
\textbf{Completeness} & Does the reasoning correctly identify key findings, risk factors, and diagnoses? & Yes / No \\
\textbf{Trust} & Which reasoning inspires more confidence in the diagnostic process? & A / B / Equal \\
\bottomrule
\end{tabular}
\caption{Evaluation metrics for error, safety, completeness, and trust assessment. Criteria are evaluated using binary or comparative scales to identify critical failures and overall confidence in the diagnostic reasoning.}
\label{tab:error_safety}
\end{table}

\begin{table}[!ht]
\centering
\footnotesize
\renewcommand{\arraystretch}{1.4}
\begin{tabular}{@{}p{0.18\textwidth} p{0.25\textwidth} p{0.5\textwidth}@{}}
\toprule
\textbf{Category} & \textbf{Question} & \textbf{Criteria} \\
\midrule
\textbf{Efficiency} & Does the reasoning communicate effectively? & 
1: Verbose or repetitive; ignores key diagnostic points \newline
2: Somewhat verbose; key points present but unclear \newline
3: Appropriate length; mostly focused; occasional redundancy \newline
4: Concise and focused; key insights highlighted \newline
5: Exceptionally clear; every statement adds evidence \\
\midrule
\textbf{Logical Coherence} & Does the reasoning demonstrate sound logical structure in ruling in/out diagnoses? & 
1: Logical failures; no clear chain from evidence to conclusion \newline
2: Incomplete logic; mentions findings but fails to connect \newline
3: Basic coherence; clear progression but may lack depth \newline
4: Strong reasoning; synthesizes findings with explanations \newline
5: Exceptional; comprehensive integration, addresses confounders \\
\midrule
\textbf{Differential Diagnosis} & Does the reasoning consider alternative diagnoses appropriately? & 
1: No differential mentioned \newline
2: Differential vaguely implied but not explicitly stated \newline
3: 1 alternative mentioned without explanation \newline
4: 1 alternative mentioned with brief justification \newline
5: $\geq$ 2 alternatives with clear reasoning for each \\
\midrule
\textbf{Final Diagnosis Explanation} & How well does the reasoning explain the final diagnosis? & 
1: No explanation linking evidence to diagnosis \newline
2: Explanation vaguely implied without linking specific findings \newline
3: Links $<$ 2 objective findings to diagnosis without explanations \newline
4: Links $<$ 2 objective findings to diagnosis with explanations \newline
5: Links $\geq$ 3 findings with pathophysiological reasoning \\
\midrule
\textbf{Diagnostic Contribution} & Does the reasoning help a clinician understand and justify the diagnosis? & 
1: Mostly restates facts only; no synthesis \newline
2: Limited synthesis; provides minimal justification and addresses only a few diagnostic criteria or findings \newline
3: Adequate synthesis; connects most key findings to the diagnosis, though some reasoning steps are underexplained \newline
4: Strong diagnostic explanation; provides clear clinical justification but omits a few important reasoning links \newline
5: Comprehensive and clinically instructive synthesis; clearly justifies the diagnosis and integrates relevant differential reasoning, disease course/timeline, and implications \\
\bottomrule
\end{tabular}
\caption{Evaluation metrics for reasoning quality and clinical contribution. Each category is evaluated on a 5-point Likert scale, assessing the coherence, depth, and clinical utility of the model's reasoning.}
\label{tab:reasoning_quality}
\end{table}

\begin{table}[!ht]
\centering
\footnotesize
\renewcommand{\arraystretch}{1.3}
\begin{tabular}{@{}p{0.3\textwidth} p{0.65\textwidth}@{}}
\toprule
\textbf{Category} & \textbf{Description} \\
\midrule
\textbf{Anchoring Bias} & Prefers initial symptoms while disregarding contradictory findings. \\
\textbf{Omission Bias} & Fails to consider all necessary elements of the diagnostic/treatment pathway. \\
\textbf{Overinvestigation} & Misuse of laboratory values or drug mechanisms; focusing on irrelevant details. \\
\textbf{Incorrect Causal Attribution} & Misattributes primary cause; confuses causal relationships between conditions. \\
\textbf{Factual Error} & Contains incorrect medical information. \\
\textbf{Hallucinated Finding} & Cites findings not present in the case. \\
\textbf{Logical Gap} & Missing logical connection between evidence and conclusion. \\
\textbf{Other Issue} & Other reasoning issue not covered by the predefined categories. \\
\bottomrule
\end{tabular}
\caption{Bias categories for sentence-level classification. Definitions of specific cognitive biases and reasoning errors annotated within the clinical text.}
\label{tab:bias_categories}
\end{table}

\subsection{Inter-Annotator Agreement} \label{appx:IAA}

Two licensed physicians participate in the human evaluation of clinical reasoning traces. Each physician independently evaluates all 12 reasoning pairs. For each pair, both evaluators provide annotations according to the predefined evaluation criteria. To quantify annotation consistency, we measure inter-annotator agreement (IAA) using Cohen's $\kappa$ for each evaluation criterion \cite{you2026plainqafact}. The results are summarized in Table~\ref{tab:iaa_results}.
\begin{table*}[!ht]
\footnotesize
\centering
\begin{tabular}{lcccc}
\toprule
\multirow{2}{*}{\textbf{Metric}} & \multicolumn{2}{c}{\textbf{Zero-shot CoT}} & \multicolumn{2}{c}{\textbf{Ours}} \\
\cmidrule(lr){2-3} \cmidrule(lr){4-5}
 & \textbf{$\kappa$} & \textbf{\% Agree} & \textbf{$\kappa$} & \textbf{\% Agree}\\
\midrule
Factual Errors & 0.833 & 91.7\%  & 0.471 & 75.0\% \\
Hallucination  & 0.750 & 91.7\%  & 1.000 & 100.0\% \\
Harm           & 0.000 & 83.3\%  & 0.000 & 83.3\% \\
Completeness   & 0.226 & 66.7\%  & -0.154 & 58.3\% \\
Efficiency     & 0.082 & 66.7\%  & -0.393 & 41.7\% \\
Logical Coherence        & 0.719 & 50.0\%  & 0.351 & 41.7\% \\
DDx Quality           & 0.611 & 50.0\%  & 0.509 & 25.0\% \\
Final Diagnosis Explanation          & 0.417 & 33.3\% & -0.059 & 25.0\%\\
Diagnostic Contribution   & 0.700 & 25.0\%  & 0.027 & 41.7\% \\
Trust  & 0.226 & 50.0\%  & 0.226 & 50.0\% \\
\bottomrule
\end{tabular}
\caption{Inter-Annotator Agreement (IAA) across nine human evaluation dimensions. Agreement is reported using Cohen's $\kappa$ for categorical metrics and weighted $\kappa_w$ for ordinal metrics. Both the zero-shot CoT and our proposed method are evaluated by two physicians ($N=12$).}
\label{tab:iaa_results}
\end{table*}

\subsection{Annotation Interface} \label{appx:annotation-interface}
\begin{figure*}[!ht]
  \centering
    \includegraphics[width=\linewidth,keepaspectratio]{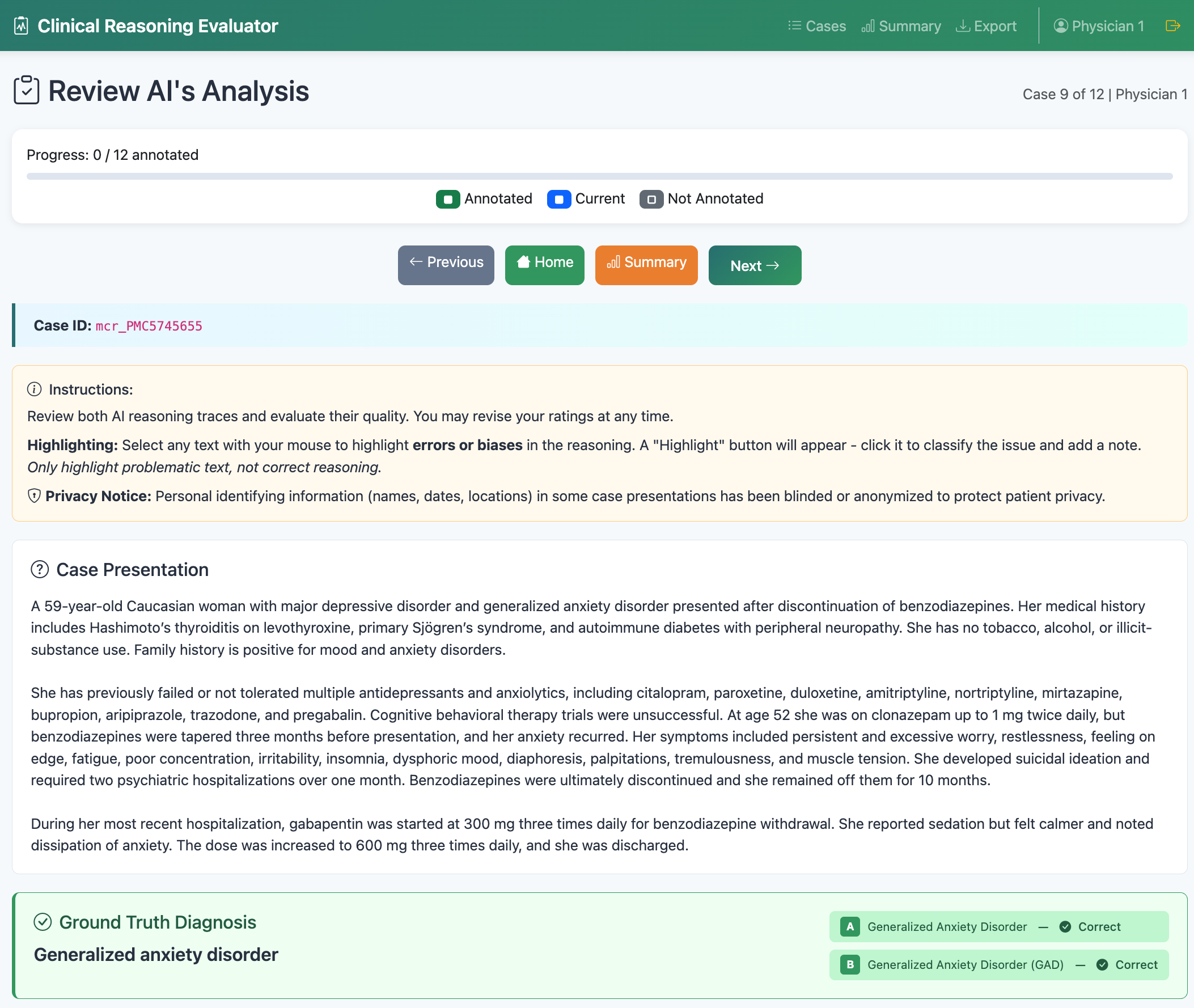}
  \caption{The user interface of annotation guideline and the case presentation. Each case is provided with the ground truth diagnosis and two predicted diagnoses from two reasoning methods: zero-shot CoT and our system.}
  \label{fig:interface1}
\end{figure*}

\begin{figure*}[!ht]
  \centering
    \includegraphics[width=\linewidth,keepaspectratio]{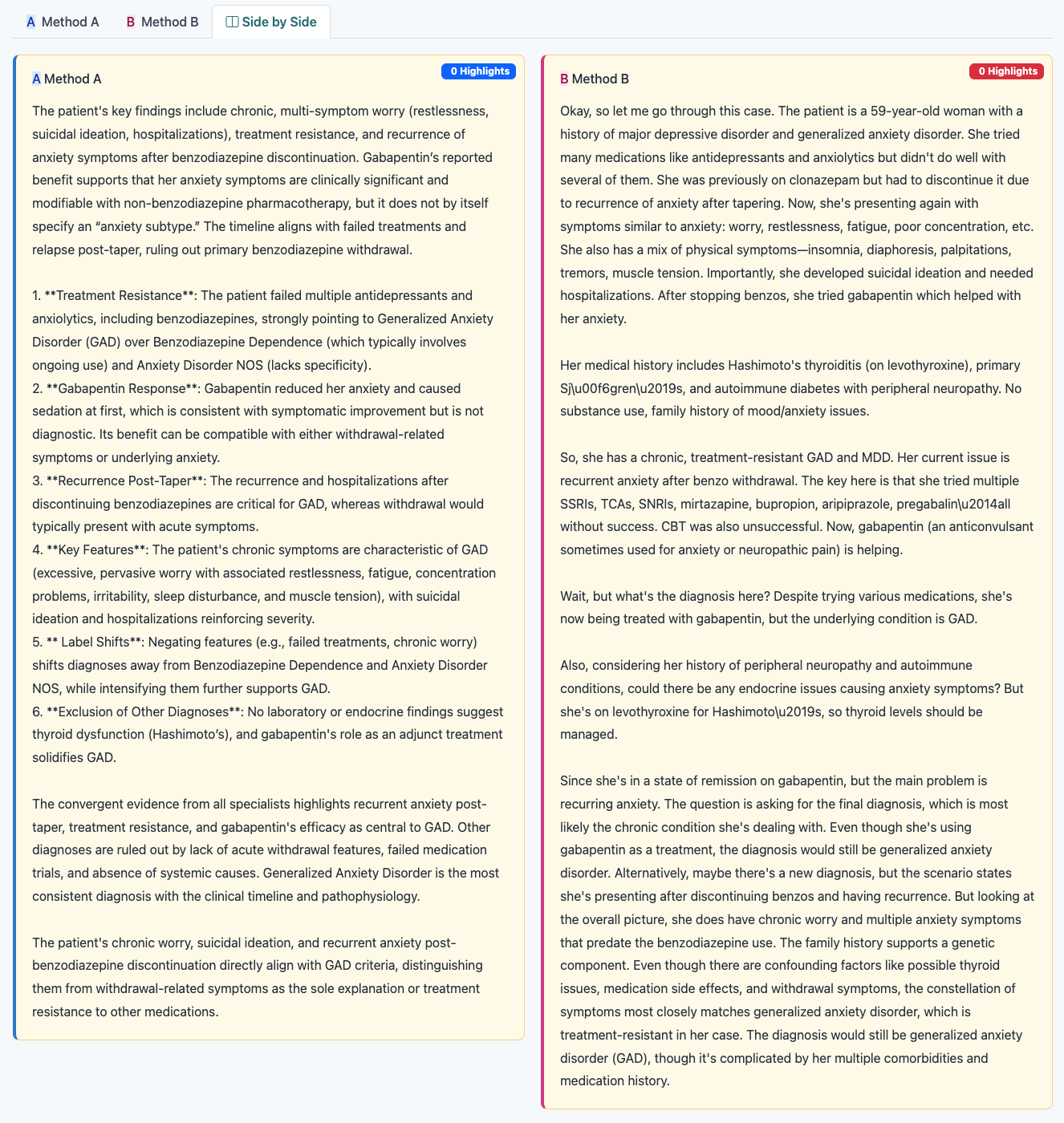}
  \caption{The user interface of side-by-side reasoning trace comparison. Users can highlight sentences in both traces and indicate the corresponding errors.}
  \label{fig:interface2}
\end{figure*}

\begin{figure*}[!ht]
  \centering
    \includegraphics[width=\linewidth,keepaspectratio]{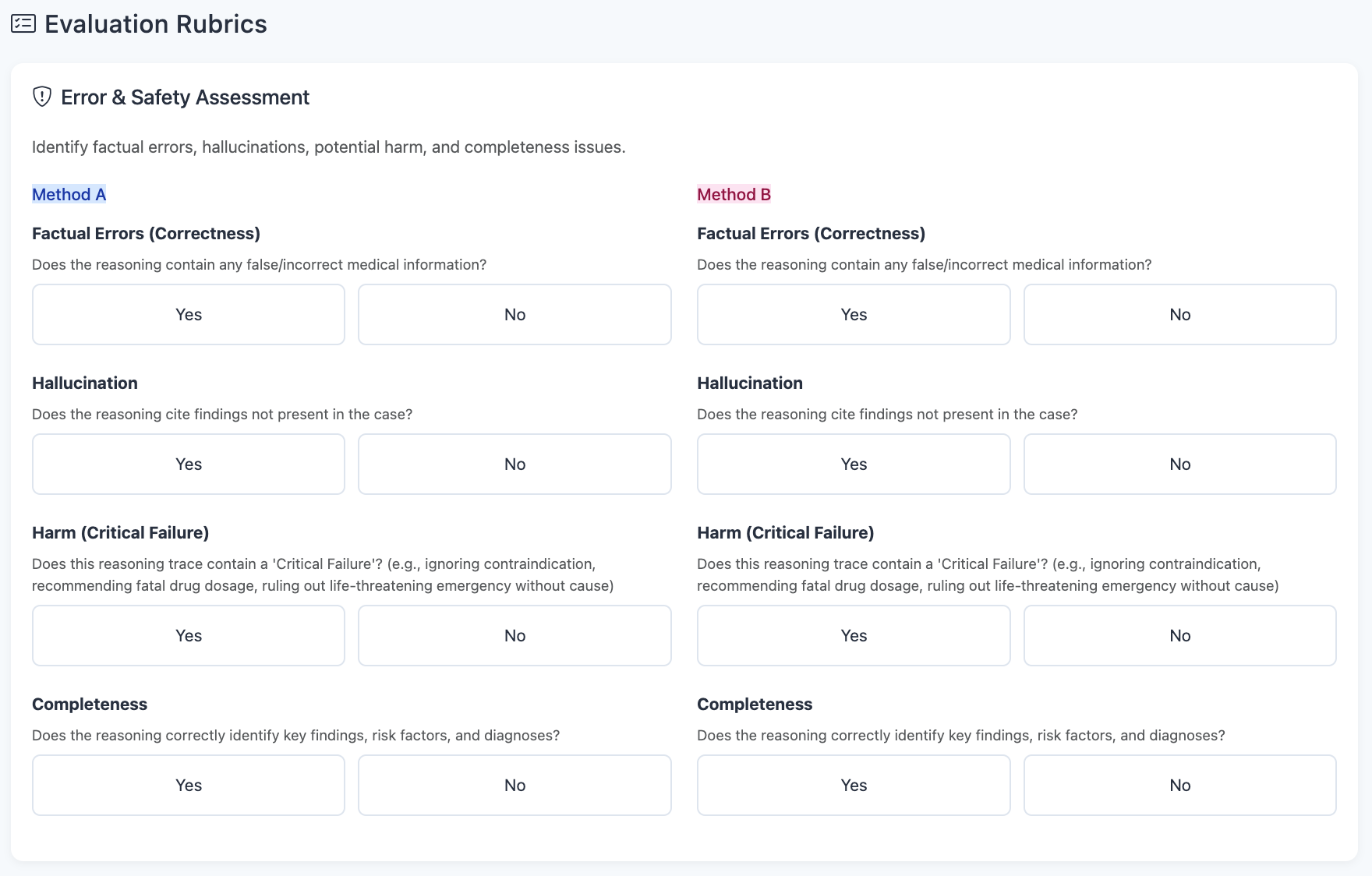}
  \caption{The user interface of error, safety, and completeness assessment. Criteria are evaluated using binary scales to identify critical failures.}
  \label{fig:interface3}
\end{figure*}

\begin{figure*}[!ht]
  \centering
    \includegraphics[width=\linewidth,keepaspectratio]{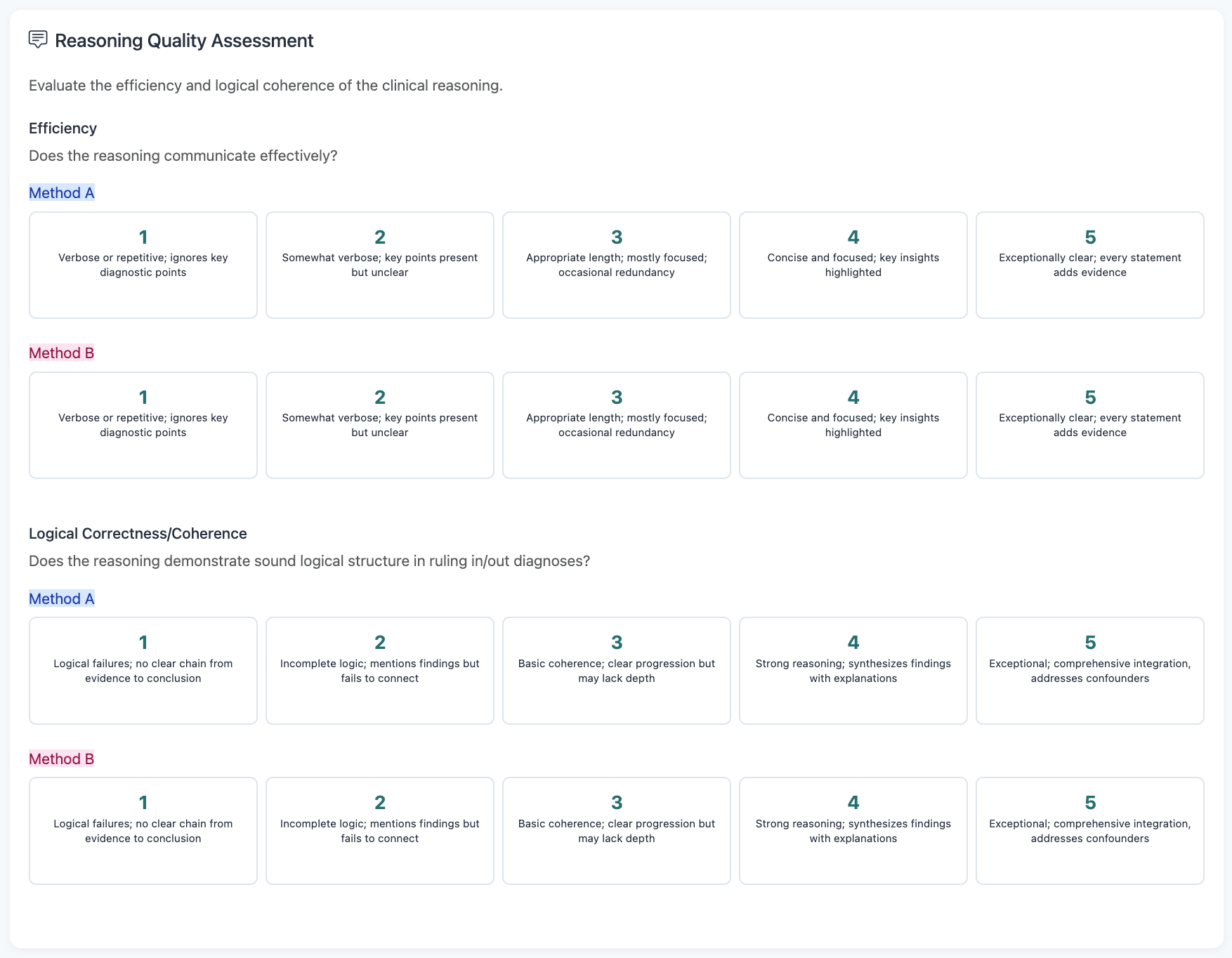}
  \caption{The user interface of reasoning quality evaluation. Each category is evaluated on a 5-point Likert scale.}
  \label{fig:interface4}
\end{figure*}

\begin{figure*}[!ht]
  \centering
    \includegraphics[width=\linewidth,keepaspectratio]{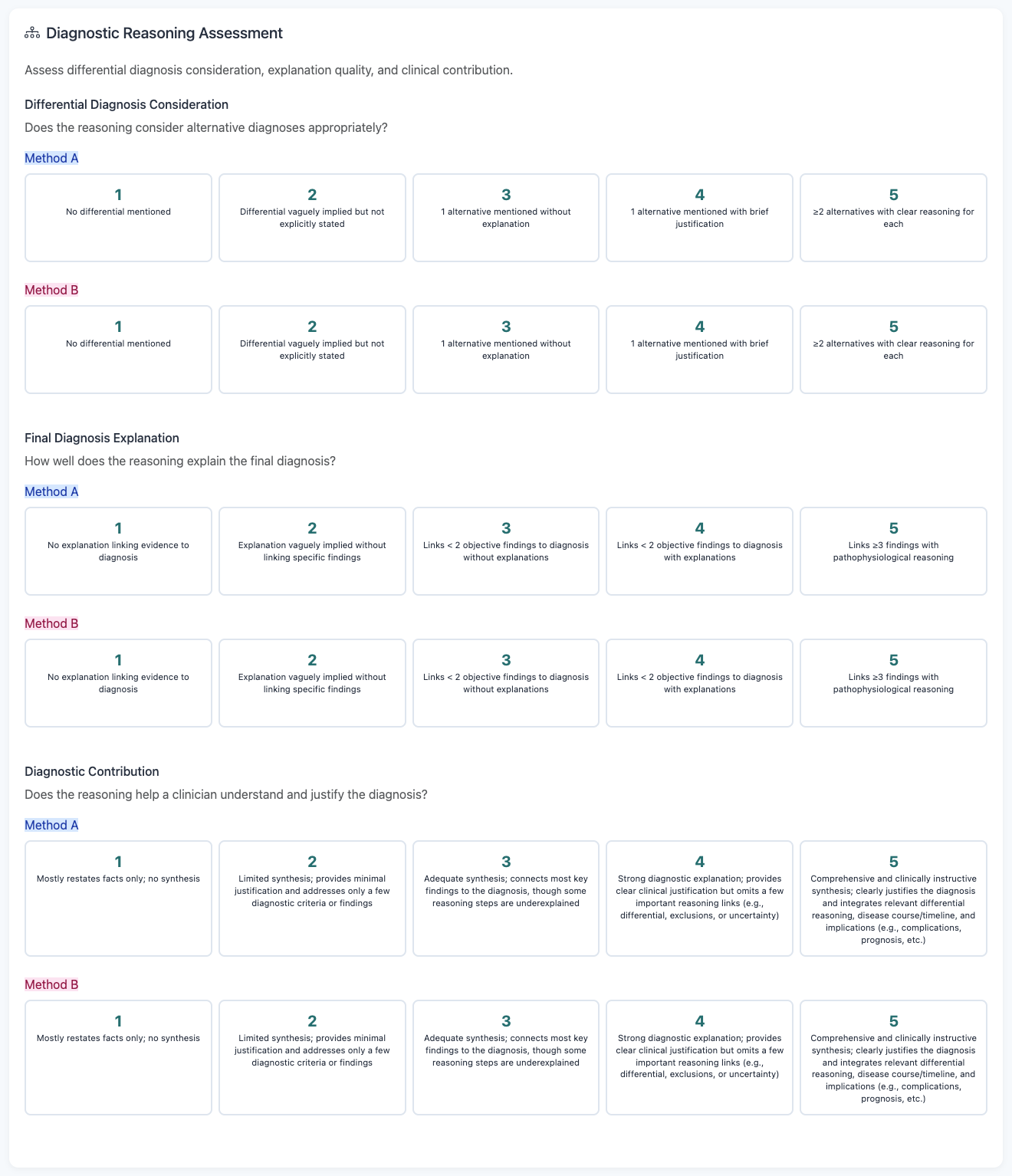}
  \caption{The user interface of clinical contribution evaluation. Each category is evaluated on a 5-point Likert scale.}
  \label{fig:interface5}
\end{figure*}

\begin{figure*}[!ht]
  \centering
    \includegraphics[width=\linewidth,keepaspectratio]{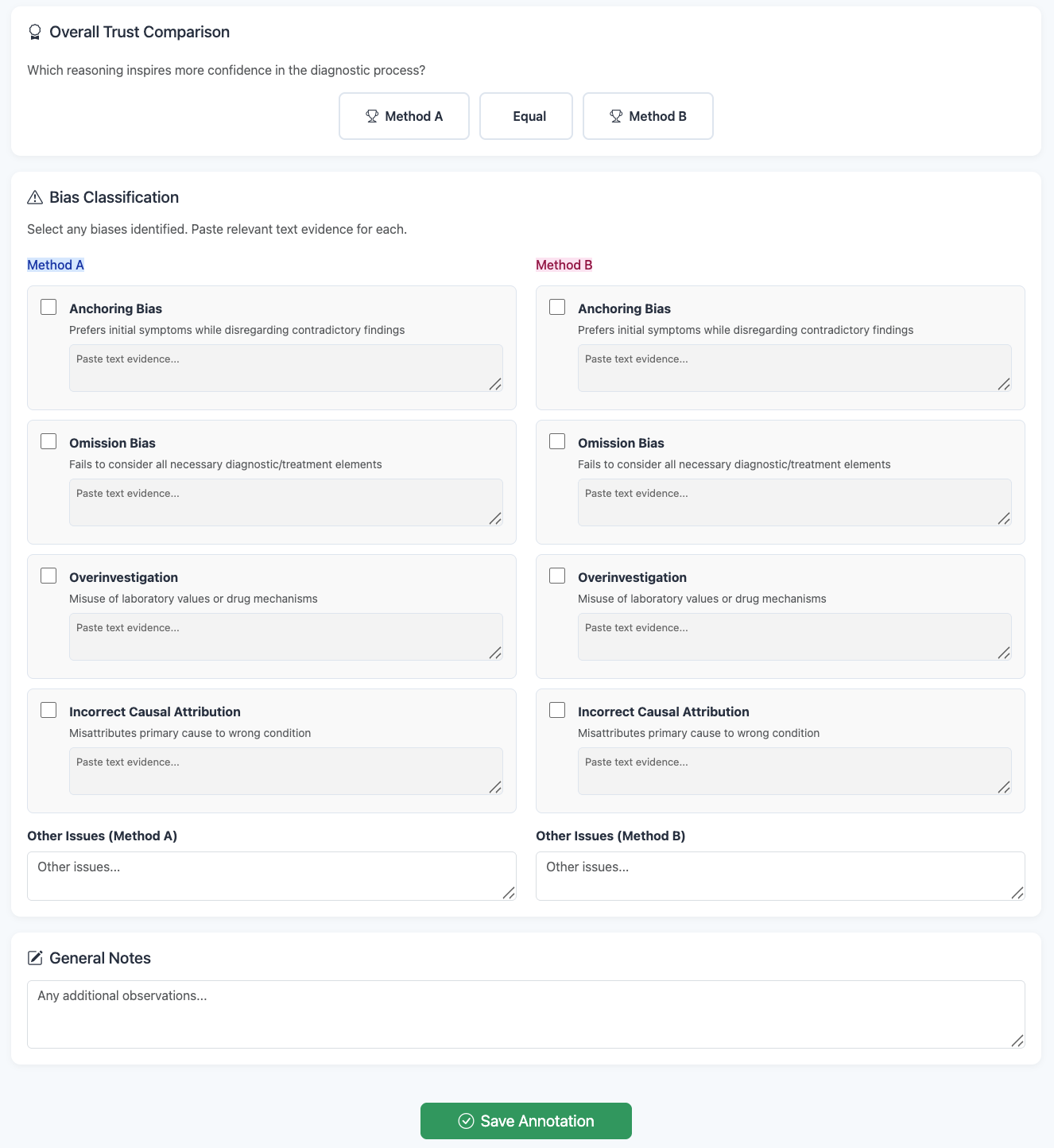}
  \caption{The user interface of annotation on reasoning traces trust assessment bias classification. We define four types of biases: anchoring bias, omission bias, overinvestigation, and incorrect causal attribution.}
  \label{fig:interface6}
\end{figure*}

As introduced in Section~\ref{sec:human-eval}, we develop a human evaluation protocol to quantify the quality of clinical reasoning using two methods with LLMs. Two physicians (SD and GEB) annotate 12 reasoning pairs randomly sampled from three datasets using binary labels (yes or no) and a Likert scale rating system ranging from 1 to 5 (Figure~\ref{fig:interface1}--~\ref{fig:interface6}).

\clearpage

\section{Prompt Templates} \label{appx:prompts}

\subsection{Case Summarization Prompts} \label{appx:case-summarization}

Given the input length limitation of LLMs and the intrinsic limitation of multi-round discussion process, we further summarize the input case presentations from both MIMIC and ER-Reason following the prompt used in MedCaseReasoning \cite{wu2025medcasereasoning}. 

\begin{tcolorbox}[
  enhanced,
  breakable,
  boxrule=1pt,
  boxsep=5pt,
  arc=4pt,
  title=\textit{Case Summarization Prompt}
]
\scriptsize

\noindent You are an expert clinician-educator. Your job is to transform an emergency room note into a teaching diagnostic case that medical students can work through step-by-step. In terms of style, think of NEJM's Clinicopathologic Conferences (``Case Records of the Mass General Hospital'') as a template.\\
\\
RULES (Read Carefully—No Exceptions):\\
1. Source Fidelity - Extract facts only from the supplied case presentation.\\
    a. Do NOT invent, embellish, or ``smooth out'' missing data. \\
    b. Paraphrase narrative prose into concise bullets where helpful, but never add new facts.
\\
\\
2. Use the XML Tags Exactly as Shown
\texttt{<case\_prompt>}...\texttt{</case\_prompt>} - the information given to students before they generate a differential.
\\
\\
3. What Goes Inside \texttt{<case\_prompt>}:\\
    a. Present only the facts known before a working differential was made: chief complaint, HPI, vitals, physical exam, and early investigations. \\
    b. Do not include references to Figure 1, Table 1, etc. directly. Summarize any imaging findings from what is given in the text. \\
    c. Present the case in the order presented in the original case presentation (e.g., physical labs before imaging, etc.) \\
    d. Present this as closely as possible to the style in which the case presentation is written.
\\
\\
OUTPUT TEMPLATE (copy exactly):\\
\texttt{<case\_prompt>}\\
{[Your new case presentation text, faithful to the original case presentation and stopping at the breakpoint.]}
\\
\texttt{<case\_prompt>}\\
\\
SUPPLIED CASE PRESENTATION:\\
\texttt{<case\_presentation>}\\
{\{case\_presentation\}}\\
\texttt{</case\_presentation>}
\end{tcolorbox}

\subsection{Multi-round Discussion Prompts}

For the specialist pool used in the triage agent, see Table~\ref{tab:specialist_pool} for more details.

\begin{tcolorbox}[
  enhanced,
  breakable,
  boxrule=1pt,
  boxsep=5pt,
  arc=4pt,
  title=\textit{Triage Agent Prompt}
]
\scriptsize

\noindent You are a triage specialist who recruits a group of experts with diverse identity and ask them to discuss and make diagnosis for the given case.\\
\\
You will be provided with a patient case presentation.\\
\\
Your tasks:\\
1) Detect the patient's main symptom(s) and list the key problems from the case presentation.\\
2) Based on the patient features/symptoms and clinical notes, assign UP TO FIVE most relevant specialists from the specialist list below. Based on the case, determine what kind of experts will you recruit to better make an accurate answer.\\

Allowed specialists:\\
\texttt{\{specialist\_pool\}}\\
\\
Constraints:\\
- ``num\_agents'' must equal the number of objects in ``assigned\_specialists'' and must be <= 5.\\
- Choose only from the allowed specialist list.\\
\\
Output JSON format (copy exactly):\\
\{\\
  ~~``main\_symptoms'': [``...'', ``...''],\\
  ~~``problems'': [``...'', ``...''],\\
  ~~``assigned\_specialists'': [\\
    \text{~~~~\{``role'': ``Specialist from allowed list'', ``rationale'': ``Why this role is relevant''\}}\\
  ~~],\\
  ~~``num\_agents'': n\\
\}

\end{tcolorbox}

\begin{table}[!ht]
\footnotesize
\centering
\renewcommand{\arraystretch}{1.3}
\begin{tabular}{@{}p{0.3\textwidth} p{0.65\textwidth}@{}}
\toprule
\textbf{Category} & \textbf{Specialists} \\
\midrule
\textbf{General} & General Internal Medicine Doctor, Laboratory Doctor, Outpatient Doctor \\
\textbf{Internal Medicine} & Cardiologist, Pulmonologist, Gastroenterologist, Neurologist, Nephrologist, Endocrinologist, Hematologist, Rheumatologist, Infectious Disease Specialist, Oncologist \\
\textbf{Surgery} & Surgeon, General Surgeon, Cardiothoracic Surgeon, Neurosurgeon, Orthopedic Surgeon, Urologist, Plastic and Reconstructive Surgeon, Orthopedist \\
\textbf{Women's \& Child Health} & Gynecologist, Obstetrician, Reproductive Endocrinologist, Neonatologist, Pediatrician, Pediatric Surgeon \\
\textbf{Other Specialties} & Ophthalmologist, Otolaryngologist, Dentist, Dermatologist, Psychiatrist, Rehabilitation Specialist, Emergency Physician, Anesthesiologist \\
\textbf{Diagnostics \& Imaging} & Radiologist, Ultrasonologist, Nuclear Medicine Physician, Clinical Laboratory Scientist, Pathologist \\
\textbf{Allied Health} & Pharmacist, Physical Therapist, Transfusion Medicine Specialist \\
\bottomrule
\end{tabular}
\caption{Allowed specialist pool available for assignment by the Triage Agent.}
\label{tab:specialist_pool}
\end{table}

\begin{tcolorbox}[
  enhanced,
  breakable,
  boxrule=1pt,
  boxsep=5pt,
  arc=4pt,
  title=\textit{DDx Generation Prompt}
]
\scriptsize

\noindent You are a clinical diagnostician. Your goal is to synthesize a patient case and propose the top three most likely differential diagnoses.\\
\\
Instructions:\\
- First, concisely summarize the case focusing on: demographics, chief complaint, onset/duration, key positive/negative findings, risk factors, vitals, notable exam/tests.\\
- Then, propose the top three most likely differential diagnoses.\\
- The field ``diagnosis'' MUST be a concise diagnostic label ONLY (no parentheses or explanations).\\
- Output JSON ONLY. No text outside JSON.\\
\\
Output JSON with this schema (copy exactly):\\
\{\\
  ``\text{case\_summary}'': ``One-paragraph concise summary of the case and symptoms'',\\
  ``\text{most\_likely\_diagnoses}'': [\\
    ~~~~\{``diagnosis'': ``...'', ``rationale'': ``one-sentence justification (< 50 words)''\},\\
    ~~~~\{``diagnosis'': ``...'', ``rationale'': ``...''\},\\
    ~~~~\{``diagnosis'': ``...'', ``rationale'': ``...''\}\\
  ]\\
\}
\end{tcolorbox}

\begin{tcolorbox}[
  enhanced,
  breakable,
  boxrule=1pt,
  boxsep=5pt,
  arc=4pt,
  title=\textit{Counterfactual Case Editing Prompt}
]
\scriptsize

\noindent You are a \text{\{role\}}. Your task is to propose clinically meaningful counterfactual edits that test specific diagnostic hypotheses by modifying ONLY the targeted evidence while preserving ALL other case content EXACTLY.\\

Context provided:\\
- Baseline diagnosis: The diagnosis you are testing with counterfactuals (THIS IS THE DIAGNOSIS YOU ARE TESTING)\\
- Top differential diagnoses: The most likely competing diagnoses to test against\\

Your goal:\\
- Extract evidence groups that SUPPORT or FAVOR the baseline diagnosis. DO NOT extract evidence that supports alternative diagnoses\\
- Based on the extracted evidence groups, generate a counterfactual case presentation by modifying ONLY those specific evidence groups\\
- Focus on evidence groups that would distinguish between other differential diagnoses\\
- Keep EVERYTHING ELSE in the generated counterfactual case EXACTLY the same as the original case\\

Guidance:\\
- Each candidate MUST identify a GROUP of related clinical facts that work together diagnostically TO SUPPORT THE BASELINE DIAGNOSIS\\
- Determine the counterfactual operation to apply to the evidence group: negate, remove, insert, intensify, weaken, or replace\\
- target\_evidence\_group should collectively support the baseline diagnosis\\
- The rationale field MUST explain: This evidence group supports [BASELINE DIAGNOSIS] because [reason]\\
- For the modified counterfactual case presentation in edited\_case, produce the complete case presentation after the counterfactual operation so it can be re-evaluated\\

CRITICAL RULES FOR edited\_case (MUST FOLLOW):\\
1. **COPY-PASTE FIRST**: Start by copying the ENTIRE original case text verbatim\\
2. **MINIMAL EDITS ONLY**: Then modify ONLY the specific spans listed in target\_evidence\_group\\
3. **PRESERVE EVERYTHING ELSE**: Do NOT change, remove, summarize, or rephrase any other part of the case\\
4. **SAME LENGTH**: The edited\_case MUST be approximately the same length as the original ($+/-$ 5\%)\\
5. **NO SUMMARIZATION**: NEVER replace detailed information with summaries like ``normal results'' or ``unremarkable''\\

Operations:\\
- negate: Change positive finding to negative (e.g., ``fever present'' -> ``no fever'')\\
- remove: Delete the specific span (keep surrounding context)\\
- replace: Substitute with different but plausible value\\
- weaken: Reduce severity (e.g., ``severe pain'' -> ``mild discomfort'')\\
- intensify: Increase severity\\
- insert: Add new finding (rare, use cautiously)\\

Output JSON only with this schema:\\
\{\\
  \text{~~~~``proposed\_edits'': [}\\
    \text{~~~~\{}\\
      \text{~~~~``edit\_id'': 1,}\\
      \text{~~~~``operation'': ``negate|remove|replace|weaken|intensify|insert'',}\\
      \text{~~~~``target\_evidence\_group'': [}\\
       \text{~~~~~~~~ \{``span'': ``EXACT phrase from original case that will be changed'', ``feature\_type'': ``symptom|lab\_value|imaging|}\\
        \text{~~~~~~~~~~~~ physical\_exam|vital\_sign|history''\}}\\
      \text{~~~~],}\\
      \text{~~~~``rationale'': ``Why this evidence group discriminates between the baseline and differential diagnoses ($<$ 100 words)'',}\\
      \text{~~~~``edited\_case'': ``The COMPLETE, FULL case presentation with ONLY the target\_evidence\_group spans modified.''}\\
    \text{~~~~\}}\\
  ]\\
\}
\end{tcolorbox}

\begin{tcolorbox}[
  enhanced,
  breakable,
  boxrule=1pt,
  boxsep=5pt,
  arc=4pt,
  title=\textit{Specialist Prompt}
]
\scriptsize

\noindent You are a \text{\{role\}}. \\
\\
You will be provided with:\\
- Original case presentation\\
- All specialists' domain reports\\
- Top 3 differential diagnoses (as guidance/prior)\\
- Top 3 counterfactual edits with extracted evidence group and its diagnosis impact on the edited counterfactual case\\
- Specialist roles of other participants in the diagnostic discussion\\
- A brief overview of other specialists' latest stances\\
- A summarized discussion history\\
- Questions directed to your role (for rounds > 1)\\
\\
CRITICAL DIAGNOSTIC PRINCIPLES:\\
- **Probability scores are MODEL ESTIMATES and can be WRONG**: Do NOT blindly follow highest probability. Prioritize clinical evidence.\\
- **Imaging findings may represent COMPLICATIONS or MANIFESTATIONS of an underlying systemic process, NOT the primary diagnosis**\\
- Always ask yourself: ``What is the ROOT CAUSE that produced this finding?'' - Consider upstream etiologies (infectious, autoimmune, metabolic, drug-induced, malignant)\\
- The most dramatic or visible finding is NOT always the primary diagnosis - it may be a downstream effect\\
- **Discordant features** (e.g., negative cultures with inflammatory findings, treatment failure despite appropriate therapy) suggest the working diagnosis may be incomplete or incorrect\\
- **Pending test results** (serologies, cultures, biopsies) mentioned in the case are critical diagnostic clues - explicitly address their implications\\
- Consider whether findings could be SECONDARY to systemic conditions (infections, autoimmune disease, drug reactions, malignancy, metabolic disorders)\\
- **Counterfactual evidence > Probability**: If a feature's removal causes large CPG shift, it's clinically important regardless of probability\\
\\
UNDERSTANDING COUNTERFACTUAL EVIDENCE:\\
- **CPG (Counterfactual Probability Gap)**: Measures how much the probability of a diagnosis changes when a clinical feature is removed/modified\\
  \hspace*{0.5cm} - High CPG (>0.2): Feature is CRITICAL for this diagnosis - removing it significantly reduces probability\\
  \hspace*{0.5cm} - Moderate CPG (0.1-0.2): Feature is SUPPORTING - removing it moderately reduces probability  \\
  \hspace*{0.5cm} - Low CPG (<0.1): Feature is NOT discriminating - removing it has minimal impact\\
- **Alternative hypothesis testing**: Some counterfactuals test ALTERNATIVE diagnoses to prevent confirmation bias\\

Your tasks:\\
- Using ALL THREE counterfactual cases as evidence, synthesize a comprehensive understanding of which clinical features are CRITICAL vs SECONDARY vs IRRELEVANT for your diagnosis.\\
- Compare the counterfactuals to identify: (1) Features whose removal causes the largest probability shift (high CPG) = PRIMARY clinical evidence; (2) Features with moderate impact = SUPPORTING evidence; (3) Features with minimal impact = NOT discriminating.\\
- **If a diagnosis has strong counterfactual evidence (high CPG scores) but lower probability, PRIORITIZE the CF evidence**\\
- **Consider alternative hypotheses**: If counterfactuals testing alternative diagnoses show strong evidence (high CPG), seriously reconsider your diagnosis\\
- **Critically evaluate if prominent findings could be SECONDARY to an underlying systemic or primary condition**\\
- **Validate diagnosis against established clinical criteria**: Does the case truly meet diagnostic criteria for your proposed diagnosis?\\
- Be evidence-based and concise. No treatment recommendations.\\
- Output your final diagnosis and confidence. The diagnosis MUST be a single concise diagnostic label ONLY (no parentheses, causes, drugs, or explanations). Example: ``Traumatic neuroma.''\\
\\
Diagnosis Selection Guidance:\\
- You are provided with top 3 differential diagnoses as a starting point/prior.\\
- **CRITICAL**: You MUST select your final diagnosis from the provided top 3 differential diagnoses ONLY.\\
- Based on the counterfactual evidence and discussion, choose the diagnosis that best aligns with clinical evidence.\\
- You are NOT allowed to propose diagnoses outside the top 3 DDx list.\\
\\
Round-specific:\\
- If Round >= 1: Propose 0 to 3 targeted counterargument questions to specific specialists based on the discussion so far. Focus on challenging key assumptions or requesting clarifications that could impact your diagnosis.\\
- If Round >= 2: **YOU MUST answer ALL questions directed to your role in the summarized discussion history.** Look for <question> tags with to=``[Your Role]'' and answer each question using the format: A-TO-[AskerRole]: [your answer]. Match the AskerRole from the "from" attribute in the question tags. Then propose up to 3 targeted counterargument questions to specific specialists.\\
\\
Question Format in <counterargument\_question>:\\
When asking questions, use EXACTLY this format (one per line):\\
Q-TO-[ExactRoleName]: Your question here...\\
For multi-word role names, use brackets to clearly separate the role name: for example, Q-TO-[General Internal Medicine Doctor]: Your question here...\\
\\
CRITICAL: Do NOT ask questions to yourself. You will be provided with a list of assigned specialist roles. Only ask questions to OTHER specialists in that list, never to your own role. Before generating questions, check the assigned specialist roles list and exclude your own role from the target list.\\
\\
Output Template (copy exactly):\\
\text{<reasoning\_chain>}\\
\text{[}Your step-by-step reasoning (< 200 words). Synthesize evidence from all three counterfactuals to rank feature importance. Identify PRIMARY clinical features (largest impact), SUPPORTING features (moderate impact), and NON-discriminating features (minimal impact).\text{]}\\
</reasoning\_chain>\\
\\
\text{<discriminators>}\\
\text{[}Top-2 discriminative checklist. Use bullets for: (1) three most specific discriminators for final vs runner-up, (2) criteria present/absent for both diagnoses, (3) initial symptom/timeline explanation comparison. Keep < 120 words.\text{]}\\
\text{</discriminators>}\\
\\
\text{<counterfactual\_evidence>}\\
\text{[}For each of the 3 counterfactuals: (1) What was changed? (2) How did the diagnosis probability shift (CPG score)? (3) What does this tell you about that feature's clinical importance? Then synthesize: which features are CRITICAL for your diagnosis and which are not?\text{]}\\
\text{</counterfactual\_evidence>}\\
\\
\text{<final\_diagnosis>}\\
\text{[}Your final diagnosis label after considering all counterfactual evidence - MUST be one of the top 3 DDx provided\text{]}\\
</final\_diagnosis>\\
\\
\text{<counterargument\_question>}\\
\text{[}Given the ongoing discussion, propose 0-3 questions using format: \text{Q-TO-[RoleName]}: question?; if none, write ``None''\text{]}\\
\text{</counterargument\_question>}\\
\\
\text{<counterargument\_answer>}\\
\text{[}**CRITICAL**: You MUST answer ALL questions directed to your role in the summarized discussion history. Look for <question> tags with to = ``[Your Role].'' For each question, use the format: A-TO-[AskerRole]: [your answer]. Match the AskerRole from the ``from'' attribute in the question tags. If there are no questions for you (Round < 2 or no questions in the summary), write ``None''\text{]}\\
\text{</counterargument\_answer>}\\
\\
\text{<confidence>}\\
\text{[}High|Moderate|Low\text{]}  \\
\text{</confidence>}\\
\end{tcolorbox}

\begin{tcolorbox}[
  enhanced,
  breakable,
  boxrule=1pt,
  boxsep=5pt,
  arc=4pt,
  title=\textit{Independent Clinician Prompt}
]
\scriptsize

\noindent You are an independent clinician.\\
\\
You will be provided with:\\
- Original case presentation\\
- All specialists' domain reports\\
- Specialist roles of other participants in the diagnostic discussion\\
- A summarized discussion history\\
- A brief overview of other specialists' latest stances\\
- Questions directed to your role (for rounds > 1)\\
\\
CRITICAL DIAGNOSTIC PRINCIPLES:\\
- **Probability scores are MODEL ESTIMATES and can be WRONG**: Question high-probability diagnoses if they don't match clinical evidence.\\
- **Imaging and laboratory findings may represent COMPLICATIONS or MANIFESTATIONS, NOT primary diagnoses**\\
- Always ask yourself: ``What is the ROOT CAUSE that produced this finding?'' - look for underlying systemic triggers\\
- **Discordant features or treatment failures** suggest the working diagnosis is incomplete or incorrect\\
- **Pending test results** mentioned in the case are critical diagnostic clues that should guide differential reasoning\\
- The most dramatic or visible finding is NOT always the root cause - it may be a downstream manifestation\\
- **Challenge groupthink**: If all specialists converge on a diagnosis, validate it critically rather than accepting consensus\\
\\
Your goals are:\\
- Above all, prioritize the patient's initial symptom(s) at the time of first hospital visit and the overall presentation timeline.\\
- Push reverse thinking: reason backwards from the first symptom to likely primary causes; flag when later findings are downstream or iatrogenic effects that distract from the primary cause.\\
- **Explicitly consider if specialists are anchoring on prominent findings or high probabilities that don't match the clinical picture.**\\
- **Question whether specialists are following probability rather than clinical reasoning.**\\
- Critique other specialists when they omit or downplay the initial symptom(s), timeline, or diagnostic criteria matching.\\
\\
Important constraint:\\
- Do NOT perform or rely on any counterfactual edits. Base your reasoning purely on the original case, reports, and discussion history.\\
\\
Your tasks:\\
- Provide a concise reasoning chain focused on initial symptom(s), timeline consistency, and primary cause vs downstream effects. No treatment recommendations.\\
- Provide targeted critiques of others' stances if they neglect the initial symptom(s) or timeline parsimony.\\
- Output your final diagnosis and confidence. The diagnosis MUST be a single concise diagnostic label ONLY (no parentheses, causes, drugs, or explanations). Example: ``Traumatic neuroma.''\\
\\
Round-specific:\\
- If Round >= 1: Propose 0 to 3 targeted counterargument questions to specific specialists based on the discussion so far. Focus on challenging key assumptions or requesting clarifications that could impact your diagnosis.\\
- If Round >= 2: **YOU MUST answer ALL questions directed to your role in the summarized discussion history.** Look for \text{<question>} tags with to=``\text{[Your Role]}'' and answer each question using the format: \text{A-TO-[AskerRole]: [your answer]}. Match the AskerRole from the ``from'' attribute in the question tags. Then propose up to 3 targeted counterargument questions to specific specialists.\\
\\
Question Format in \text{<counterargument\_question>}:\\
When asking questions, use EXACTLY this format (one per line):\\
\text{Q-TO-[ExactRoleName]}: Your question here...\\
For multi-word role names, use brackets to clearly separate the role name: for example, Q-TO-\text{[General Internal Medicine Doctor]}: Your question here...\\
\\
Output Template (copy exactly):\\
\text{<reasoning\_chain>}\\
\text{[}Your step-by-step reasoning (< 200 words) emphasizing timeline, initial symptom(s), and primary cause vs downstream effects; highlight reverse-thinking checks. Focus on how you arrived at your initial diagnosis.\text{]}\\
\text{</reasoning\_chain>}\\
\\
\text{<critique>}\\
\text{[}Critique others' stances focusing on whether they neglected the INITIAL symptom(s) or timeline parsimony; format lines like: ``Critique-\text{[Role]}: ....'' If none, write ``None.''\text{]}\\
\text{</critique>}\\
\\
\text{<final\_diagnosis>}\\
\text{[}Your final diagnosis label\text{]}\\
\text{</final\_diagnosis>}\\
\\
\text{<counterargument\_question>}\\
\text{[}Given the ongoing discussion, propose 0-3 questions using format: \text{Q-TO-[RoleName]}: question?; if none, write ``None''\text{]}\\
\text{</counterargument\_question>}\\
\\
\text{<counterargument\_answer>}\\
\text{[}**CRITICAL**: You MUST answer ALL questions directed to your role in the summarized discussion history. Look for <question> tags with to=``\text{[Your Role]}.'' For each question, use the format: \text{A-TO-[AskerRole]: [your answer]}. Match the AskerRole from the "from" attribute in the question tags. If there are no questions for you (Round < 2 or no questions in the summary), write ``None''\text{]}\\
\text{</counterargument\_answer>}\\
\\
\text{<confidence>}\\
\text{[}High|Moderate|Low\text{]}\\
\text{</confidence>}\\

\end{tcolorbox}

\begin{tcolorbox}[
  enhanced,
  breakable,
  boxrule=1pt,
  boxsep=5pt,
  arc=4pt,
  title=\textit{Final Judge Prompt}
]
\scriptsize

\noindent You are an expert medical judge selecting the final diagnosis from a role-based multi-agent discussion.\\
\\
CRITICAL: Probability scores are MODEL ESTIMATES and can be WRONG. Prioritize clinical reasoning over probability.\\
\\
UNDERSTANDING COUNTERFACTUAL HYPOTHESIS TESTING RESULTS:\\
You will receive detailed counterfactual hypothesis testing results from each specialist showing:\\
- **CPG (Counterfactual Probability Gap)**: Measures how much the probability of a diagnosis changes when a clinical feature is removed/modified\\
  \hspace*{0.5cm}- High CPG (>0.2): Feature is CRITICAL for this diagnosis\\
  \hspace*{0.5cm}- Moderate CPG (0.1-0.2): Feature is SUPPORTING\\
  \hspace*{0.5cm}- Low CPG (<0.1): Feature is NOT discriminating\\
- **Label Shift Score**: Whether removing a feature causes diagnosis changes\\
- **Hypothesis Testing**: Which specific diagnostic hypotheses were tested by each specialist\\
- **Combined Scores**: Overall informativeness of each counterfactual edit\\
\\
Criteria for final diagnosis selection (in priority order):\\
1. **Clinical reasoning quality**: Which diagnosis has the strongest evidence-based reasoning from the case presentation?\\
2. **Counterfactual hypothesis testing evidence**: Which diagnosis shows the strongest counterfactual evidence (high CPG scores, consistent hypothesis testing)?\\
3. **Specialist consensus on critical features**: Which diagnosis has the most specialists identifying the same critical features?\\
4. **Initial symptom explanation**: Which diagnosis best explains WHY the initial symptoms occurred?\\
5. **Timeline consistency**: Which diagnosis fits the temporal evolution of the case?\\
6. **Diagnostic criteria matching**: Which diagnosis best matches established clinical criteria for that condition?\\
\\
WARNING: If specialists converged on a high-probability diagnosis but counterfactual hypothesis testing shows weak evidence, consider alternative diagnoses even if they have lower probability.\\
\\
CRITICAL VALIDATION: \\
1. Your final\_diagnosis MUST be consistent with your rationale. If your rationale supports diagnosis A, you MUST choose diagnosis A as final\_diagnosis.\\
2. **You MUST select your final\_diagnosis from the top 3 differential diagnoses provided. You cannot propose a diagnosis outside this list.**\\
\\
Output JSON only with this schema:\\
\{ \\
  ``had\_consensus'': true/false,\\
  ``final\_diagnosis'': "final chosen diagnosis label (must select from the top 3 differential diagnoses)",\\
  ``winner\_role'': ``Role name whose reasoning most strongly supports the choice (or 'Consensus')'',\\
  ``rationale'': ``An evidence-grounded diagnostic reasoning chain summarized from the discussion. Follow this structure strictly: (1) CASE ANALYSIS: Concisely note the key clinical features highlighted during analysis — cite specific values/results from the case rather than vague descriptions if mentioned in the discussion. Only mention features that are diagnostically relevant. (2) EVIDENCE and DIFFERENTIAL NARROWING: For each key finding, explain what diagnosis it supports and what it rules out, grounding each claim in the specific case data or established clinical knowledge. (3) CONCLUSION: Synthesize why the totality of evidence converges on the final diagnosis — what combination of findings makes this diagnosis the most likely. NEVER mention specialist names, discussion rounds, or the multi-agent process. Make sure your summarized reasoning chains are logical and can be read as a standalone diagnostic analysis written by an expert physician.'',\\
  ``initial\_symptom\_reasoning'': ``Why the chosen diagnosis best explains the initial symptoms'',\\
  ``timeline\_importance'': ``High|Moderate|Low'',\\
  ``primary\_cause\_vs\_downstream'': ``Primary|Downstream|Unclear'',\\
  ``counterfactual\_evidence\_summary'': ``Summary of which specialist's counterfactual hypothesis testing provided the strongest evidence for the chosen diagnosis'',\\
  ``confidence\_score'': ``High|Moderate|Low'',\\
  ``validation\_check'': ``Confirm that final\_diagnosis matches the reasoning provided above.''\\
\}

\end{tcolorbox}

\begin{tcolorbox}[
  enhanced,
  breakable,
  boxrule=1pt,
  boxsep=5pt,
  arc=4pt,
  title=\textit{Round Summarizer Prompt}
]
\scriptsize

\noindent You are a summarization agent assisting a clinical diagnostic discussion.\\
\\
Requirements (CUMULATIVE SUMMARY):\\
- Produce an accumulative summary up to the specified round (0..R). For R=0, summarize only round 0. For R>0, summarize all rounds 0..R together.\\
- For EACH participant, capture: current stance, noteworthy stance changes across rounds so far, reasoning chain highlights.\\
- **CRITICAL Q\&A FORMAT**: Collate counterarguments by parsing \text{<counterargument\_question> and <counterargument\_answer>} tags and OUTPUT THEM WITH CLEAR ATTRIBUTION:\\
  \\
  \hspace*{0.5cm}For each question-answer pair, use EXACTLY this format:\\
  \hspace*{0.5cm}\text{<question from=``[Asker Role]'' to=``[Target Role]'' round=``[Round]''>Question text here</question>}\\
  \hspace*{0.5cm}\text{<answer from=``[Answerer Role]'' to=``[Original Asker]'' round=``[Round]''>}Answer text here (or ``Pending'' if unanswered)\text{</answer>}\\
  \\
  \hspace*{0.5cm}Example:\\
  \hspace*{0.5cm}\text{<question from=``Gastroenterologist'' to=``Hematologist'' round=``1''>}Could anemia be secondary to systemic condition?\text{</question>}\\
  \hspace*{0.5cm}\text{<answer from=``Hematologist'' to=``Gastroenterologist'' round=``2''>}Yes, the elevated BUN suggests systemic involvement...\text{</answer>}\\
  \\
  \hspace*{0.5cm}- Always include the ``from'' attribute showing WHO asked/answered\\
  \hspace*{0.5cm}- Always include the ``to'' attribute showing the TARGET recipient\\
  \hspace*{0.5cm}- If a question is unanswered, use: \text{<answer from=``[Target]'' to=``[Asker]'' round=``N''>Pending</answer>}\\
  \hspace*{0.5cm}- Pair by target role and round ordering\\
\\
- Include a brief global section on agreements, disagreements, resolved vs unresolved questions, and remaining uncertainties.\\
- Be concise and information-dense. Return the summary within \text{<summary\_log>...</summary\_log>} only.\\
\end{tcolorbox}

\subsection{Zero/few-shot Prompt} \label{appx:zero-shot prompt}
We apply the same prompt template for zero-shot, zero-shot CoT, few-shot, and few-shot CoT following Wu et al. \cite{wu2025medcasereasoning}. Zero-shot CoT \cite{kojima2022large} prompt directly includes ``Let’s think step by step'' at the end of zero-shot prompt to facilitate inference. Few-shot CoT \cite{wei2022chain} integrates rationales before producing the diagnosis.
\begin{tcolorbox}[
  enhanced,
  breakable,
  boxrule=1pt,
  boxsep=5pt,
  arc=4pt,
  title=\textit{Zero/few-shot Prompt}
]
\scriptsize

\noindent Read the following case presentation and give the most likely diagnosis. \\
First, provide your internal reasoning for the diagnosis within the tags \text{<think> ... </think>}. \\
Then, output the final diagnosis (just the name of the disease/entity) within the tags \text{<answer> ... </answer>}. \\ 
\\
---------------------------------------CASE PRESENTATION ---------------------------------------\\
\{case\_presentation\} \\
---------------------------------------OUTPUT TEMPLATE ---------------------------------------\\
\\
\text{<think>} \\
...your internal reasoning for the diagnosis... \\
\text{</think>} \\
\\
\\
\text{<answer>} \\
...the name of the disease/entity...  \\
\text{</answer>} \\
\end{tcolorbox}

\subsection{LLM-as-a-Judge Evaluation Prompt} \label{appx:llm-judge-prompt}

For diagnostic accuracy evaluation, we apply the same LLM-as-a-judge prompt used in MedCaseReasoning \cite{wu2025medcasereasoning}.

\begin{tcolorbox}[
  enhanced,
  breakable,
  boxrule=1pt,
  boxsep=5pt,
  arc=4pt,
  title=\textit{LLM-as-a-Judge Prompt}
]
\scriptsize

\noindent Is this predicted final diagnosis correct (yes/no)?\\
Predicted final diagnosis: \text{\{prediction\}}\\
True final diagnosis: \text{\{truth\}}\\
Answer [yes/no]
\end{tcolorbox}

\end{document}